\documentclass[10pt,twocolumn,letterpaper]{article}
\usepackage{cvpr} 

\usepackage[accsupp]{axessibility}
\usepackage{times}
\usepackage{epsfig}
\usepackage{graphicx}
\usepackage[export]{adjustbox}
\usepackage{verbatim}
\usepackage{amsmath}
\usepackage{amssymb}
\usepackage{colortbl, color, xcolor}
\usepackage{makecell}
\usepackage{booktabs}
\usepackage{multirow}

\definecolor{cvprblue}{rgb}{0.21,0.49,0.74}
\usepackage[pagebackref,breaklinks,colorlinks,allcolors=cvprblue]{hyperref}

\usepackage{amstext} 
\usepackage{array}   
\newcolumntype{L}{>{$}c<{$}} 

\newcommand{\dataset}{RealBokeh\xspace}
\newcommand{\sdataset}{\dataset}

\newcommand{\fnum}[1]{\emph{f/#1}}
\newcommand{\fstop}{\emph{f}-stop\xspace}
\newcommand{\fstops}{\emph{f}-stops\xspace}
 
\def\cvprPaperID{NTIRE80}

\usepackage[capitalise]{cleveref}

\begin{document}

\title{The First Controllable Bokeh Rendering Challenge at NTIRE 2026}

\author{
Tim Seizinger$^\dagger$ \and Florin-Alexandru Vasluianu$^\dagger$ \and  Jeffrey Chen$^\dagger$ \and Zhuyun Zhou$^\dagger$ \and Zongwei Wu$^\dagger$ \and Radu Timofte$^\dagger$ \and
Dafeng Zhang
\and
Yipeng Lin
\and
Qi Yan
\and
Junhao Chen
\and
Yang Yang
\and
Divyavardhan Singh
\and
Hariom Thacker
\and
Hammad Mohammad
\and
Aanchal Maurya
\and
Kishor Upla
\and
Kiran Raja
\and
Wei Zhou
\and
Hongyu Huang
\and
Yujin Cho
\and
Grigory Malivenko
\and
Jiachen Tu
\and
Yaokun Shi
\and
Guoyi Xu
\and
Yaoxin Jiang
\and
Jiajia Liu
}

\maketitle


\begin{abstract}
This study presents the outcomes of the first Controllable Bokeh Rendering Challenge at NTIRE and highlights the most effective submitted methodologies.
In total, 44 participants registered for the competition, of which 8 teams submitted valid solutions after the conclusion of the final test phase.
All submissions were evaluated on unseen images, focusing on portraits and intricate subjects with complex and visually appealing bokeh phenomena.
In addition to the first track focusing on established quantitative fidelity metrics, we conducted a qualitative user study with a panel of experts for a second track focusing on perceptual assessment.
As this was the inaugural challenge on this topic, most of the participants focused on refining and extending the Bokehlicious baseline method.
\end{abstract}

\let\thefootnote\relax\footnotetext{
\hspace{-5mm} 
$\dagger$ 
Tim Seizinger, Florin-Alexandru Vasluianu, Jeffrey Chen, Zhuyun Zhou, Zongwei Wu, and Radu Timofte are the NTIRE 2026 challenge organizers. 
The other authors participated in the challenge.
\\
\cref{sec:team} (in the supplementary) provides names and affiliations.
\\ 
\url{https://cvlai.net/ntire/2026}
}

\section{Introduction}

\textit{Bokeh} is one of the most distinct aspects that separates the visual aesthetics of smartphone photography from those associated with professional cameras.
The term is derived from the Japanese word \textit{boke}, meaning blur or haze, and refers to the smooth and pleasant appearance of out-of-focus image backgrounds. 
This effect emphasizes the subject by transforming potentially distracting backgrounds into visually pleasing smooth blurs in low-contrast or intricate circular patterns in high-contrast areas~\cite{kennerdell1997bokeh}.

In the photographic process, exact control over the strength of the Bokeh effect is crucial.
For professional cameras with large dedicated lenses, this is achieved with a variable circular \textit{aperture} diaphragm mechanism inside the optical path of the lens.
The settings of this aperture mechanism are denoted in so-called \textit{\fstops}, which represent the ratio between the focal-length of the lens and the diameter of the aperture opening.
Small \fstops such as \fnum{2.0} indicate an open aperture, which results in a strong Bokeh effect.
To weaken the strength of the effect, the aperture is gradually closed down, which is indicated by increasingly \textit{larger} \fstop numbers such as \fnum{4.0} or \fnum{8.0}~\cite{mercado2019filmmaker}.

While Bokeh naturally occurs with large DSLM cameras and wide-aperture lenses, it is barely noticeable on smartphones due to their tiny optical systems.
This necessitates computational systems to render an equivalent Bokeh effect onto smartphone images~\cite{wadhwa2018synthetic, ignatov2020rendering, barron2015fast}.

The origins of Bokeh rendering are not in the field of computational photography, but rather in computer graphics, with early work simulating Bokeh in the context of 3D rendering~\cite{cook1984distributed, potmesil1981lens}.
Later methods adapt classical rendering to the image space, often focusing on stability, since unlike in a 3D scene, semantic information is not always accurate or available~\cite{bae2007defocus, peng2022mpib, BokehMeHybrid, peng2021interactive}, necessitating specialized camera hardware~\cite{hach2015cinematic, wadhwa2018synthetic} or computational pre-processing~\cite{sheng2024dr, zhang2019synthetic}.
Other works aim to speed up the rendering process~\cite{barron2015fast, zhu2017fast} or improve the visual fidelity of the effect~\cite{kraus2007depth, luo2023defocus, wadhwa2018synthetic, busam2019sterefo, peng2024bokehme++, hach2015cinematic, zhang2019synthetic}.
However, the Bokeh produced by classical rendering methods is usually based on hand-crafted kernels, limiting realism~\cite{ignatov2020rendering, seizinger2025bokehlicious}.

More recently, Bokeh rendering approaches that incorporate neural rendering mechanisms have gained popularity. 
Although some combine neural and classical rendering to alleviate artifact and stability issues~\cite{BokehMeHybrid, peng2024bokehme++}, others attempt to fully replicate the Bokeh effect in its entirety~\cite{ignatov2020rendering, wang2018deeplens, zhu2025bokehdiff} by learning from paired small- and large-aperture images.
This can offer numerous benefits, with some work focusing on photorealistic Bokeh Effect reproduction~\cite{ignatov2020aim, ignatov2020rendering, seizinger2025bokehlicious, Mandl2024NeuralBokeh} that utilizes real-world data, forgoing semantic priors~\cite{seizinger2025bokehlicious, BokehLossGAN, huang2026bokehflow} or enabling control over Bokeh generation similarly to classical methods~\cite{seizinger2023bokeh, zhu2025bokehdiff, mu2025generative}.

As of late. controllable neural Bokeh rendering has received increased attention. 
Early large-scale datasets only modeled a binary scenario with Bokeh-free inputs and fixed-aperture references showing strong Bokeh effects~\cite{ignatov2020rendering}; consequently, many methods~\cite{BokehLossGAN, dutta2021stackedbokeh, nagasubramaniam2023BEViT, purohit2019depth} offer no way to adjust the Bokeh strength.
Although several works have explored controllability using synthetic training data generated with 3D-based rendering solutions~\cite{zhu2025bokehdiff, seizinger2023bokeh, conde2023lens, kong2023bokeh}, such solutions can suffer from a domain gap when applied to real images~\cite{peng2024bokehme++} and attempt to mitigate this by relying on explicit depth-priors~\cite{zhu2025bokehdiff, mu2025generative}.

Bokeh rendering is a technically challenging task, as it requires: 
\textbf{(1)} detecting the image subject and maintaining its sharpness throughout the rendering process~\cite{sheng2024dr, wadhwa2018synthetic}, 
\textbf{(2)} highly detailed spatial awareness to separate fine structures~\cite{zhu2025bokehdiff, seizinger2025bokehlicious}, 
\textbf{(3)} implicit high dynamic range modeling of light intensities and colors~\cite{zhang2019synthetic}, 
\textbf{(4)} background inpainting~\cite{peng2022mpib, sheng2024dr}, 
\textbf{(5)} accurate modeling of the intricate space-varying point spread function (PSF) of a photographic lens~\cite{hach2015cinematic, seizinger2025bokehlicious, Mandl2024NeuralBokeh}, and 
\textbf{(6)} correctly rendering potentially extensive PSFs while simulating complex hierarchical Bokeh phenomena such as \textit{eclipsing}~\cite{debevec2020phenonenon}.

With our previous work presenting the \textbf{Bokehlicious} architecture and the \textbf{RealBokeh} dataset, we have taken the foundational step towards photorealistic and controllable Bokeh rendering.
Now, with the \textit{first Controllable Bokeh Rendering Challenge at NTIRE}, we hope to excite fellow researchers for this interesting and challenging application.

\begin{figure*}[!ht]
    \centering
    \small
    \setlength{\tabcolsep}{1pt}
    \renewcommand{\arraystretch}{0.7}
    \def\widthcomp{0.19}
    \def\hcomp{2mm}
    \begin{tabular}{cccccc}
    & Input \fnum{22.0} & GT \fnum{14.0} & GT \fnum{4.0} & GT \fnum{2.5} & GT \fnum{2.0}  \\
    \addlinespace[0.2pt]
    \rotatebox{90}{\hspace{\hcomp}\footnotesize Focallength 35mm} & \includegraphics[width=\widthcomp\linewidth]{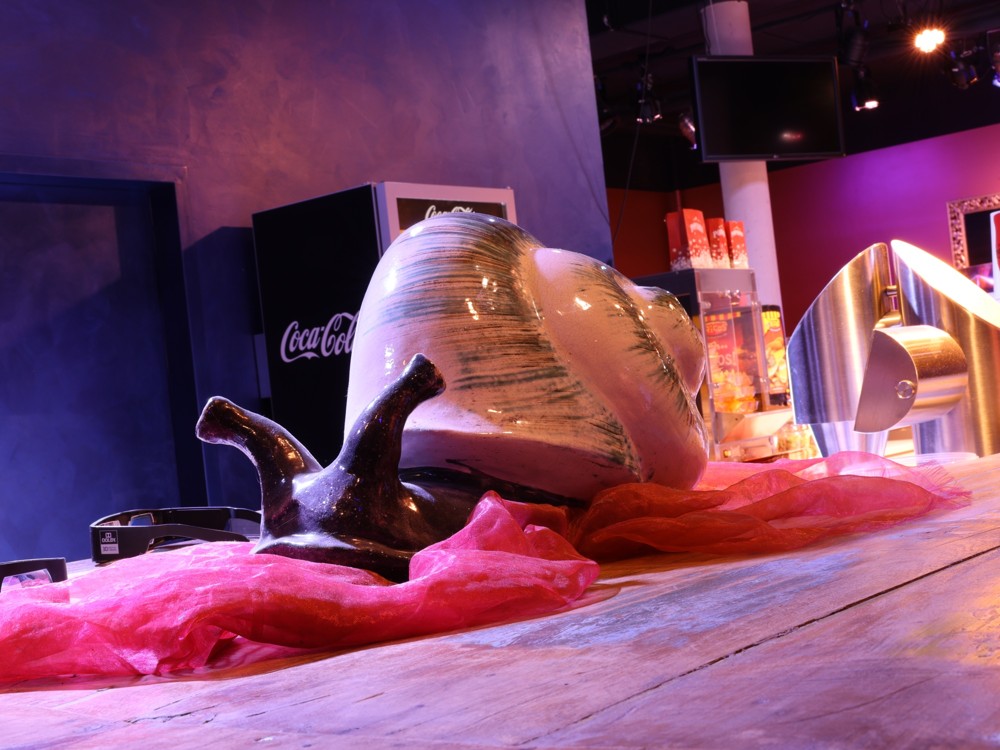} & \includegraphics[width=\widthcomp\linewidth]{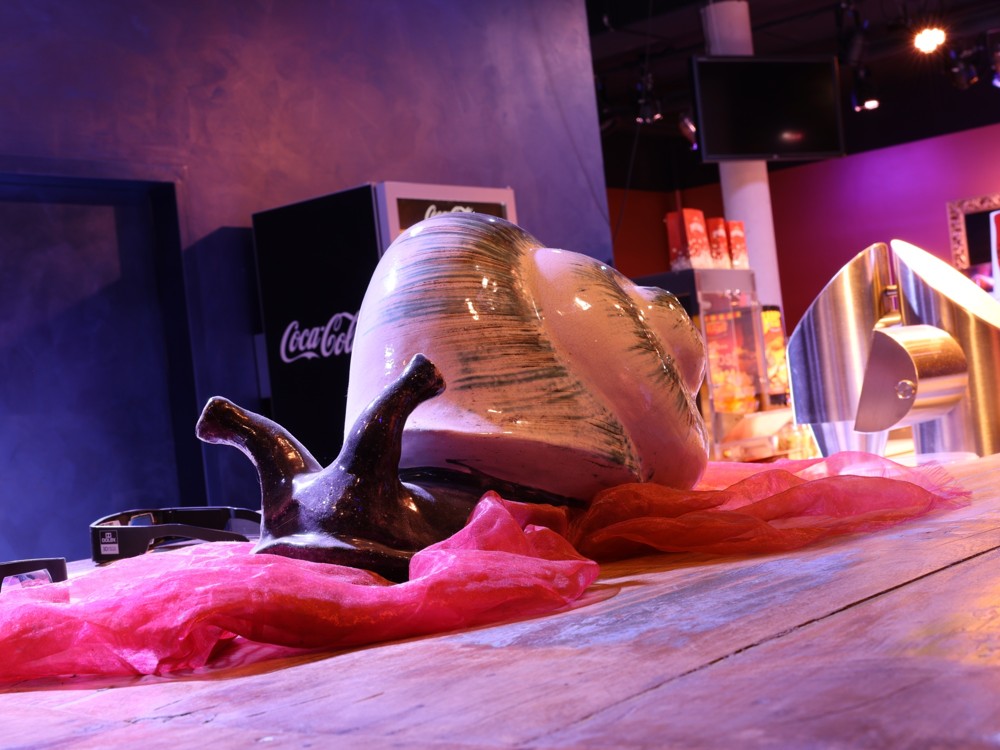} & \includegraphics[width=\widthcomp\linewidth]{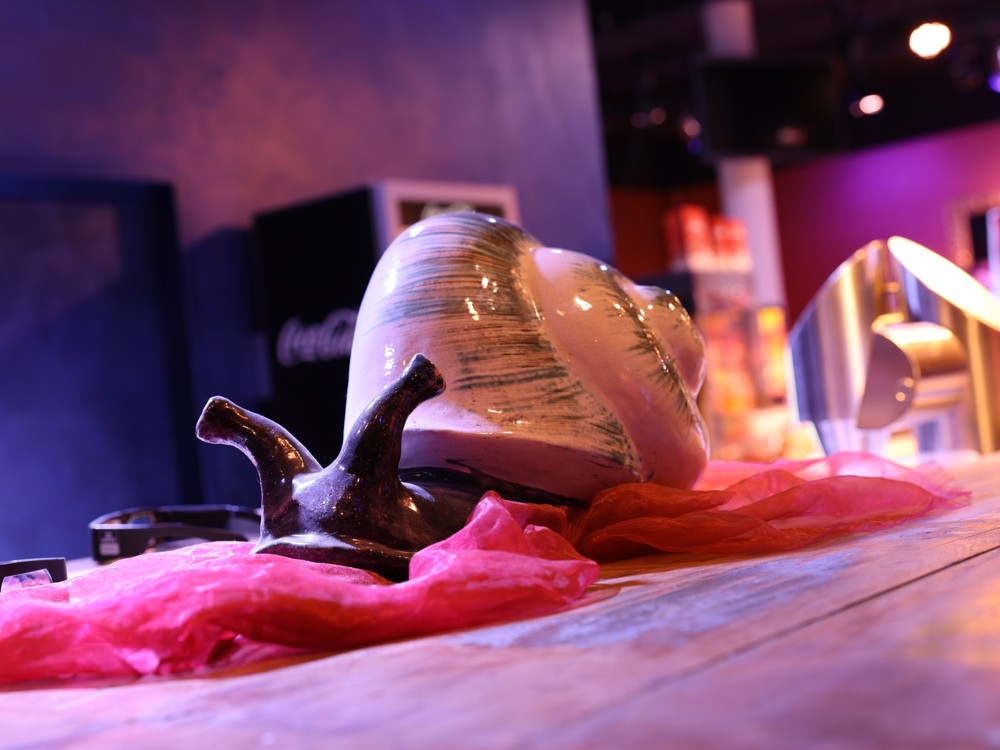} & \includegraphics[width=\widthcomp\linewidth]{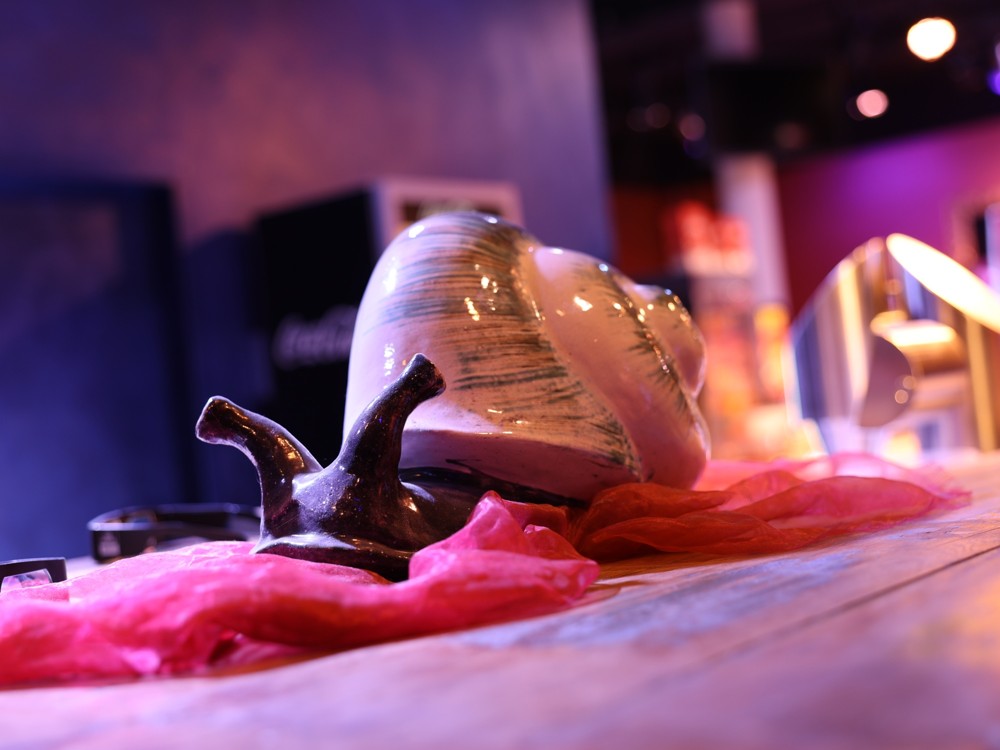} & \includegraphics[width=\widthcomp\linewidth]{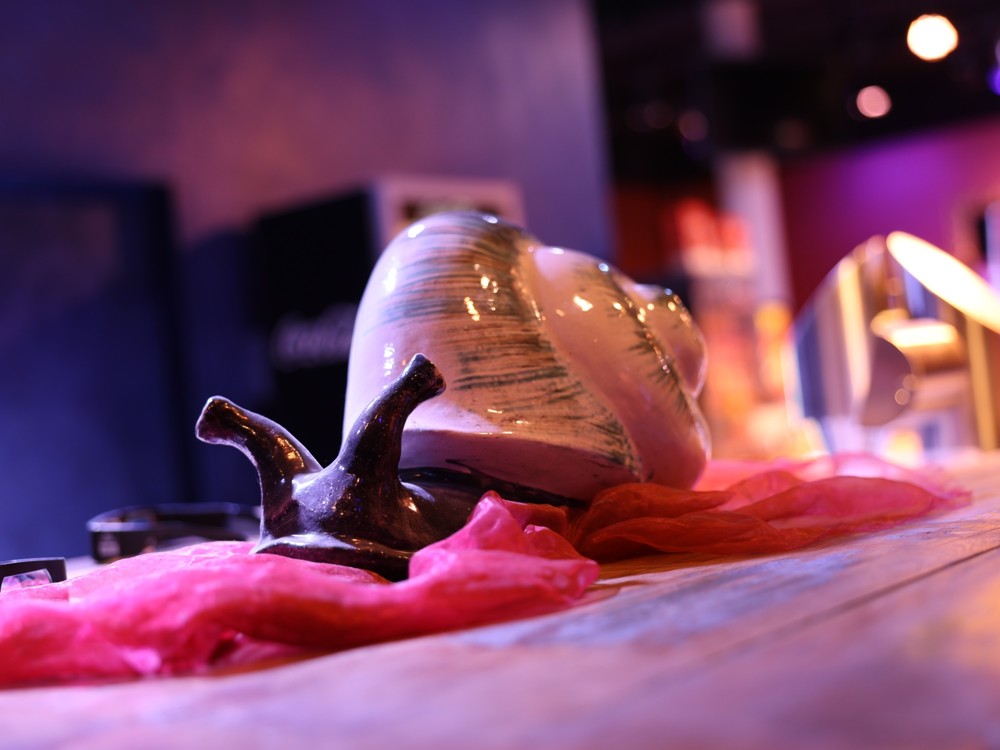} \\
    \addlinespace[0.5pt]
    &Input \fnum{22.0} & GT \fnum{13.0} & GT \fnum{8.0} & GT \fnum{5.0} & GT \fnum{2.0}  \\
    \addlinespace[0.2pt]
    \rotatebox{90}{\hspace{\hcomp}\footnotesize Focallength 52mm} & \includegraphics[width=\widthcomp\linewidth]{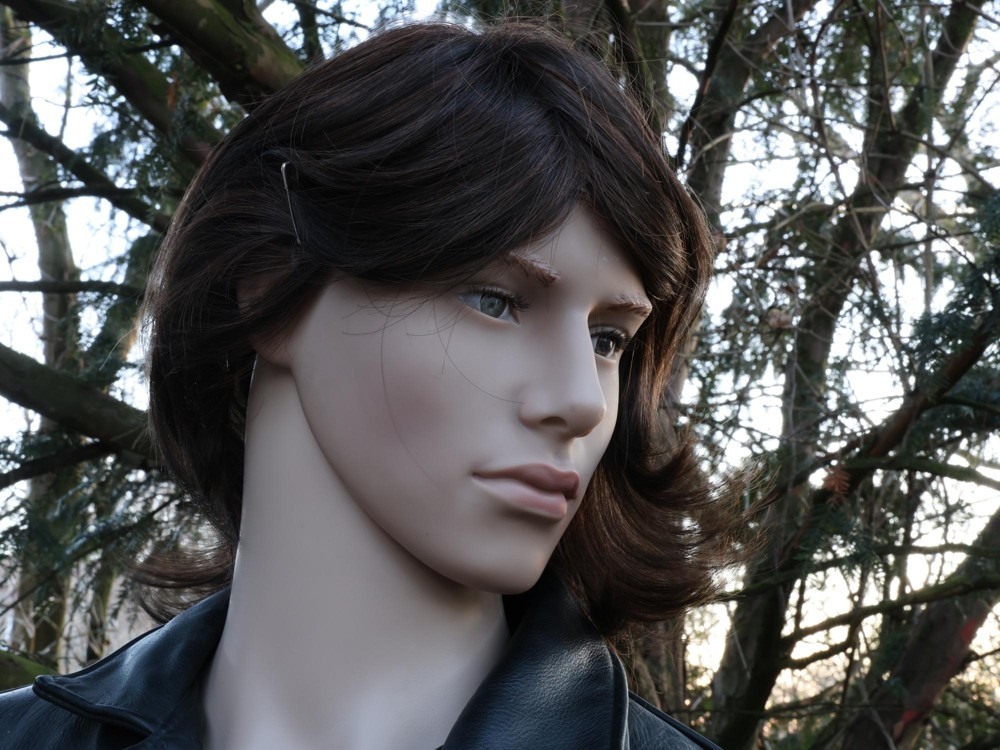} & \includegraphics[width=\widthcomp\linewidth]{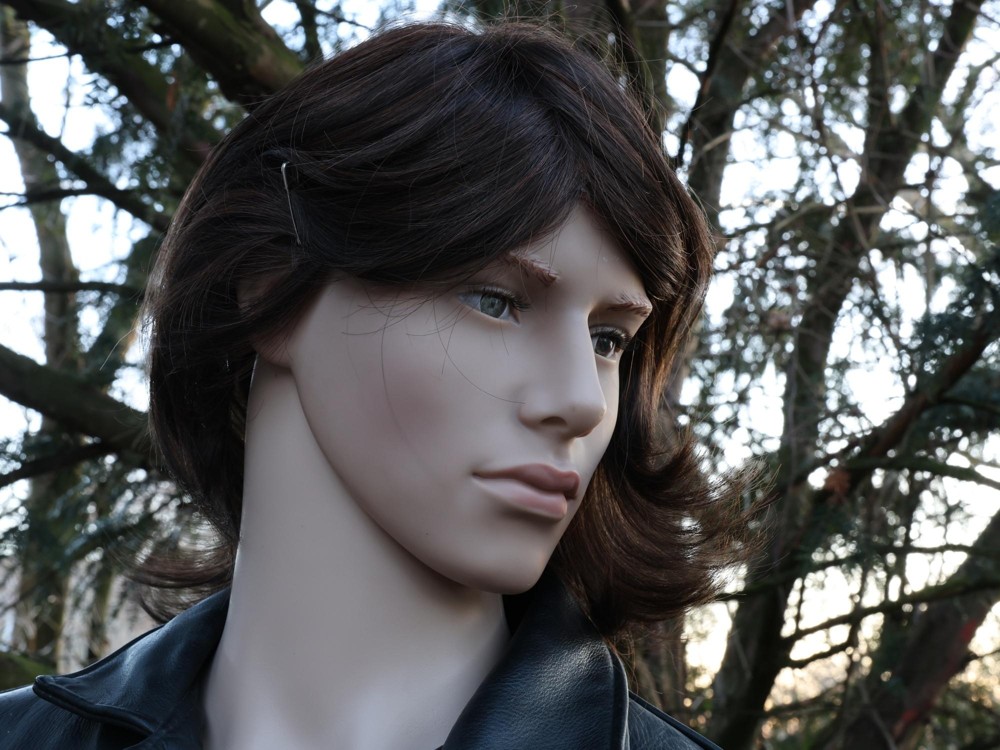} & \includegraphics[width=\widthcomp\linewidth]{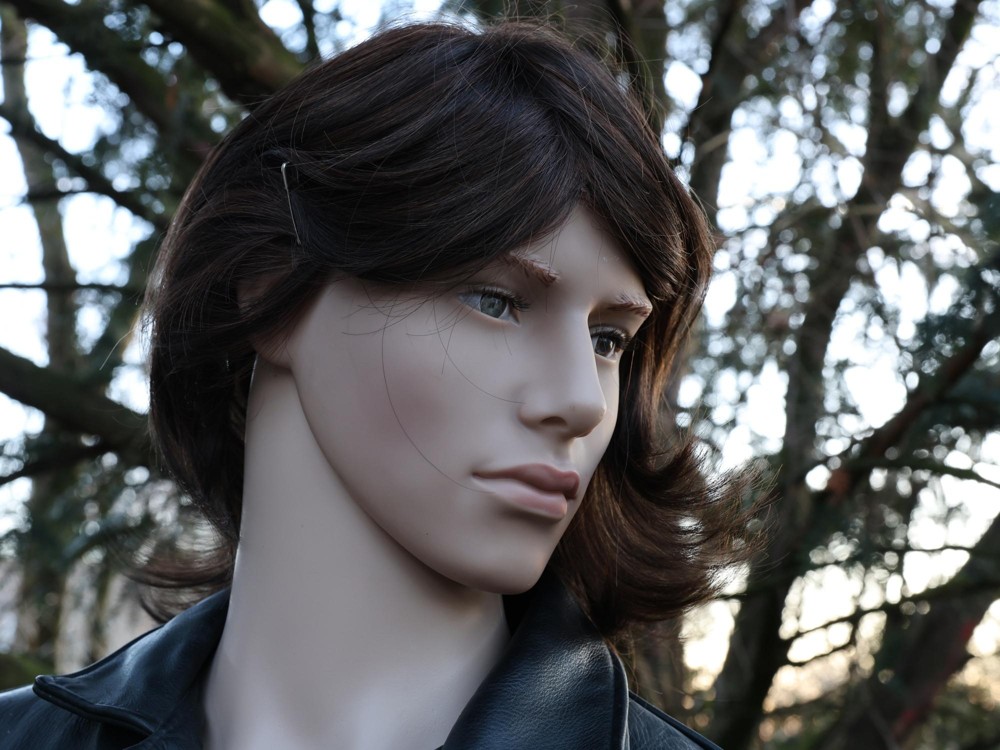} & \includegraphics[width=\widthcomp\linewidth]{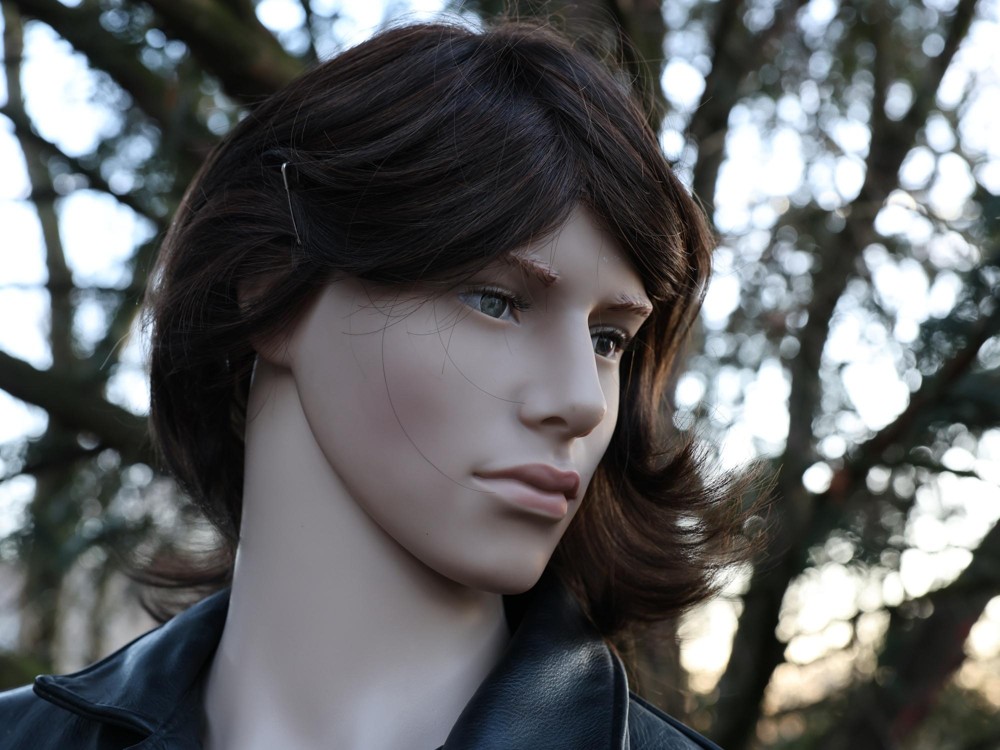} & \includegraphics[width=\widthcomp\linewidth]{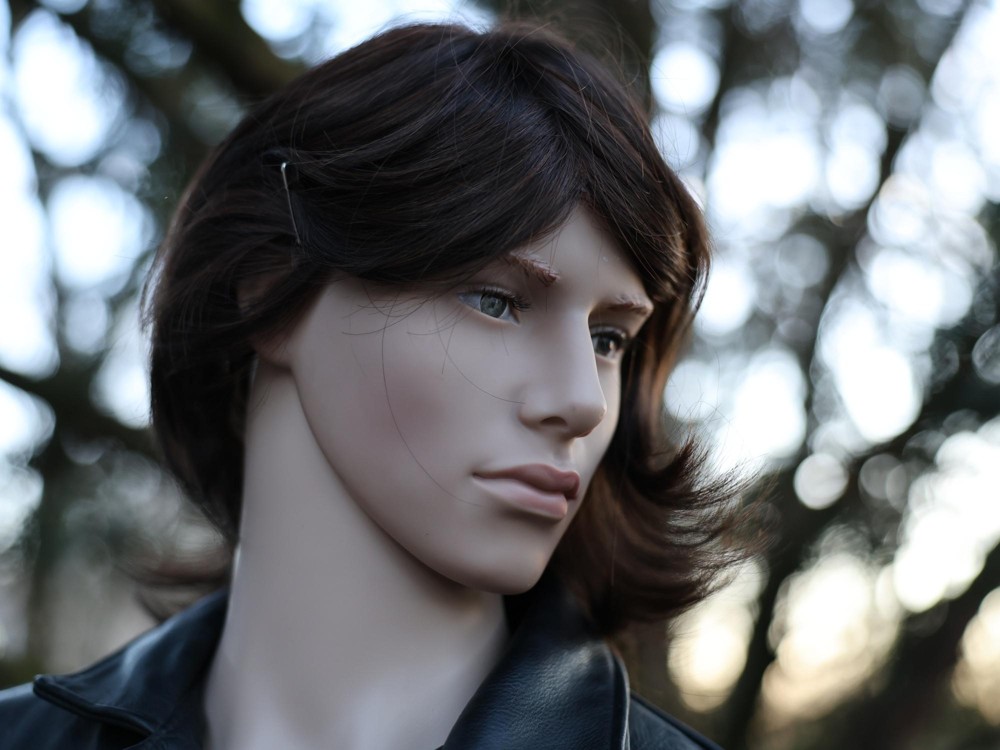} \\
    \addlinespace[0.5pt]
    &Input \fnum{22.0} & GT \fnum{7.1} & GT \fnum{5.0} & GT \fnum{2.2} & GT \fnum{2.0}  \\
    \addlinespace[0.2pt]
    \rotatebox{90}{\hspace{\hcomp}\footnotesize Focallength 70mm} & \includegraphics[width=\widthcomp\linewidth]{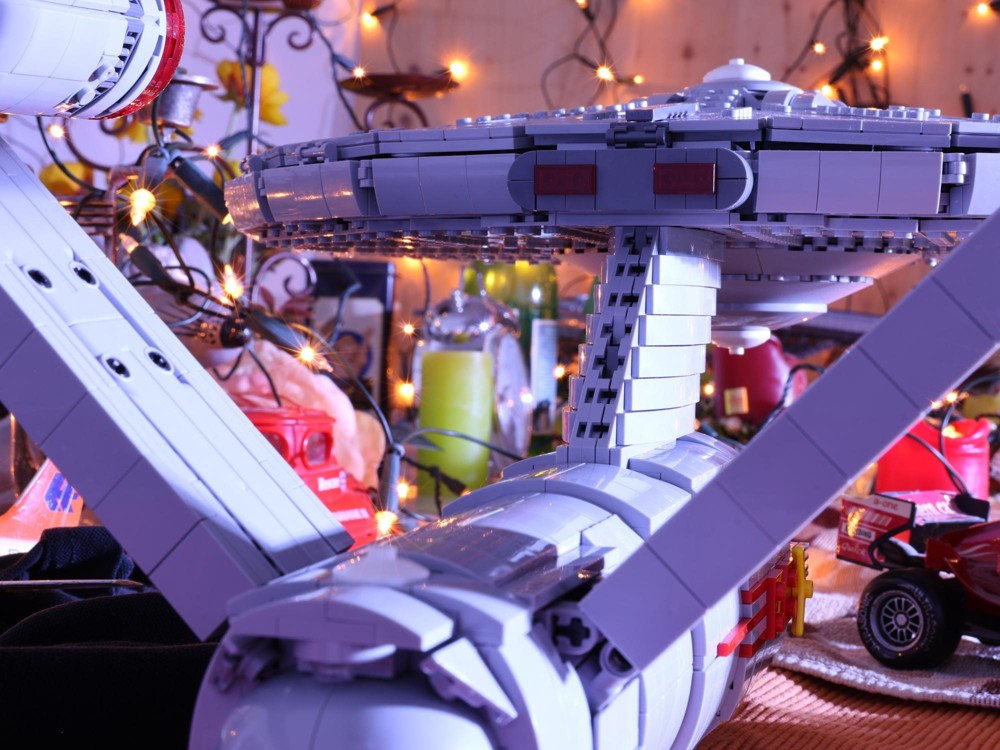} &  \includegraphics[width=\widthcomp\linewidth]{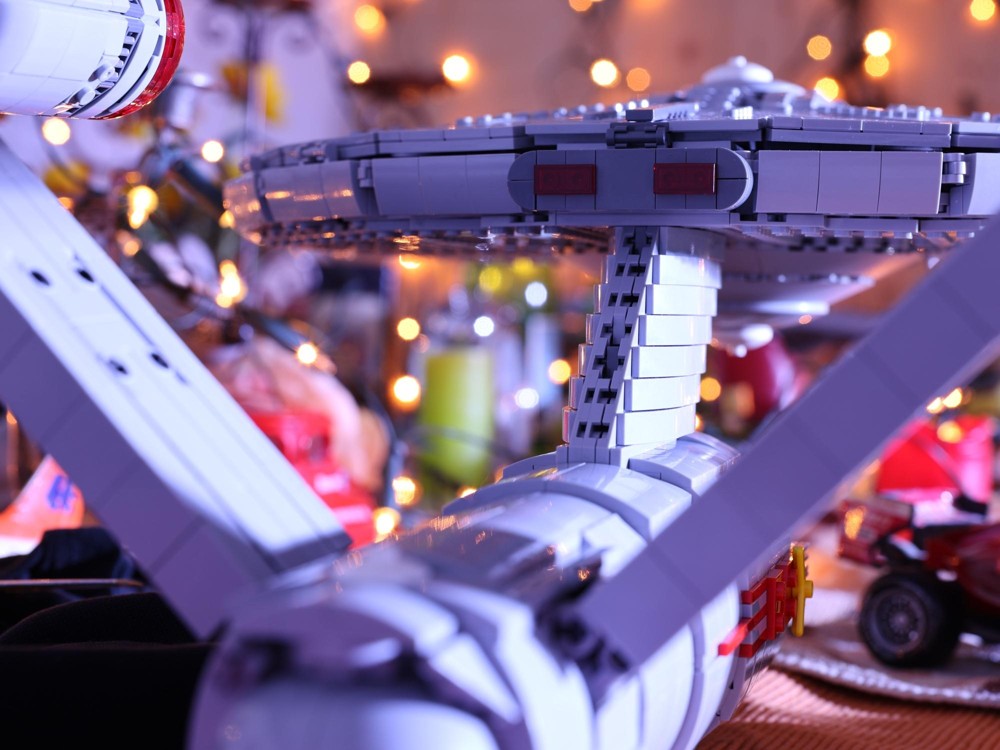} & \includegraphics[width=\widthcomp\linewidth]{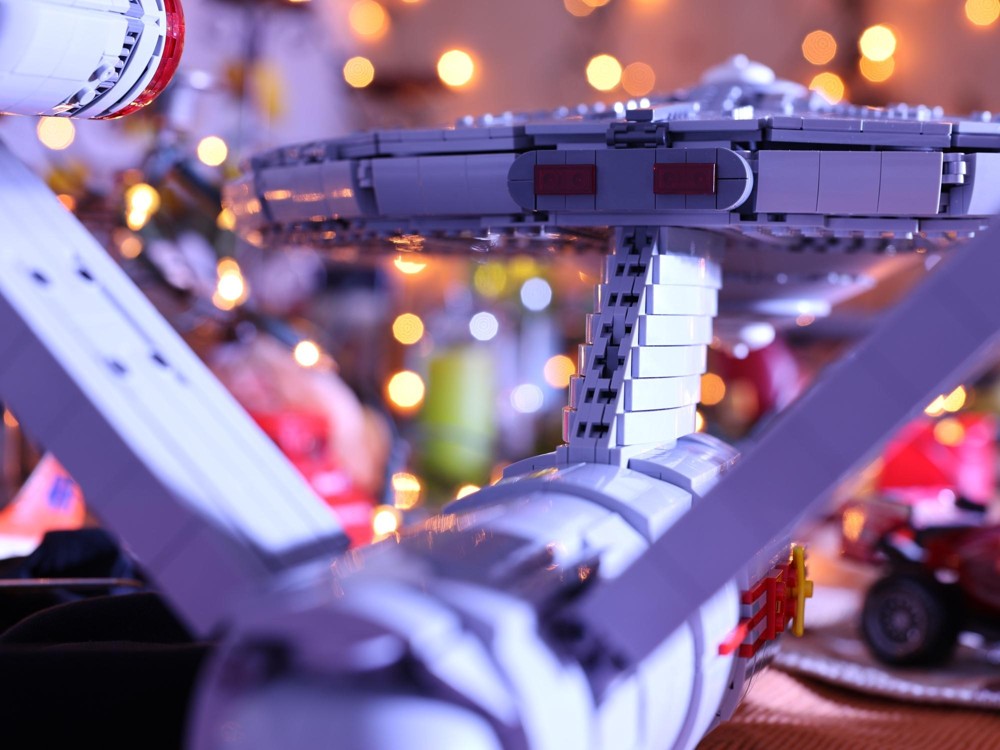} & \includegraphics[width=\widthcomp\linewidth]{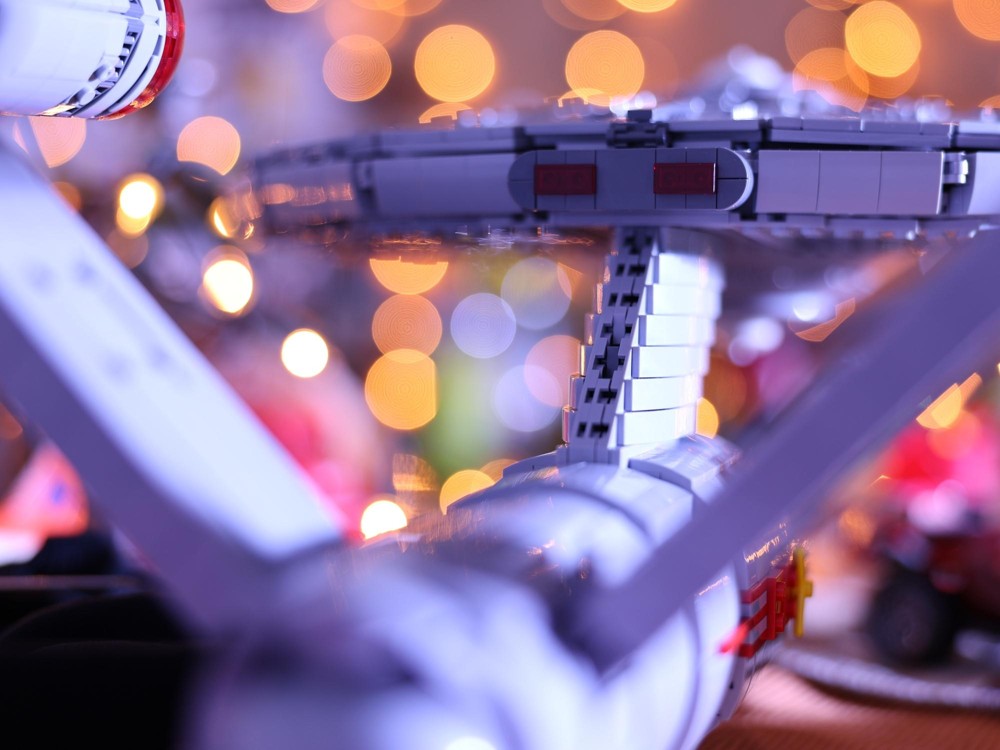} & \includegraphics[width=\widthcomp\linewidth]{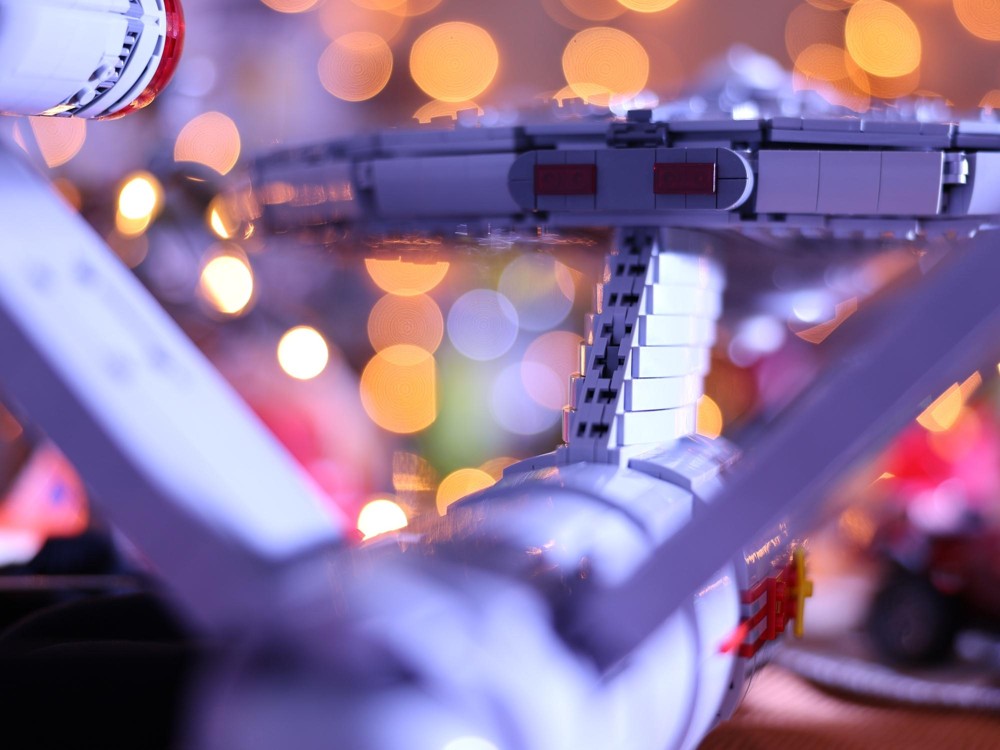} \\
    \end{tabular}
    \vspace{-0.3cm}
    \caption{Sample capture sequences from the \textbf{\textit{\dataset}} dataset. 
    The first image in the scene capture protocol is taken at \fnum{22.0} and the last image is always a \fnum{2.0} reference image while all captures in between are randomly sampled and shot in order of decreasing \fstop number, showing an increasingly stronger Bokeh effect.
    }
    \vspace{-3.5mm}
    \label{fig:RealBokeh}
\end{figure*}

This challenge is one of the challenges associated with the NTIRE 2026 Workshop~\footnote{\url{https://www.cvlai.net/ntire/2026/}} on:
deepfake detection~\cite{ntire26deepfake}, 
high-resolution depth~\cite{ntire26hrdepth},
multi-exposure image fusion~\cite{ntire26raim_fusion}, 
AI flash portrait~\cite{ntire26raim_portrait}, 
professional image quality assessment~\cite{ntire26raim_piqa},
light field super-resolution~\cite{ntire26lightsr},
3D content super-resolution~\cite{ntire263dsr},
bitstream-corrupted video restoration~\cite{ntire26videores},
X-AIGC quality assessment~\cite{ntire26XAIGCqa},
shadow removal~\cite{ntire26shadow},
ambient lighting normalization~\cite{ntire26lightnorm},
rip current detection and segmentation~\cite{ntire26ripdetseg},
low light image enhancement~\cite{ntire26llie},
high FPS video frame interpolation~\cite{ntire26highfps},
Night-time dehazing~\cite{ntire26nthaze,ntire26nthaze_rep},
learned ISP with unpaired data~\cite{ntire26isp},
short-form UGC video restoration~\cite{ntire26ugcvideo},
raindrop removal for dual-focused images~\cite{ntire26dual_focus},
image super-resolution (x4)~\cite{ntire26srx4},
photography retouching transfer~\cite{ntire26retouching},
mobile real-word super-resolution~\cite{ntire26rwsr},
remote sensing infrared super-resolution~\cite{ntire26rsirsr},
AI-Generated image detection~\cite{ntire26aigendet},
cross-domain few-shot object detection~\cite{ntire26cdfsod},
financial receipt restoration and reasoning~\cite{ntire26finrec},
real-world face restoration~\cite{ntire26faceres},
reflection removal~\cite{ntire26reflection},
anomaly detection of face enhancement~\cite{ntire26anomalydet},
video saliency prediction~\cite{ntire26videosal},
efficient super-resolution~\cite{ntire26effsr},
3d restoration and reconstruction in adverse conditions~\cite{ntire26realx3d},
image denoising~\cite{ntire26denoising},
blind computational aberration correction~\cite{ntire26aberration},
event-based image deblurring~\cite{ntire26eventblurr},
efficient burst HDR and restoration~\cite{ntire26bursthdr},
low-light enhancement: `twilight cowboy'~\cite{ntire26twilight},
and efficient low light image enhancement~\cite{ntire26effllie}.


\section{Challenge Data}

\begin{table}[t]
\centering
\small
\begin{tabular}{@{}l|cc}
\toprule
                    & \textbf{\sdataset~\cite{seizinger2025bokehlicious} - \textit{NTIRE26}}   \\
\midrule
Total Samples       & 20,646                                \\
Training Samples    & 20,500                                \\
Validation Samples  & 78                                    \\
Test Samples        & 68                                    \\
Apertures           & \fnum{20.0} - \fnum{2.0}              \\
Focal Length        & 28mm - 70mm                           \\
Resolution          & 2000$\times$1500                      \\
Metadata            & f-stop, focus-distance, focal-length  \\
Public              & Train-only                            \\ 
\bottomrule
\end{tabular}
\vspace{-2mm}
\caption{Statistics on the \sdataset-\textit{NTIRE26} dataset used by this challenge for training, validation and final testing. The \sdataset dataset was originally proposed by Seizinger \etal~\cite{seizinger2025bokehlicious}.}
\label{tab:datasets}
\vspace{-3mm}
\end{table}

To advance neural Bokeh rendering, large-scale, well-aligned, and diverse real-world datasets are essential. 
However, many previous datasets lack realism, scale and alignment quality~\cite{zhang2019synthetic,ignatov2020rendering}. 
Our previous work \cite{seizinger2025bokehlicious} introduced \textit{\dataset}, comprising 23K wide-/small-aperture image pairs across 4.4K scenes, being the first real-world dataset covering varying aperture \fstops and focal-lengths.
The dataset was captured by expert photographers using a Canon EOS R6 II with a 28–70mm lens with a maximum aperture of \fnum{2.0} using an automated capture protocol to maintain good sample alignment.

For each scene, five images are acquired, as shown in \cref{fig:RealBokeh}. 
One input at \fnum{22.0}, three intermediate ground truths at random \fstops between \fnum{20.0} and \fnum{2.2} (1/3-stop), and one at \fnum{2.0}, capturing realistic lens variability~\cite{allen2012manual} and a final reference image shot at \fnum{2.0}.

For the purpose of this challenge, we extend our initial RealBokeh data set by replacing the validation and test splits with 146 new scenes as shown in \cref{tab:datasets} following the original capture protocol.
The reference images of these new scenes remain private to ensure fair competition among the challenge participants.

\section{Evaluation}
This challenge was organized as a dual-track competition, with each track evaluated according to its own dedicated ranking criteria. The corresponding assessment criteria are detailed below.

\begin{itemize}
    \item The reconstruction fidelity in terms of PSNR;
    \item The Structural  Similarity Index (SSIM) \cite{wang2004image} score;
    \item The LPIPS \cite{zhang2018unreasonable} distance between the rendered and the ground-truth images. We used ImageNet pretrained AlexNet \cite{krizhevsky2012imagenet} for LPIPS feature extraction;
    \item The Mean Opinion Score (MOS) from a panel of expert photographers.
\end{itemize}

\noindent The two evaluation tracks are defined as follows:

\begin{itemize}
    \item \textbf{Fidelity Track:} The Fidelity Track ranks solutions using well-established quantitative metrics such as PSNR, SSIM, and LPIPS, weighting them equally.
    \item \textbf{Perceptual Track:} The Perceptual Track ranks solutions based on a Mean Opinion Score (MOS) of their rendered Bokeh effect, as determined by a panel of photography experts.
\end{itemize}

\section{Challenge Phases}
\begin{enumerate}
    \item \textbf{Development phase:} In the first phase, participants received the RealBokeh training dataset of 20,500 samples along with an additional 78 validation samples with private ground truths that were captured with the same optical system as RealBokeh. 
    To enable measurement of validation performance, participants uploaded their predictions to a remote CodaBench evaluation server~\cite{xu2022codabench}.
    
    \item \textbf{Test phase:} A second set of 68 additional samples was sent to the challenge participants, with evaluation against the private ground truths again performed via a remote CodaBench server~\cite{xu2022codabench}. 
    Fine-tuning of the Test-set was restricted by limiting the number of submissions.
    For the final submission, each team provided a method description, their code with checkpoints, information about the team members, and the final set of 68 predictions of the private testing split.
    
\end{enumerate}

\begin{table*}[!ht]
    \centering
    \footnotesize
        \begin{tabular}{l|ccc|c|ccc|c|c}
        \toprule
        Team & \makecell{PSNR \\ $\uparrow$} & \makecell{SSIM \\ $\uparrow$} & \makecell{LPIPS \\ $\downarrow$} & \makecell{MOS \\ $\uparrow$} & \makecell{Params. \\ (M)} & \makecell{Runtime \\ (s)} & Device   & \makecell{Perceptual \\ Rank} & \makecell{Fidelity \\ Rank} \\
        \midrule
        NJUST-KMG         & 31.057$^{1\phantom{0}}$   & 0.9432$^{1\phantom{0}}$   & 0.0909$^{3\phantom{0}}$   & 7.49 & 27.93 & 13.934 &  RTX A6000        & 4 & 1 \\
        YuFans~\cite{ntire26geometric}            & 30.790$^{2\phantom{0}}$   & 0.9421$^{2\phantom{0}}$   & 0.0941$^{5\phantom{0}}$   & 7.54 & 13.96 & 6 &  H800     & 3 & 2 \\
        CV SVNIT          & 30.613$^{4\phantom{0}}$   & 0.9390$^{4\phantom{0}}$   & 0.0841$^{1\phantom{0}}$   & 7.37 & 15.07 & 0.42 & RTX A5000        & 5 & 3 \\
        Davinci           & 30.749$^{3\phantom{0}}$   & 0.9413$^{3\phantom{0}}$   & 0.0921$^{4\phantom{0}}$   & 7.58 & 27 & 5 & A6000        & 1 & 4 \\
        Baseline$_L$~\cite{seizinger2025bokehlicious} & 30.514$^{5\phantom{0}}$   & 0.9369$^{5\phantom{0}}$   & 0.0855$^{2\phantom{0}}$   & 7.56 & 13.96 & 1.63 & RTX 4090 & 2 & 5 \\
        BIT ssvgg         & 29.920$^{6\phantom{0}}$   & 0.9324$^{6\phantom{0}}$   & 0.1202$^{8\phantom{0}}$   & 5.22 &  18  & 0.229  & RTX 4090      & 8 & 6 \\
        Centre Borelli    & 29.853$^{7\phantom{0}}$   & 0.9302$^{7\phantom{0}}$   & 0.0952$^{6\phantom{0}}$   & 5.81 & 20.12 & 0.811 &  V100-SXM2      & 6 & 7 \\
        Baseline$_S$~\cite{seizinger2025bokehlicious} & 29.625$^{8\phantom{0}}$   & 0.9276$^{8\phantom{0}}$   & 0.0957$^{7\phantom{0}}$   & 5.01 & 1.21 &  0.28 & RTX 4090 & 9 & 8 \\
        higasa            & 28.993$^{9\phantom{0}}$   & 0.9262$^{9\phantom{0}}$   & 0.1403$^{9\phantom{0}}$   & 5.75 & 8.65 & 0.2 & 5070 Ti      & 7 & 9 \\
        NTR               & 23.832$^{10}$  & 0.8458$^{10}$  & 0.2422$^{10}$  & 2.51 & 142.4 & 32 &  H100       & 10 & 10 \\
        \midrule
        Inputs         & 	20.723$^{11}$  & 		0.7011$^{11}$  & 		0.3885$^{11}$  & - & - & - & -  & 	- & 	11 \\
        \bottomrule
    \end{tabular}
    \caption{
    Quantitative evaluations for the submissions of the First Controllable Bokeh Rendering Challenge at NTIRE 2026 on the Final Phase test split. The superscripts above the performance indicate the ranking in each associated metric.
    }
    \vspace{-3mm}
    \label{tab:quanti_t1}
\end{table*}

\section{User Study}
For the perceptual track, the primary evaluation criterion of the challenge was the Mean Opinion Score (MOS). 
The MOS associated with each method was determined by a panel of four photography experts. 
The rating scale ranged from 1 (lowest quality) to 10 (ground truth quality) and was calibrated so that the unprocessed input image corresponded to a score of 3.
Scores of 1 and 2 were reserved for cases in which the algorithm perceptibly degraded the input image without successfully producing the intended bokeh effect.
In contrast, the quality of the Bokeh rendered in the images was assigned scores between 3 and 10, in increments of 0.5.

According to the expert panel, the most important factors for judging the quality of a Bokeh Rendering solution are:
\begin{itemize}
    \item Maintaining critical sharpness on the subject of the image, especially with regard to fine details such as hair.
    \item Rendering well defined and aesthetically pleasing shapes for out-of-focus highlights (also called bokeh-balls) with accurate color and saturation.
    \item Correctly rendering the gradual sharpness falloff for continuous foreground to background transitions.
    \item Preserving an overall appearance that is free from processing artifacts such as blockiness or ghosting.
\end{itemize}

\begin{figure}[!ht]
    \centering
    \includegraphics[width=0.99\linewidth]{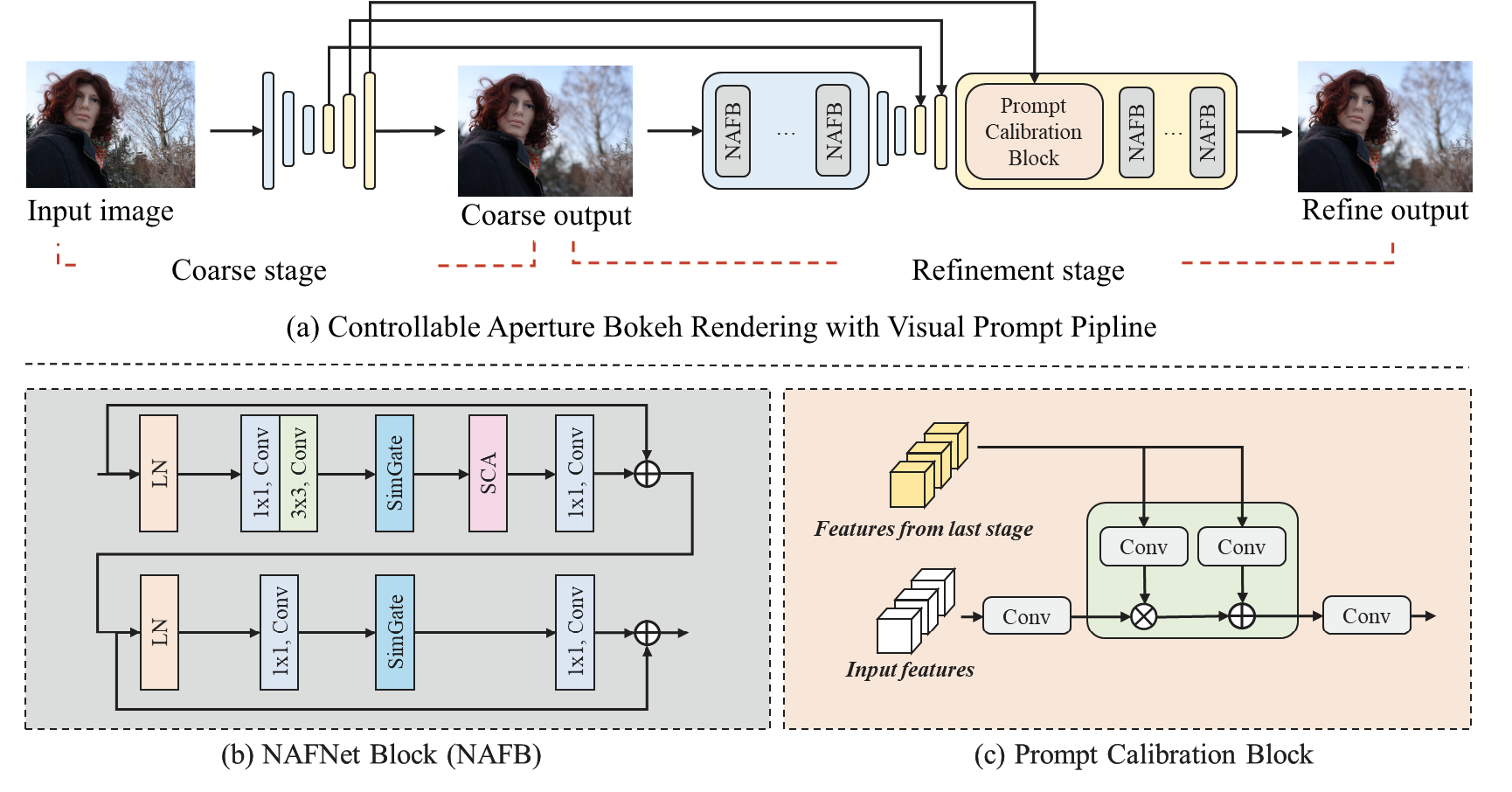}
    \vspace{-3mm}
    \caption{
        The two stage coarse-to-refinement network architecture proposed by Davinci.
    }
    \label{fig:Davinci}
    \vspace{-3mm}
\end{figure}

\section{Challenge Results}
The challenge received eight valid submissions that are presented in \cref{sec:methods}.

\begin{figure*}[ht]
\begin{center}
\renewcommand{\arraystretch}{0.4}
\setlength{\tabcolsep}{0.8pt}
\footnotesize
\def\imgcompp{.1925}
\def\imgcomp{.1214}
\def\widthcompp{.128}
\begin{tabular}[b]{c c c c c c c c}\hspace{-4mm}
    Input &  Baseline$_L$~\cite{seizinger2025bokehlicious} & Davinci & CV SVNIT & YuFans~\cite{ntire26geometric} & NJUST-KMG & GT \\
    \addlinespace[1pt]
    \multirow{3}{*}[1.4mm]{\includegraphics[height=\imgcompp\textwidth, trim={350px 0px 200px 0px},clip]{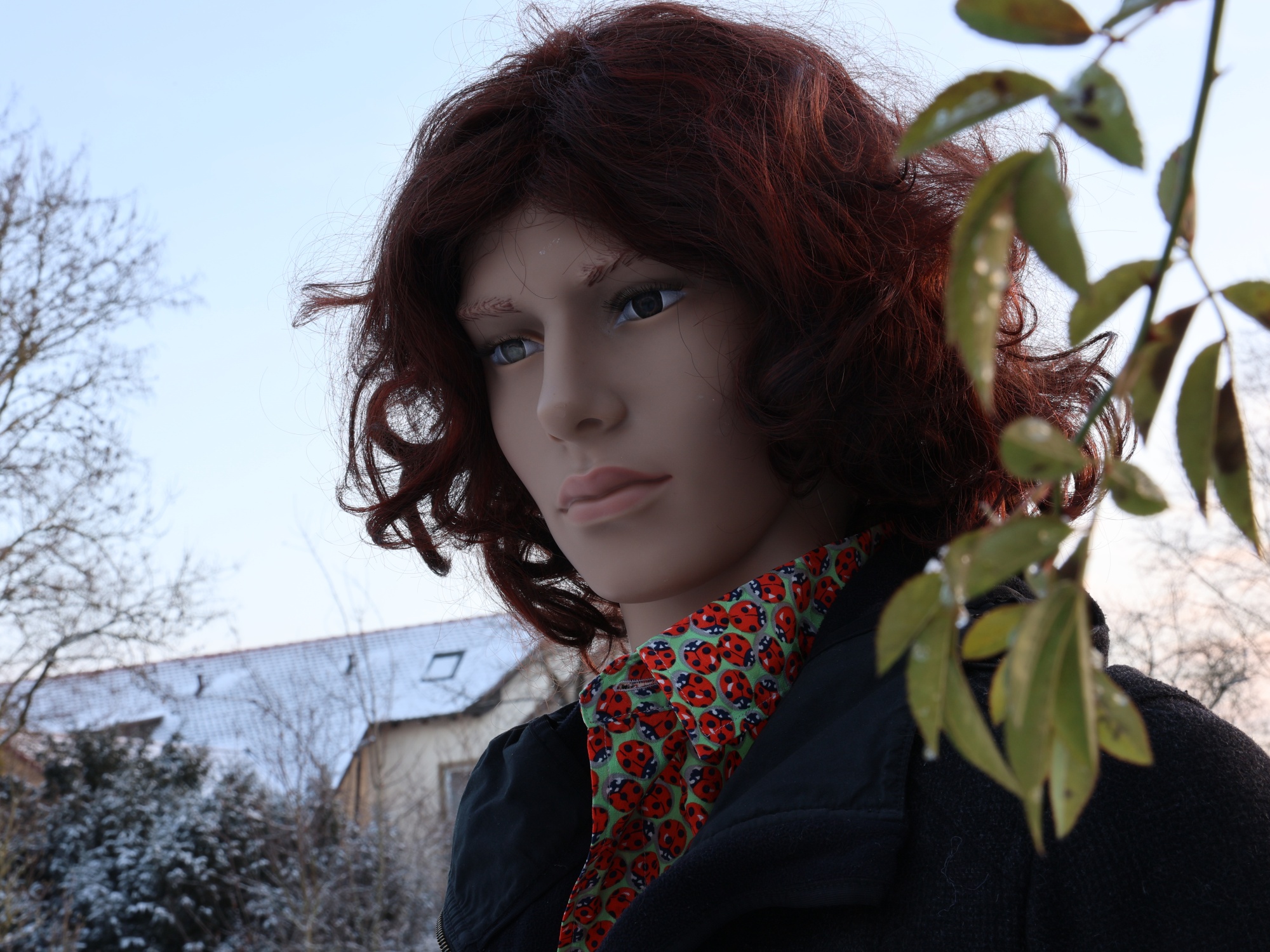}} &
        \includegraphics[width=\widthcompp\textwidth,valign=t, trim={1300px 1150px 360px 100px},clip]{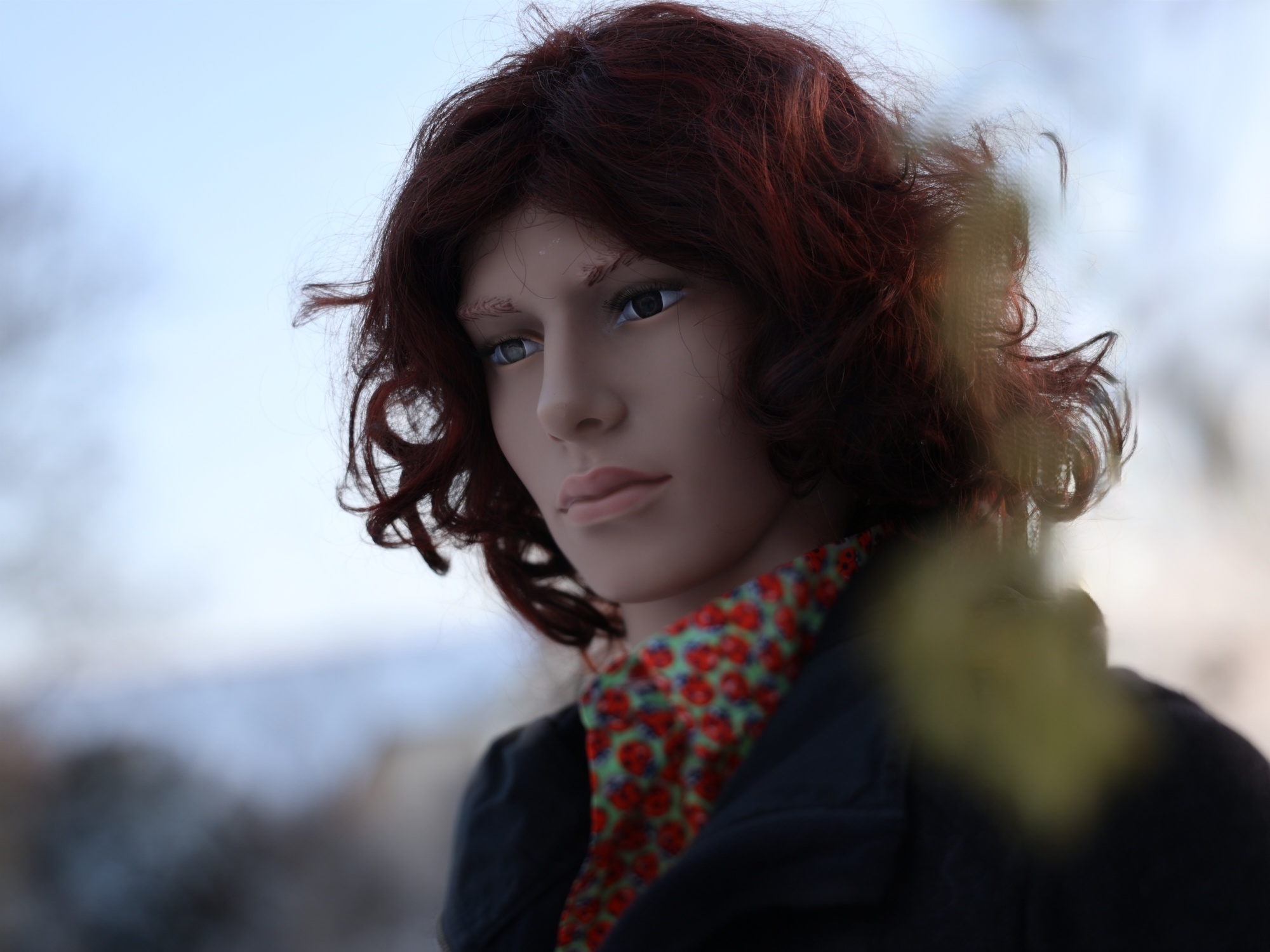} &
        \includegraphics[width=\widthcompp\textwidth,valign=t, trim={1300px 1150px 360px 100px},clip]{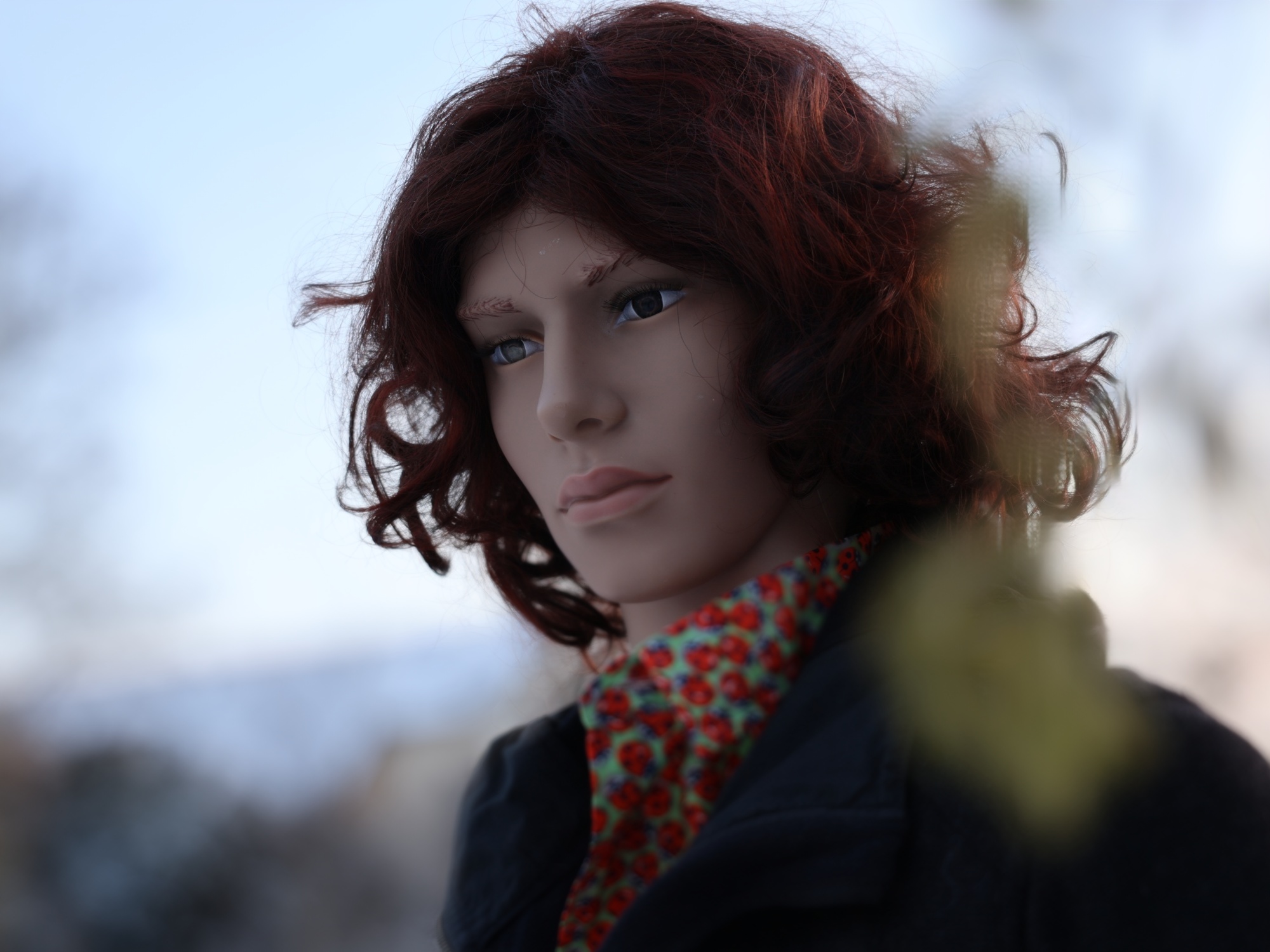} &
        \includegraphics[width=\widthcompp\textwidth,valign=t, trim={1300px 1150px 360px 100px},clip]{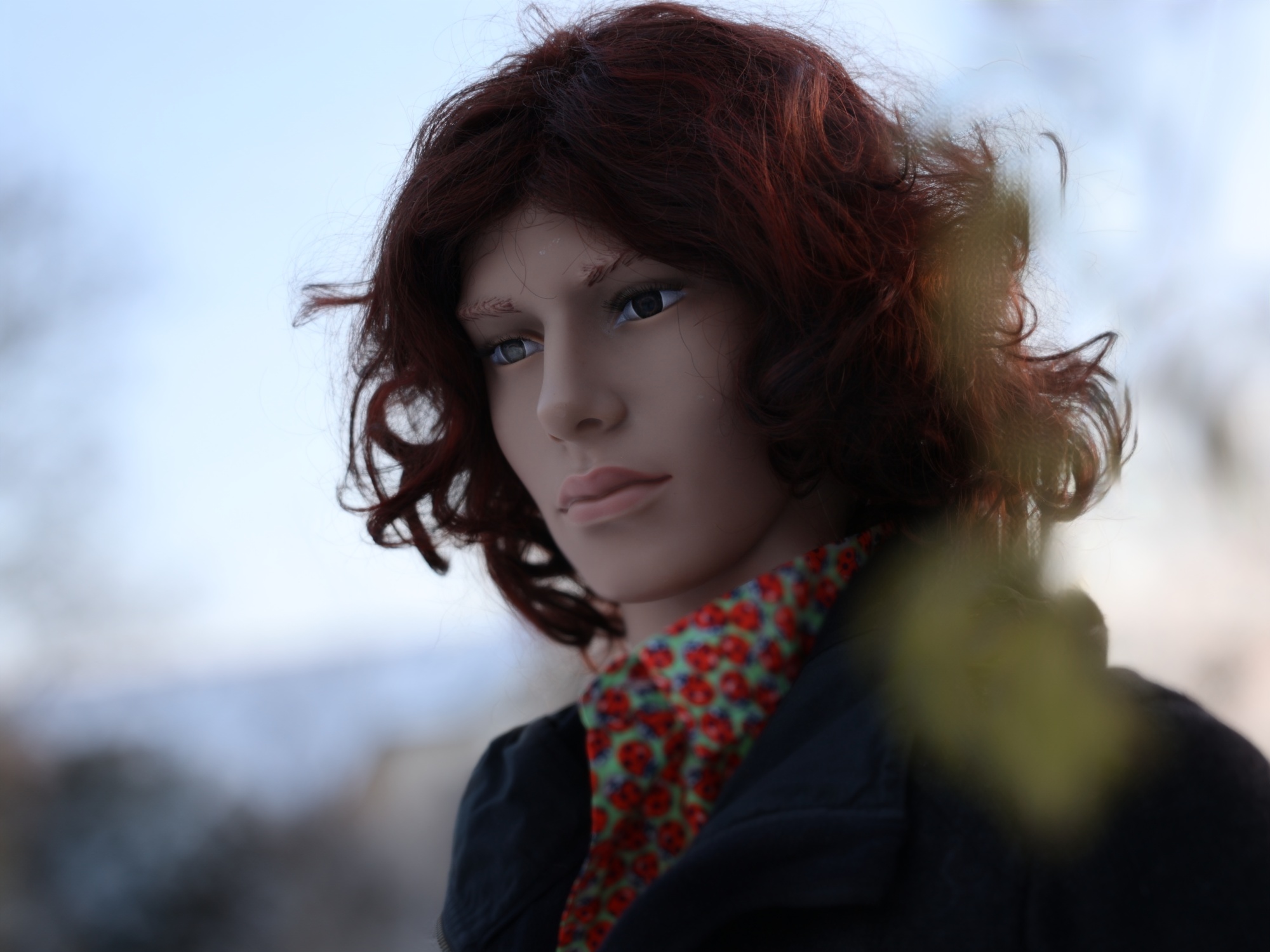} &
        \includegraphics[width=\widthcompp\textwidth,valign=t, trim={1300px 1150px 360px 100px},clip]{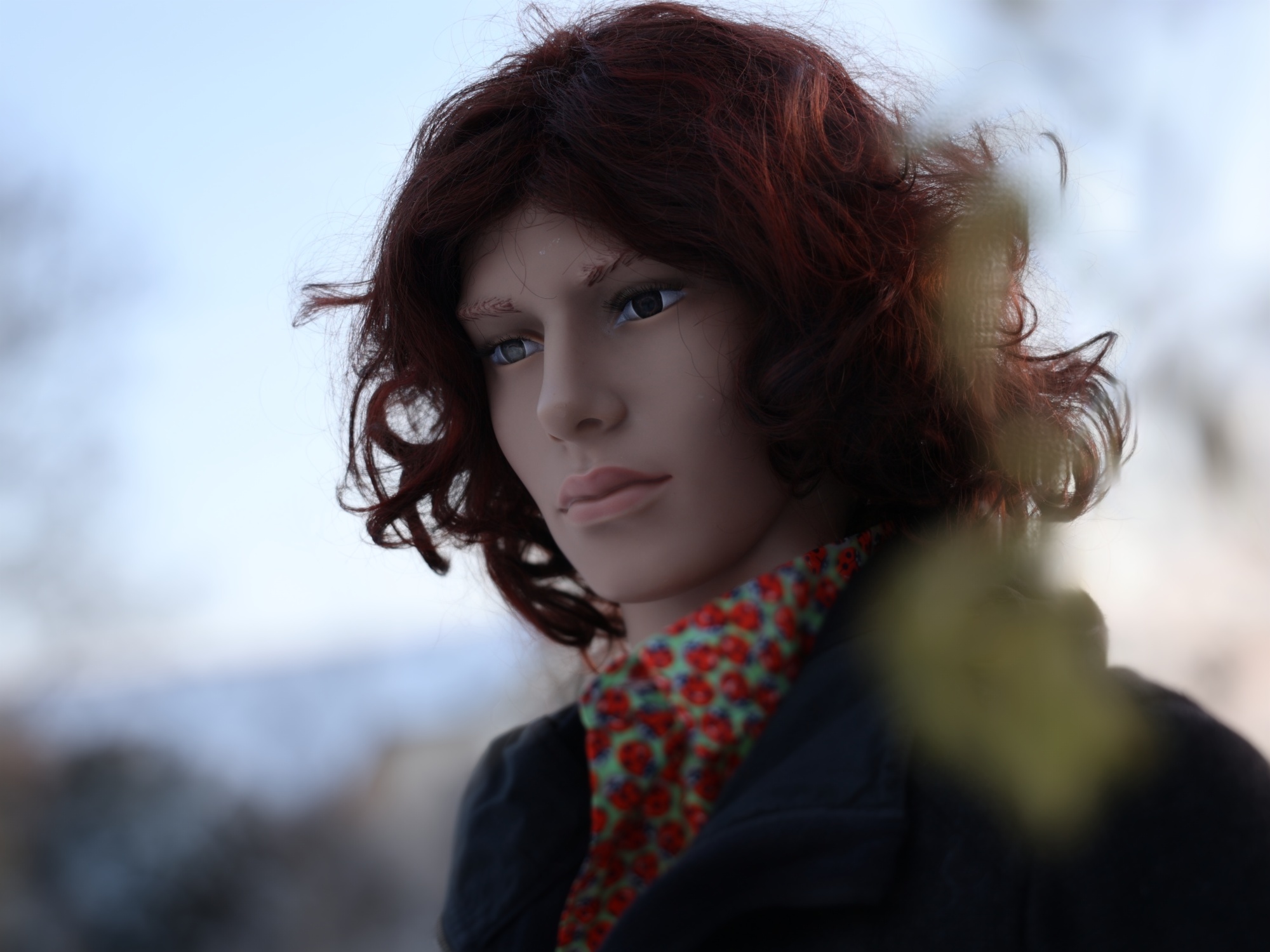} &
        \includegraphics[width=\widthcompp\textwidth,valign=t, trim={1300px 1150px 360px 100px},clip]{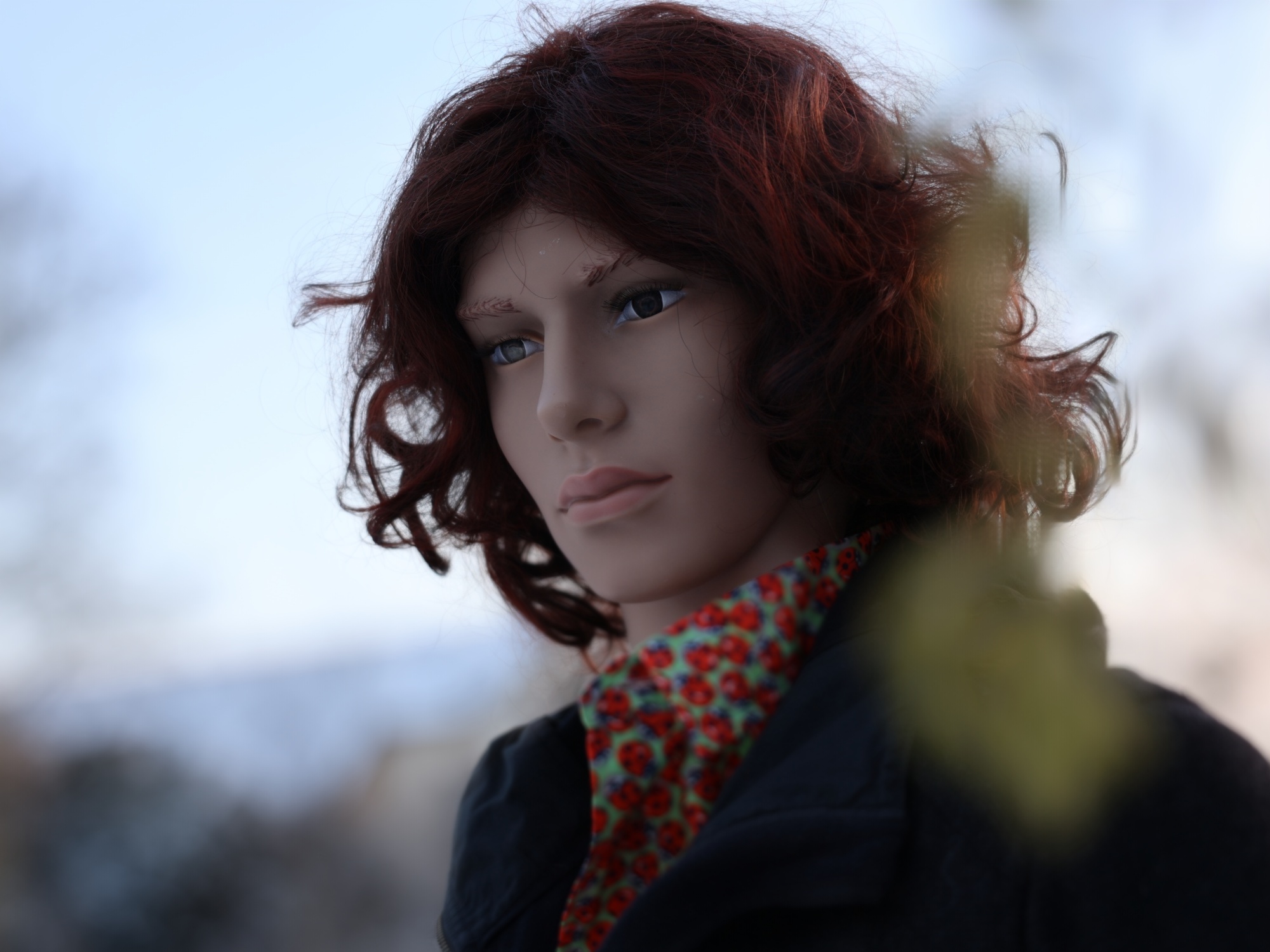} &
        \includegraphics[width=\widthcompp\textwidth,valign=t, trim={1300px 1150px 360px 100px},clip]{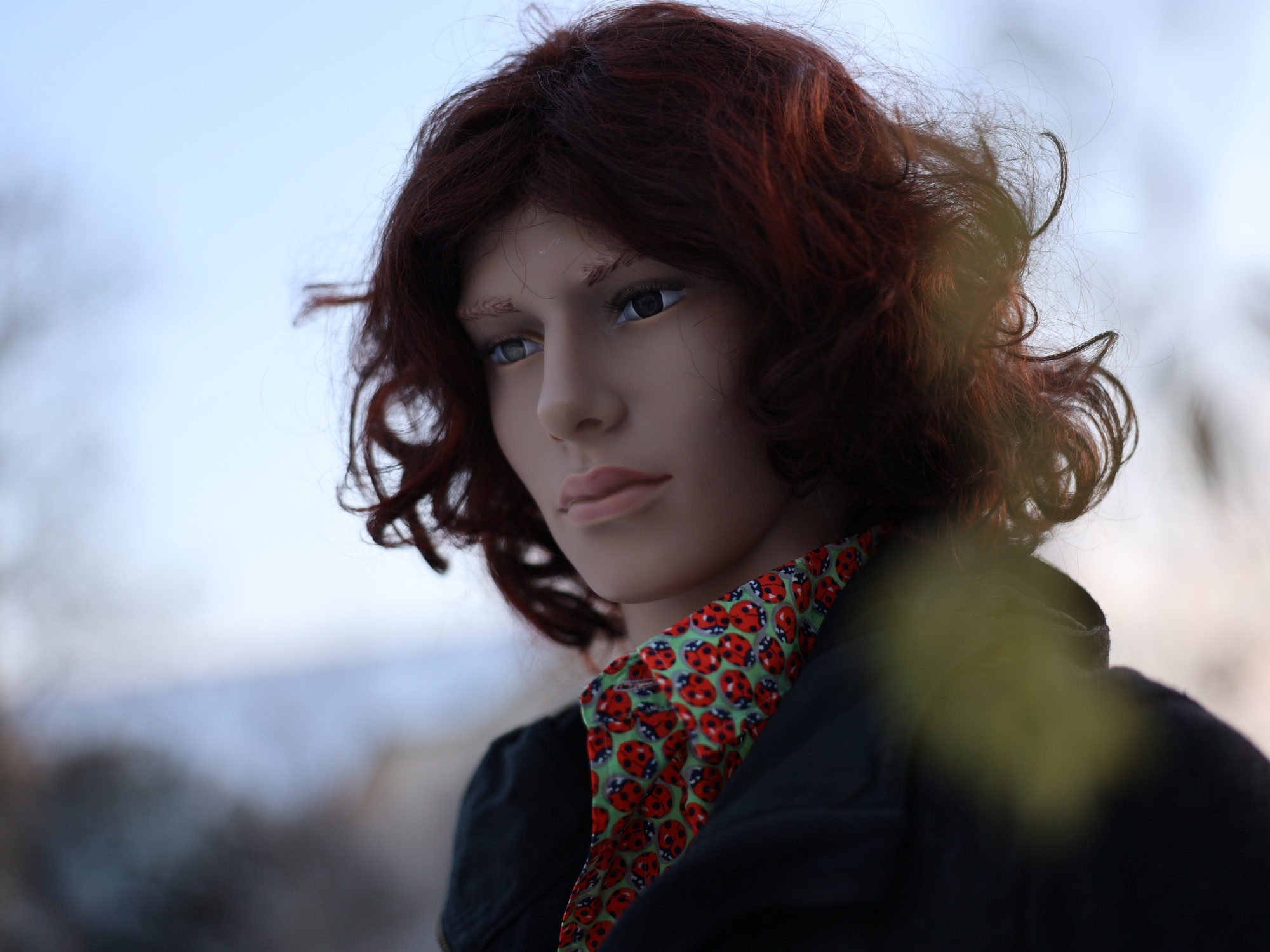} &
        \rotatebox{-90}{\hspace{3.4mm}\fnum{2.0}}
        \\
        \addlinespace[2.0pt]
        & 
        \includegraphics[width=\widthcompp\textwidth,valign=t, trim={1300px 1150px 360px 100px},clip]{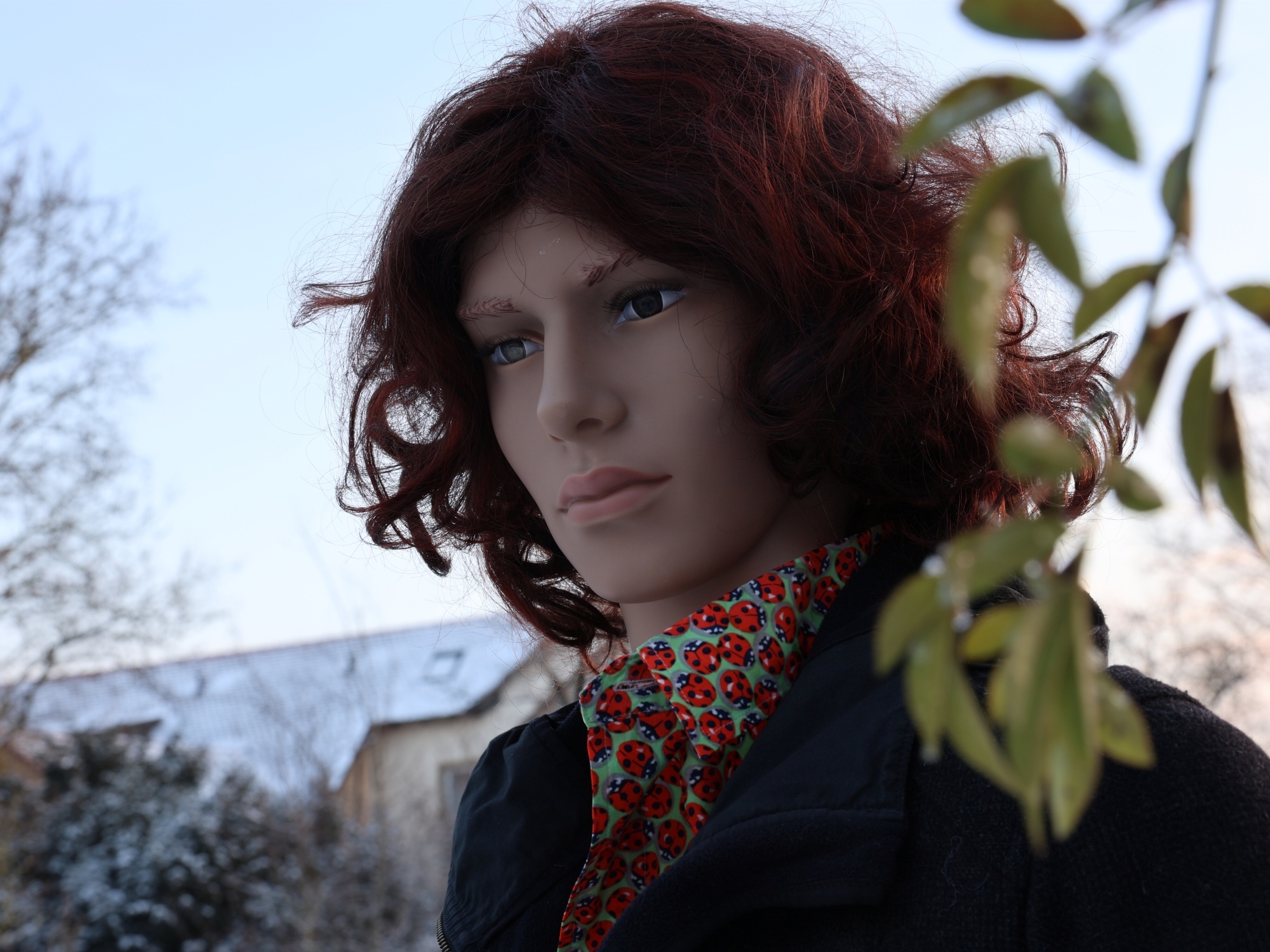} &
        \includegraphics[width=\widthcompp\textwidth,valign=t, trim={1300px 1150px 360px 100px},clip]{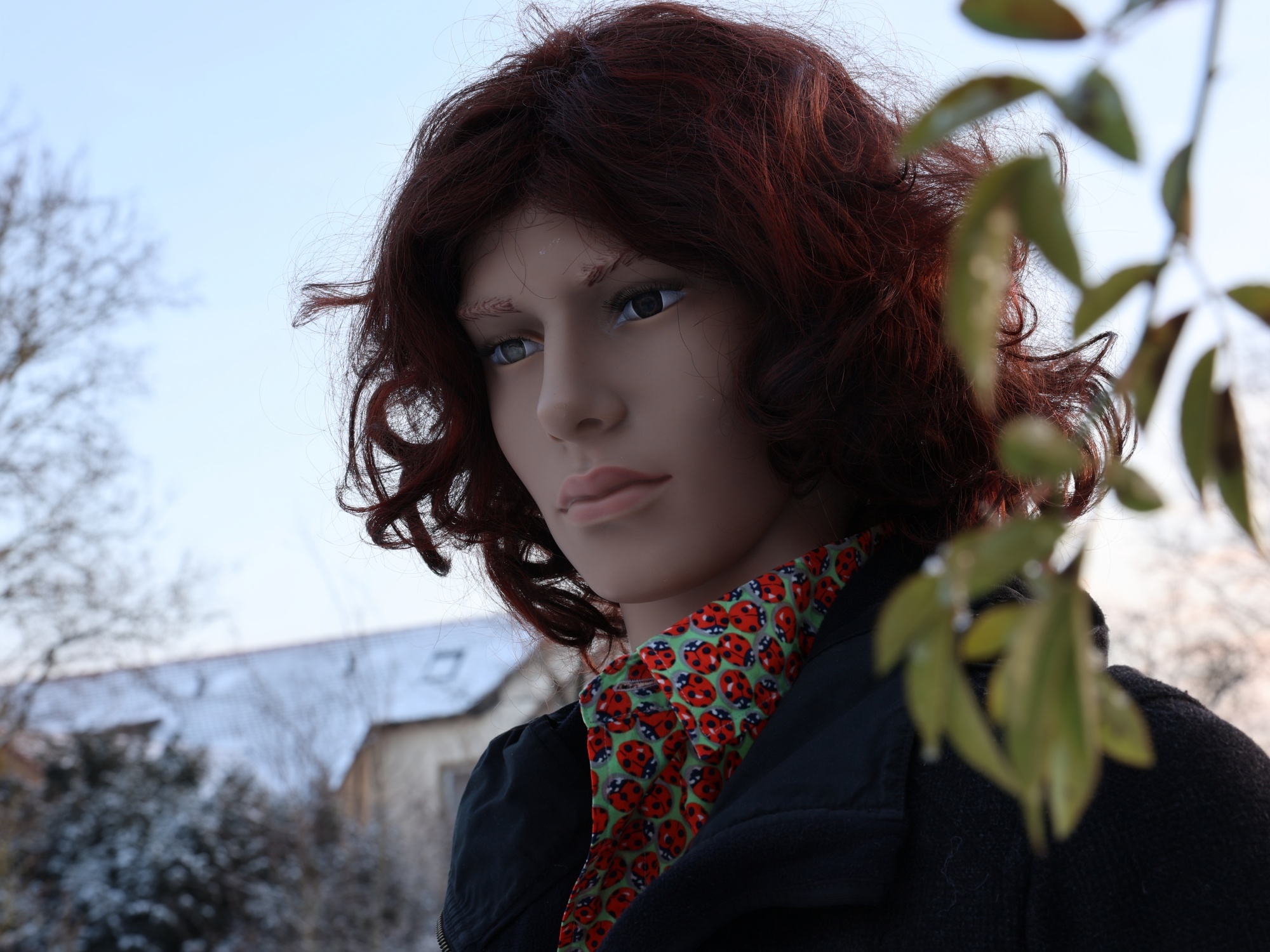} &
        \includegraphics[width=\widthcompp\textwidth,valign=t, trim={1300px 1150px 360px 100px},clip]{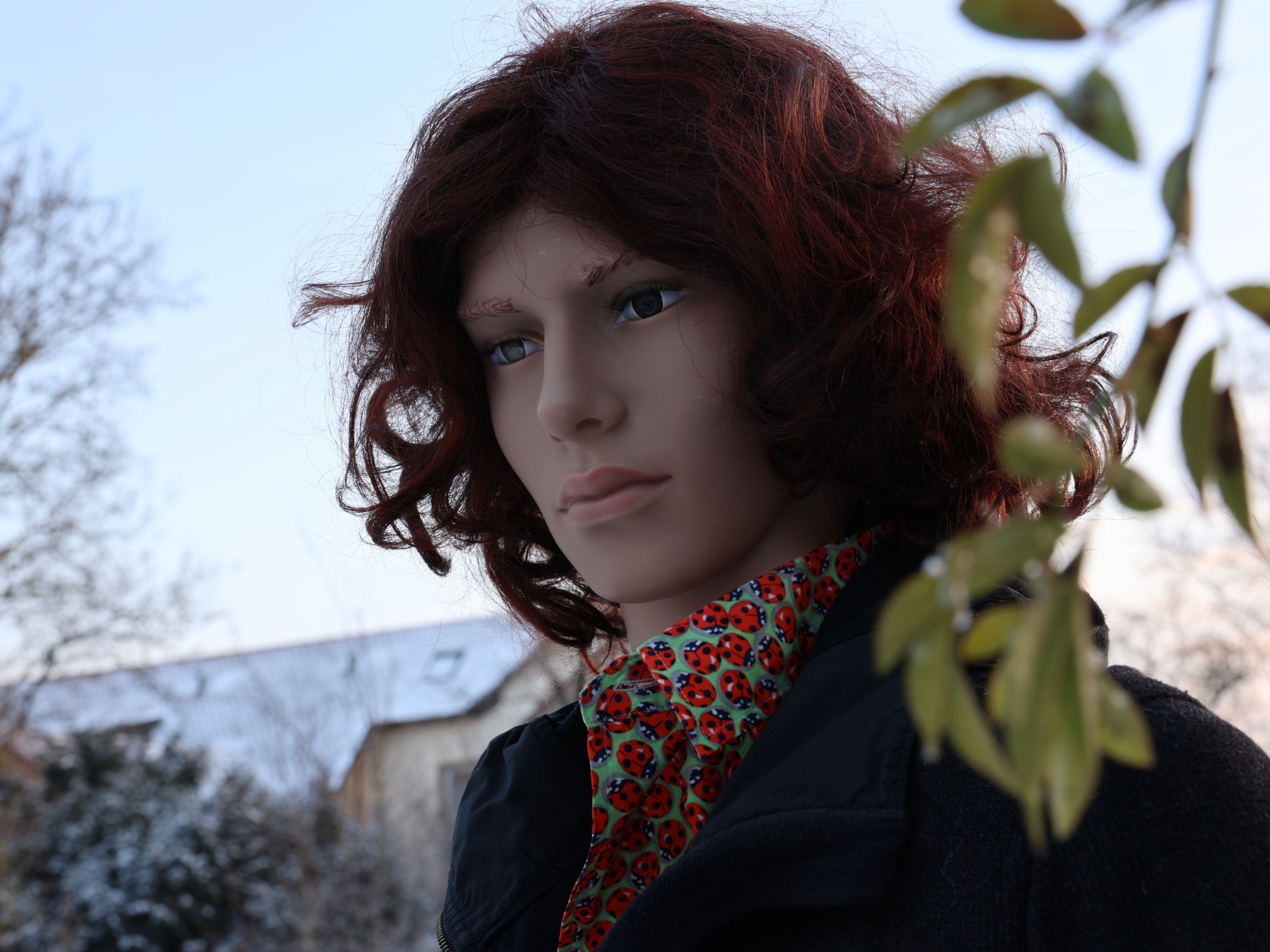} &
        \includegraphics[width=\widthcompp\textwidth,valign=t, trim={1300px 1150px 360px 100px},clip]{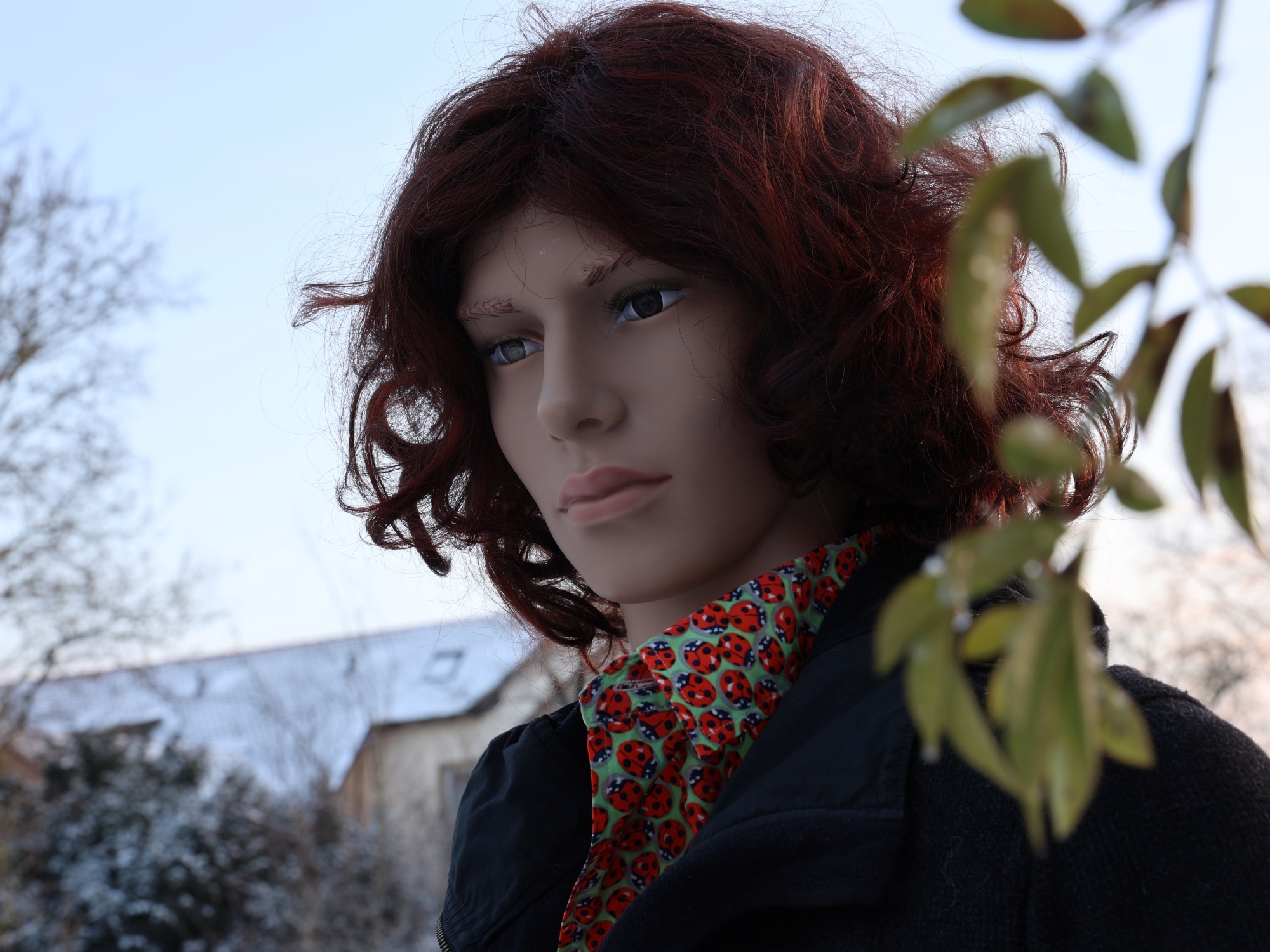} &
        \includegraphics[width=\widthcompp\textwidth,valign=t, trim={1300px 1150px 360px 100px},clip]{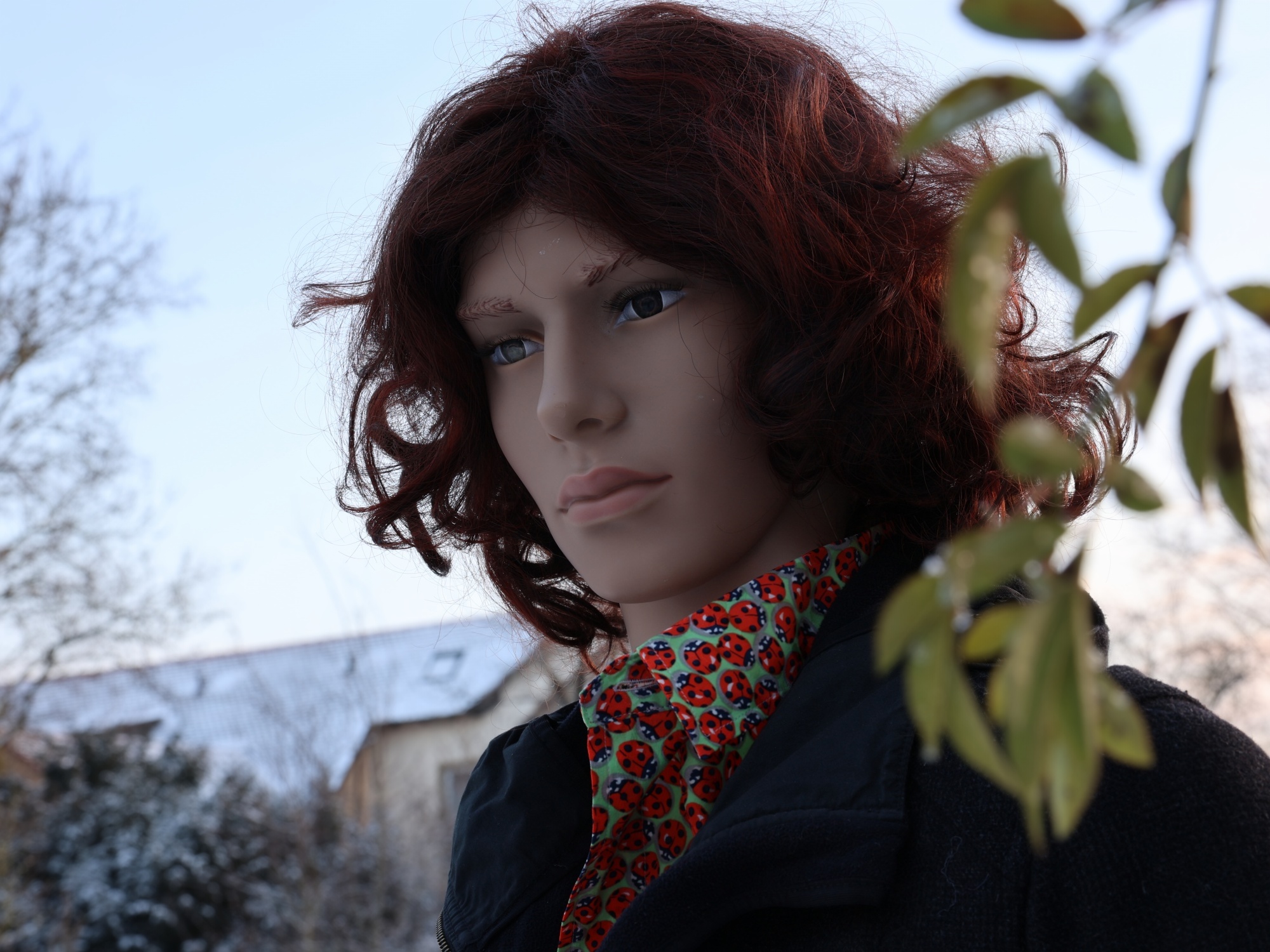} &
        \includegraphics[width=\widthcompp\textwidth,valign=t, trim={1300px 1150px 360px 100px},clip]{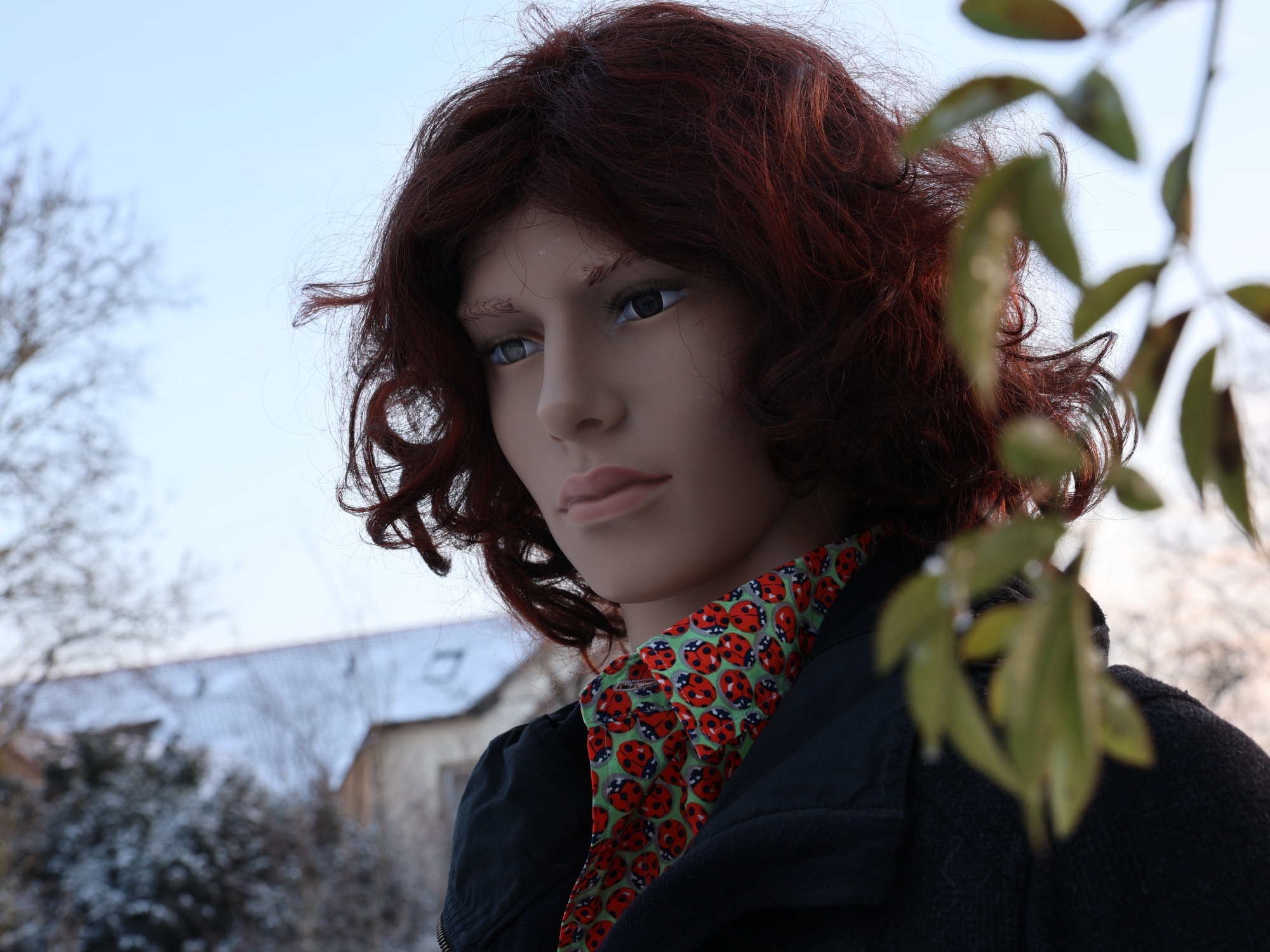} &
        \rotatebox{-90}{\hspace{2.8mm}\fnum{11.0}}
    \\
    \addlinespace[5.0pt]
    \multirow{3}{*}[1.4mm]{\includegraphics[height=\imgcompp\textwidth, trim={100px 0px 450px 0px},clip]{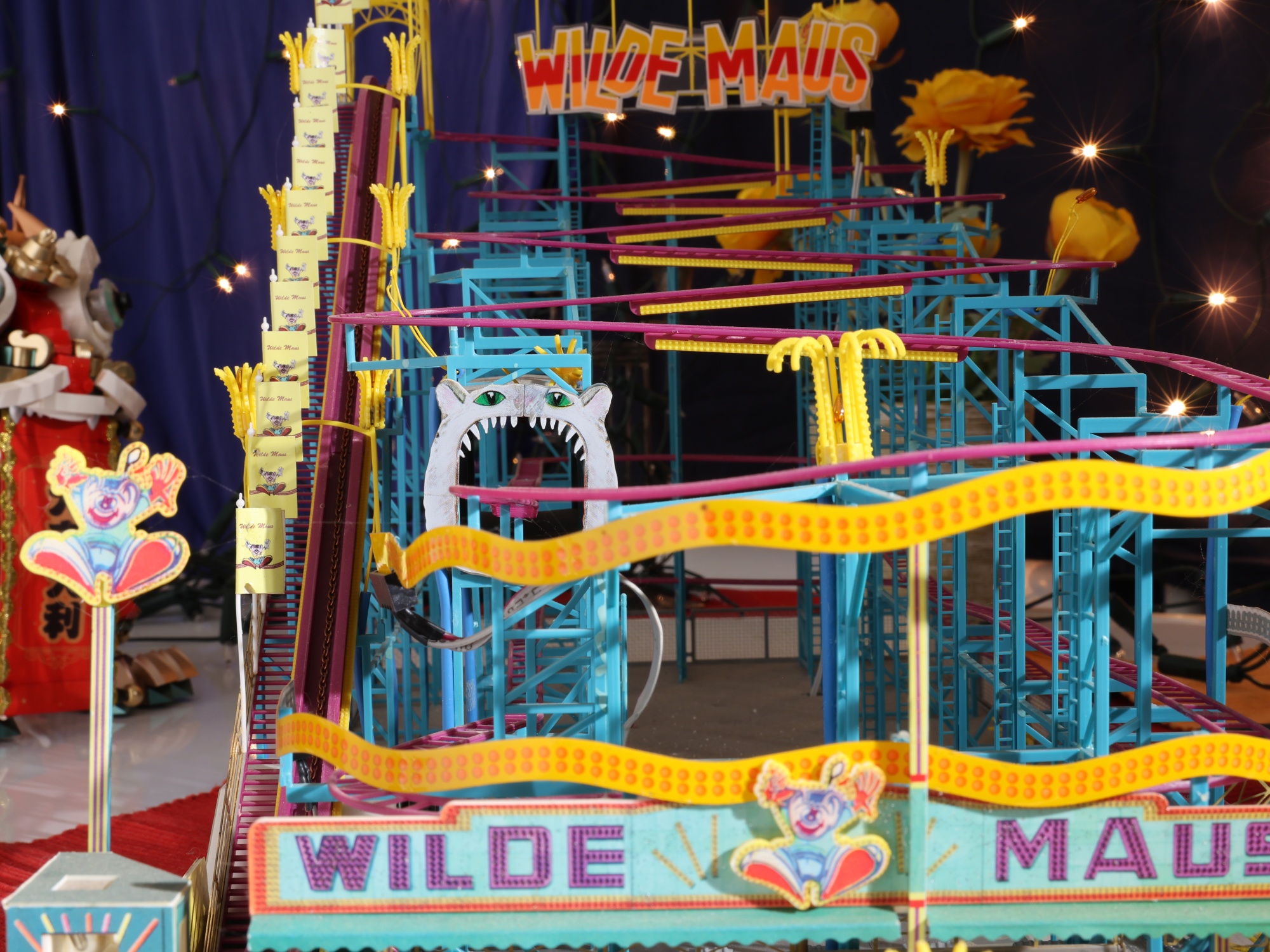}} &
        \includegraphics[width=\widthcompp\textwidth,valign=t, trim={700px 1150px 960px 100px},clip]{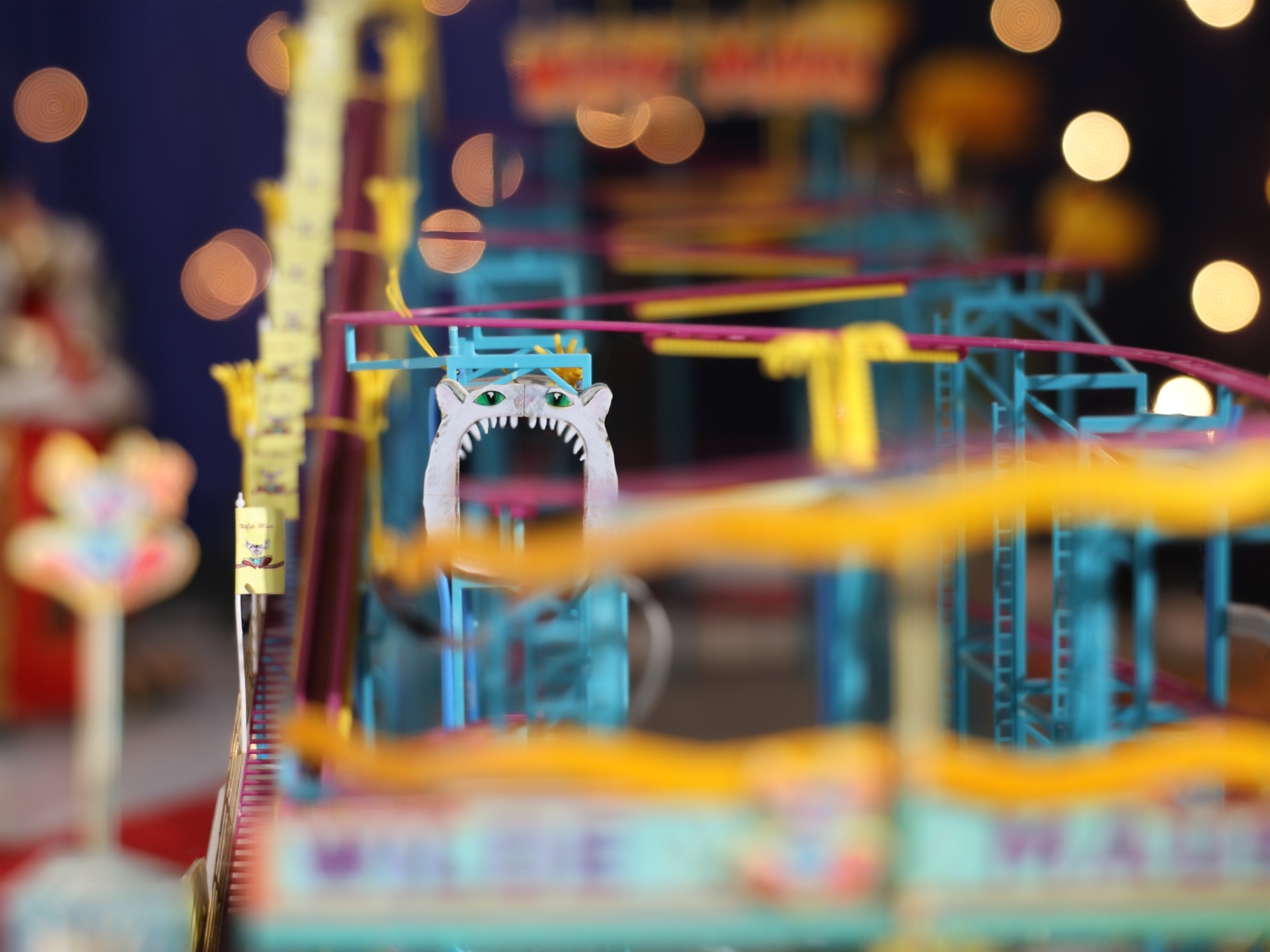} &
        \includegraphics[width=\widthcompp\textwidth,valign=t, trim={700px 1150px 960px 100px},clip]{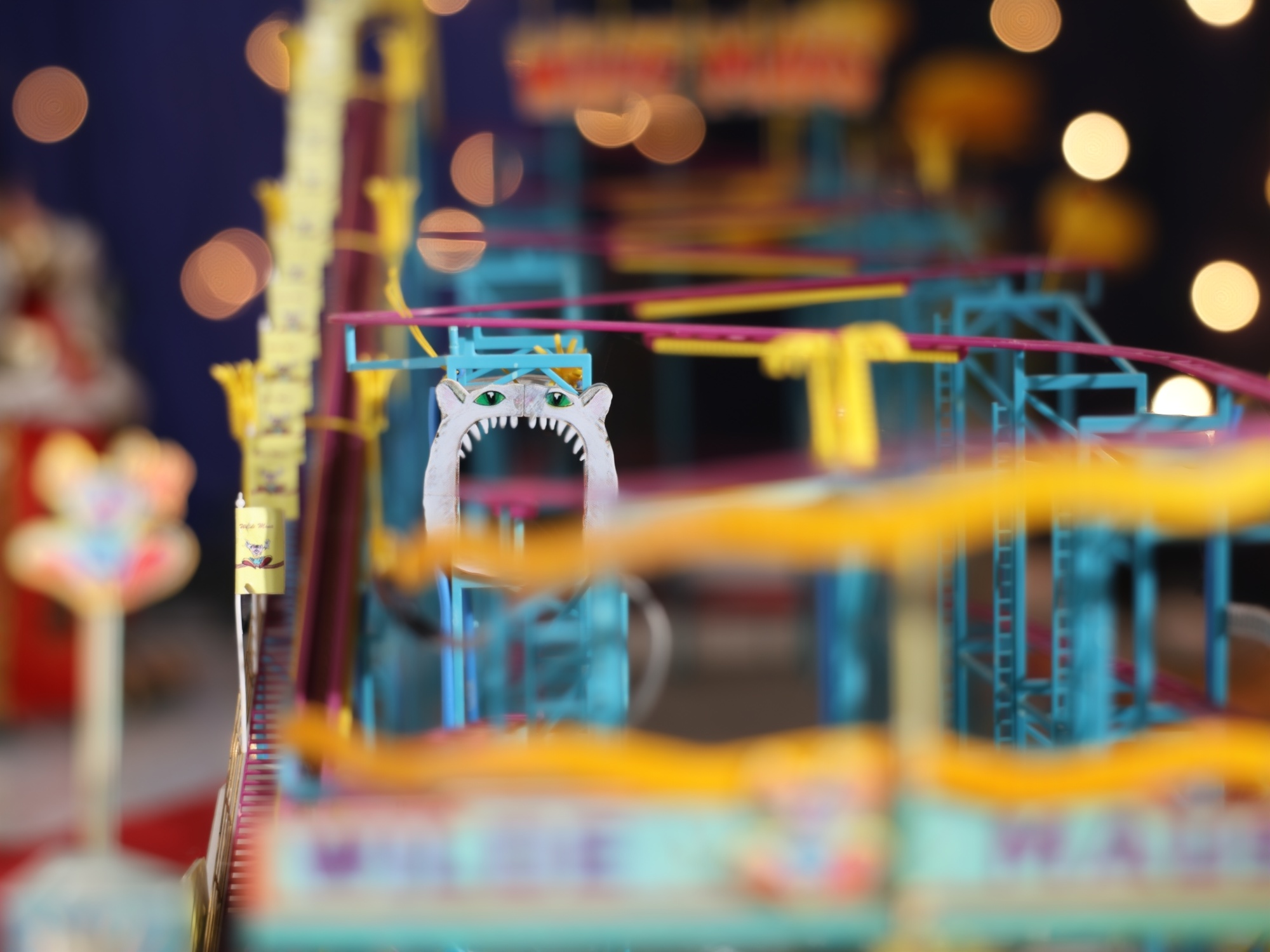} &
        \includegraphics[width=\widthcompp\textwidth,valign=t, trim={700px 1150px 960px 100px},clip]{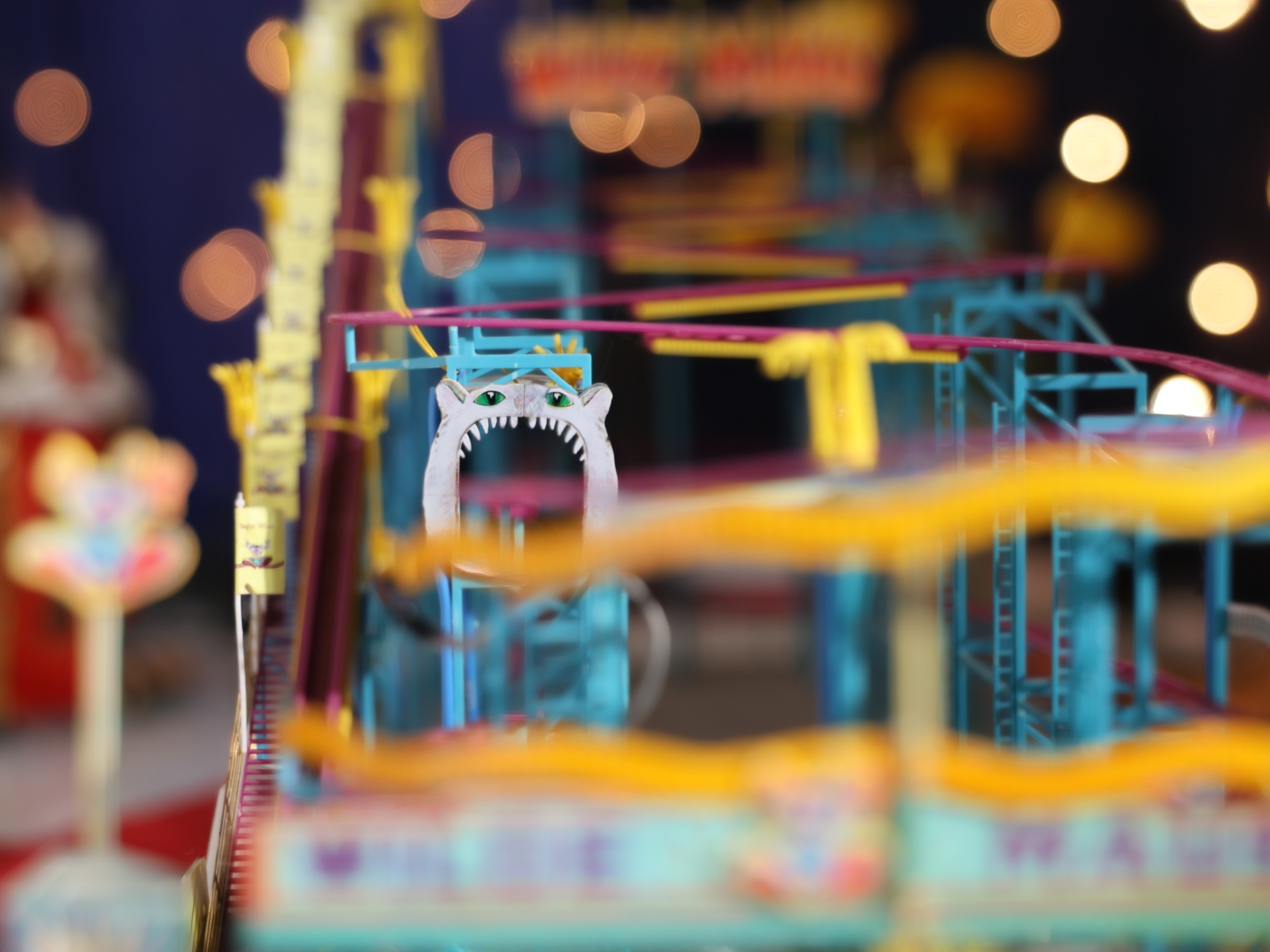} &
        \includegraphics[width=\widthcompp\textwidth,valign=t, trim={700px 1150px 960px 100px},clip]{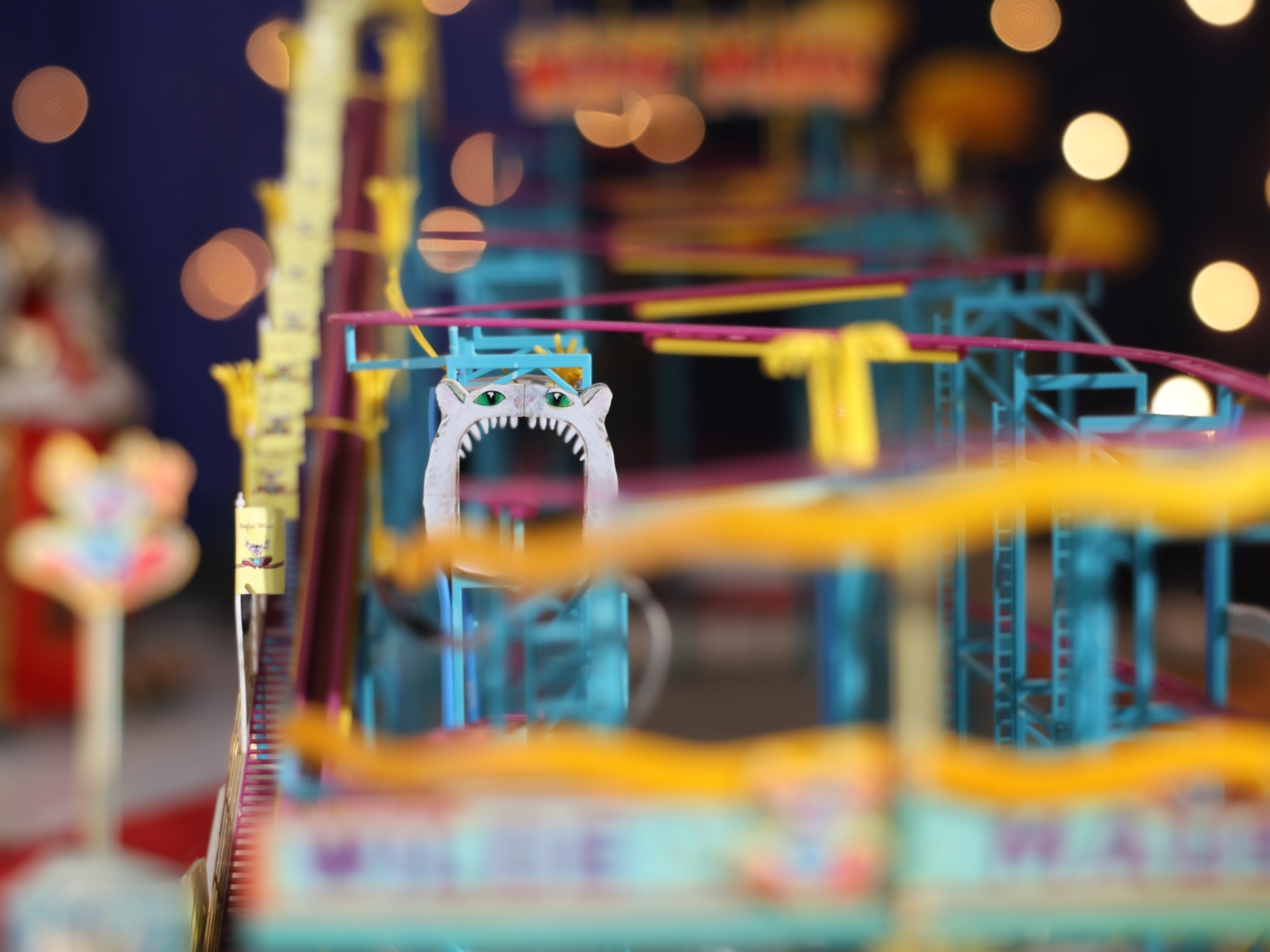} &
        \includegraphics[width=\widthcompp\textwidth,valign=t, trim={700px 1150px 960px 100px},clip]{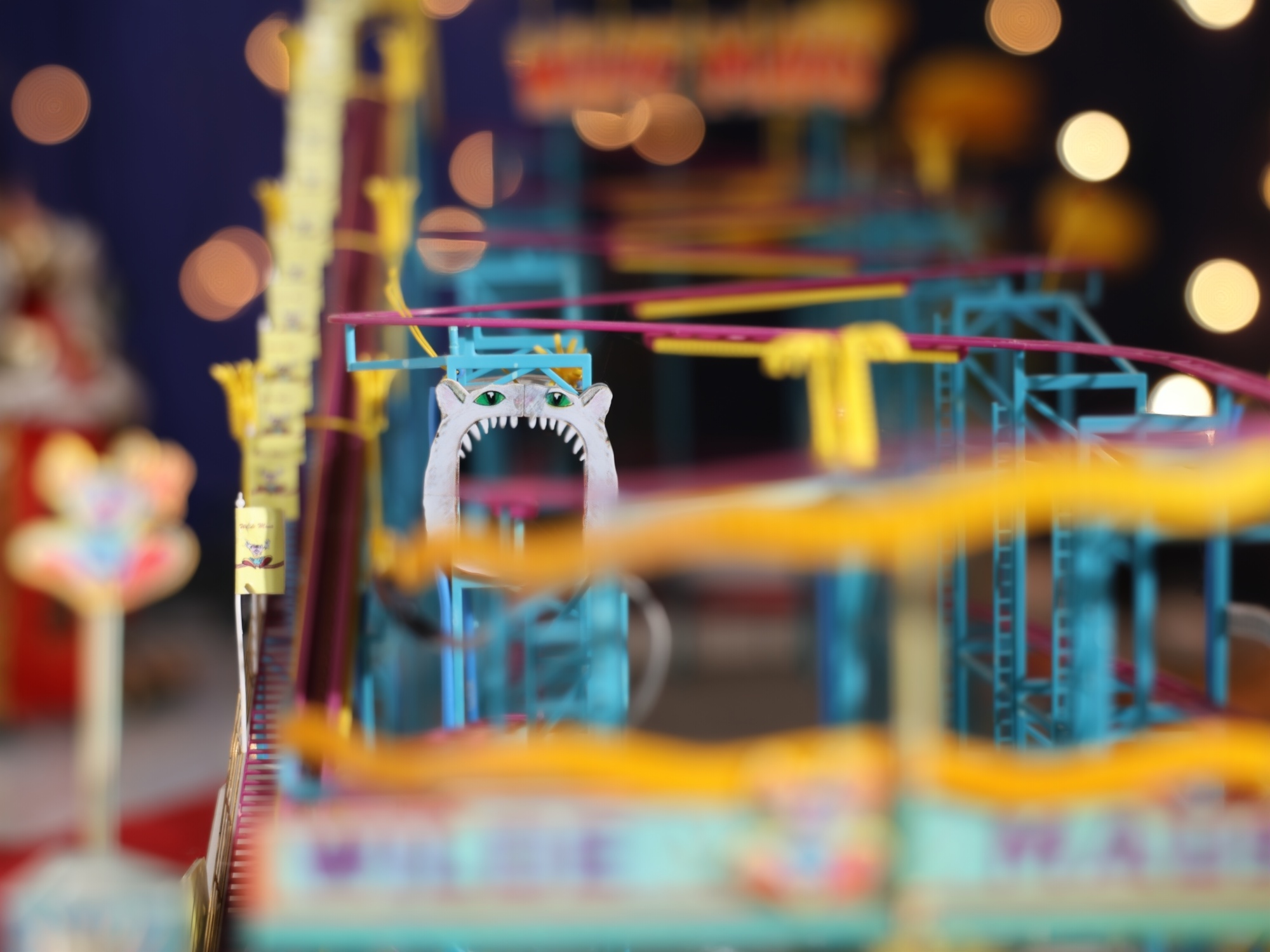} &
        \includegraphics[width=\widthcompp\textwidth,valign=t, trim={700px 1150px 960px 100px},clip]{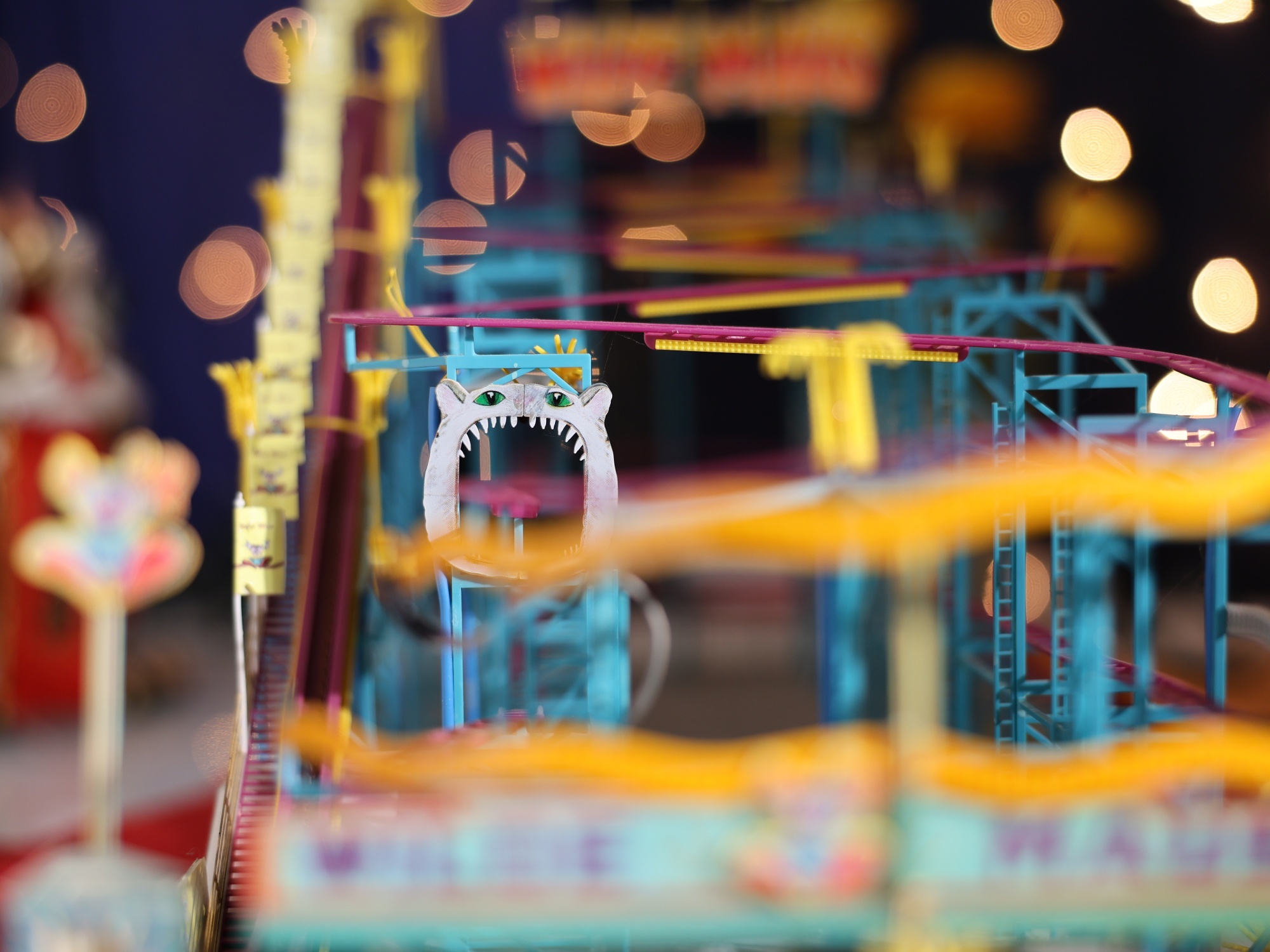} &
        \rotatebox{-90}{\hspace{3.4mm}\fnum{2.0}}
        \\
        \addlinespace[2.0pt]
        & 
        \includegraphics[width=\widthcompp\textwidth,valign=t, trim={700px 1150px 960px 100px},clip]{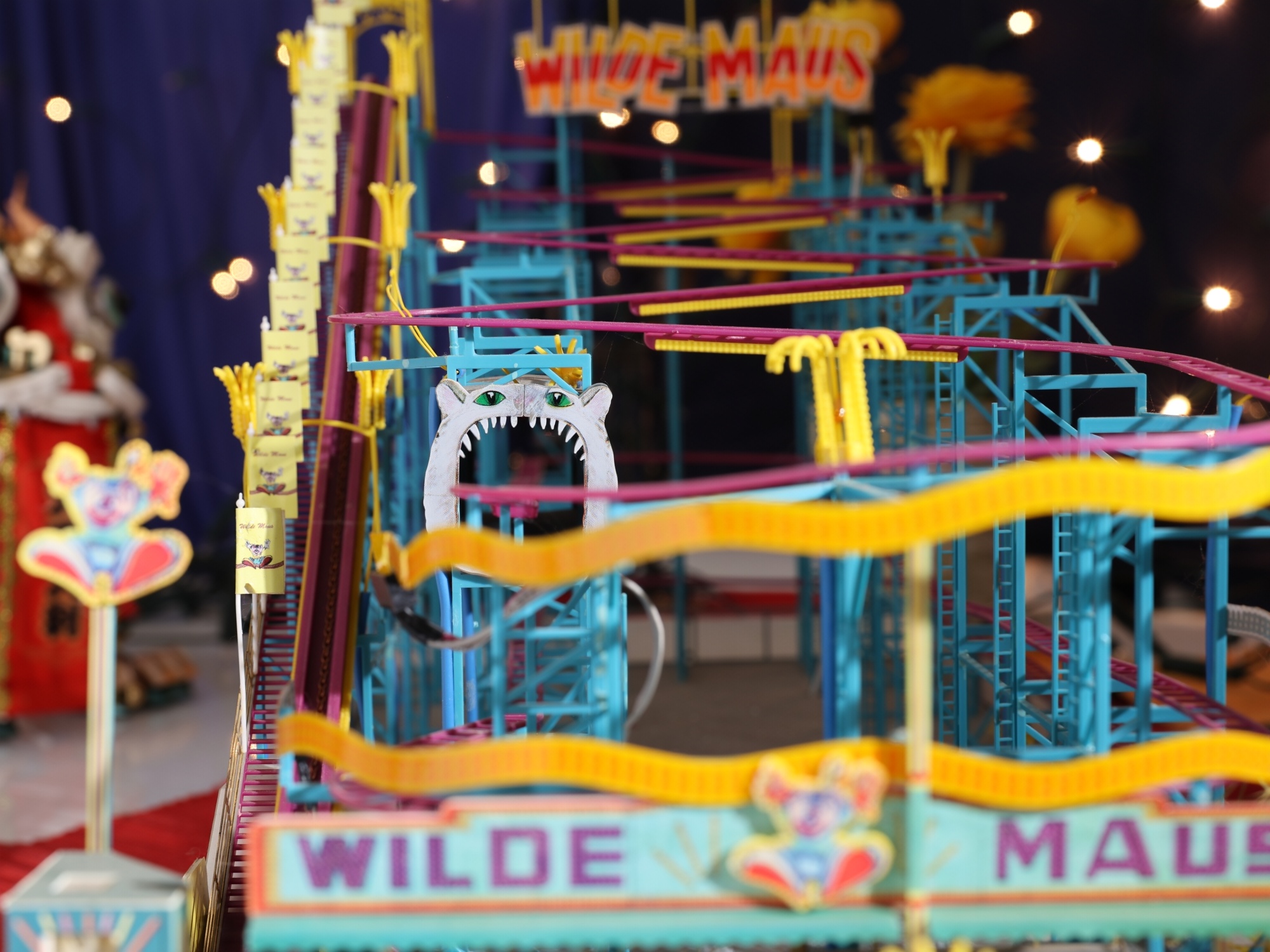} &
        \includegraphics[width=\widthcompp\textwidth,valign=t, trim={700px 1150px 960px 100px},clip]{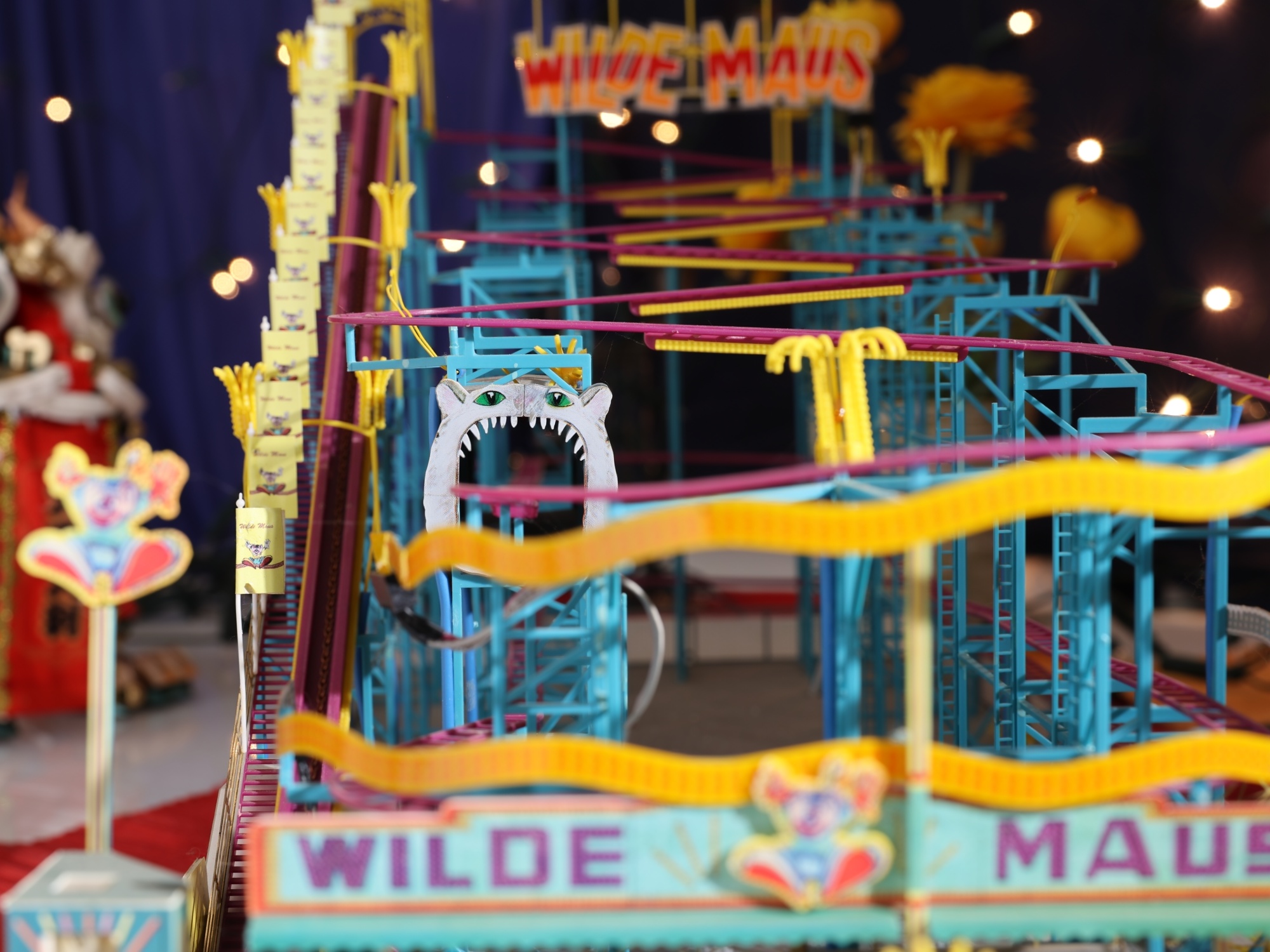} &
        \includegraphics[width=\widthcompp\textwidth,valign=t, trim={700px 1150px 960px 100px},clip]{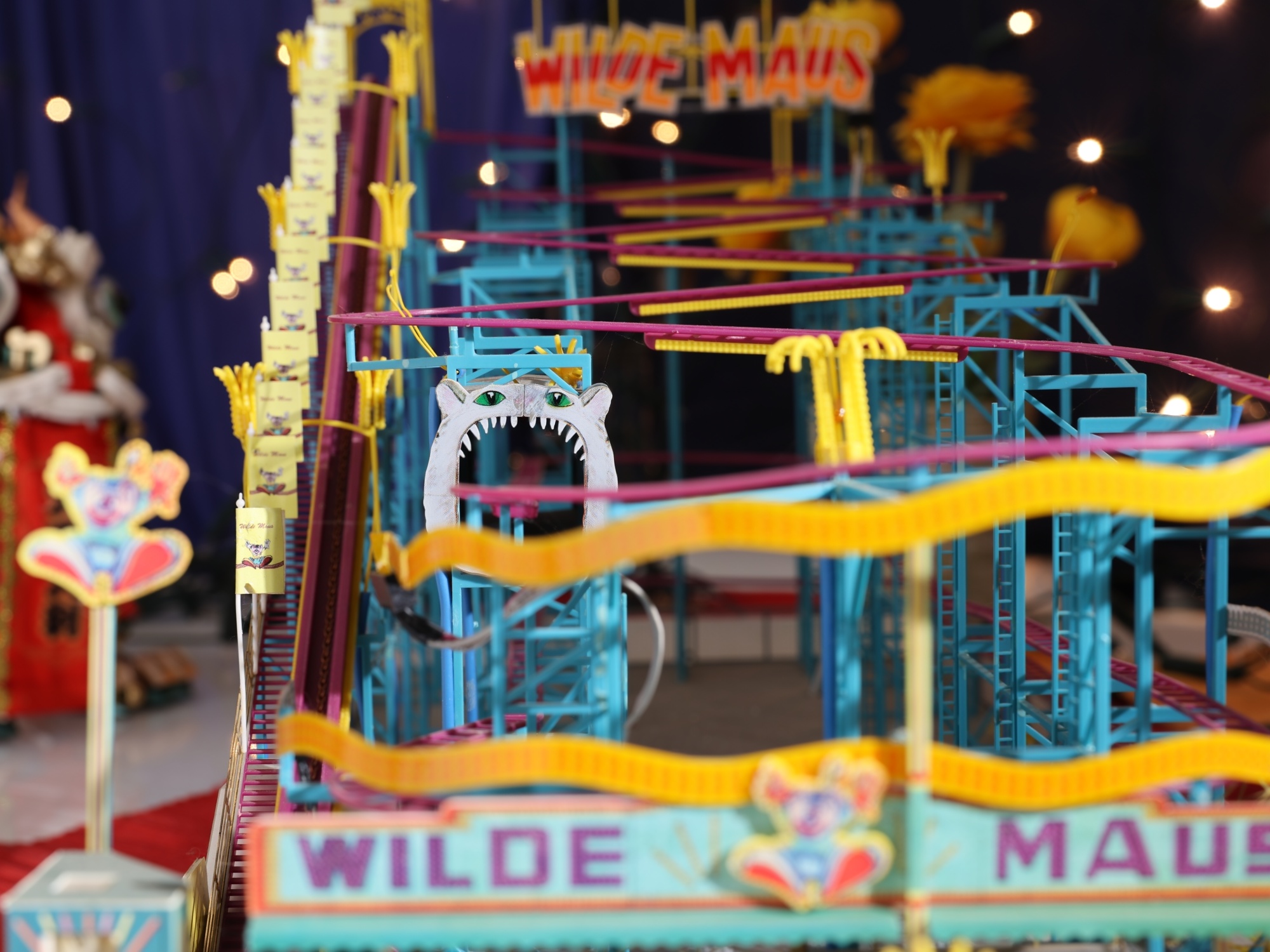} &
        \includegraphics[width=\widthcompp\textwidth,valign=t, trim={700px 1150px 960px 100px},clip]{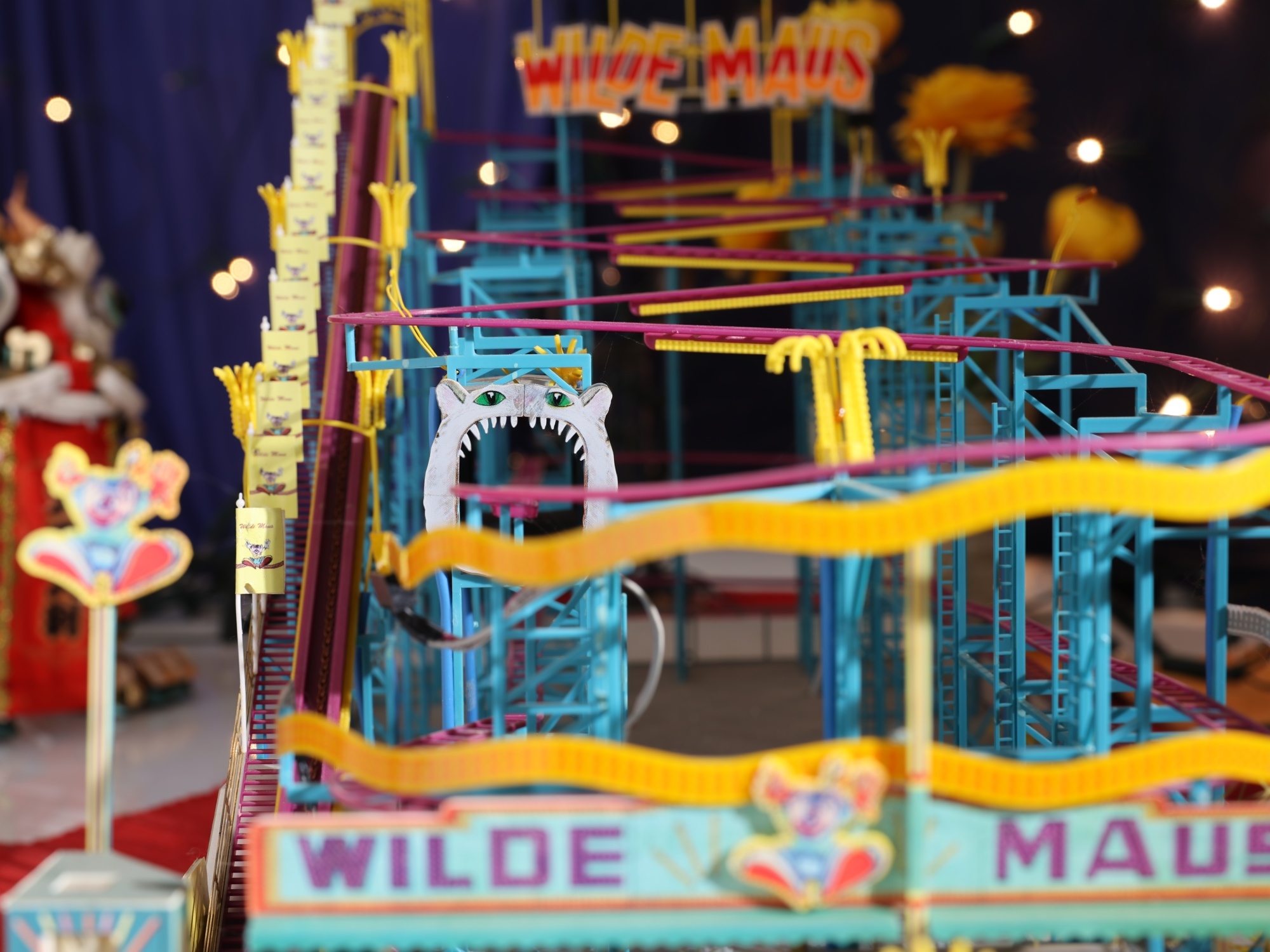} &
        \includegraphics[width=\widthcompp\textwidth,valign=t, trim={700px 1150px 960px 100px},clip]{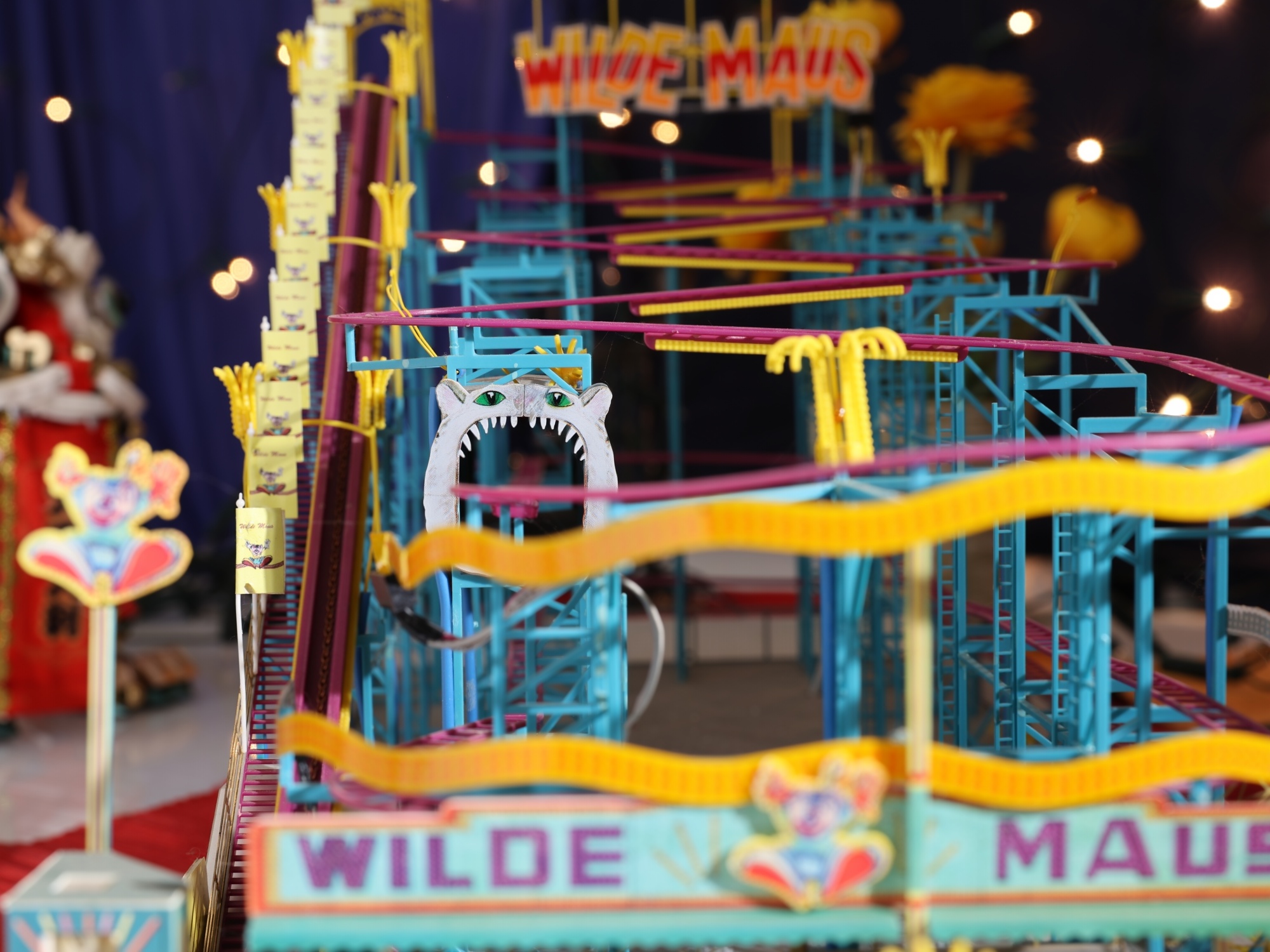} &
        \includegraphics[width=\widthcompp\textwidth,valign=t, trim={700px 1150px 960px 100px},clip]{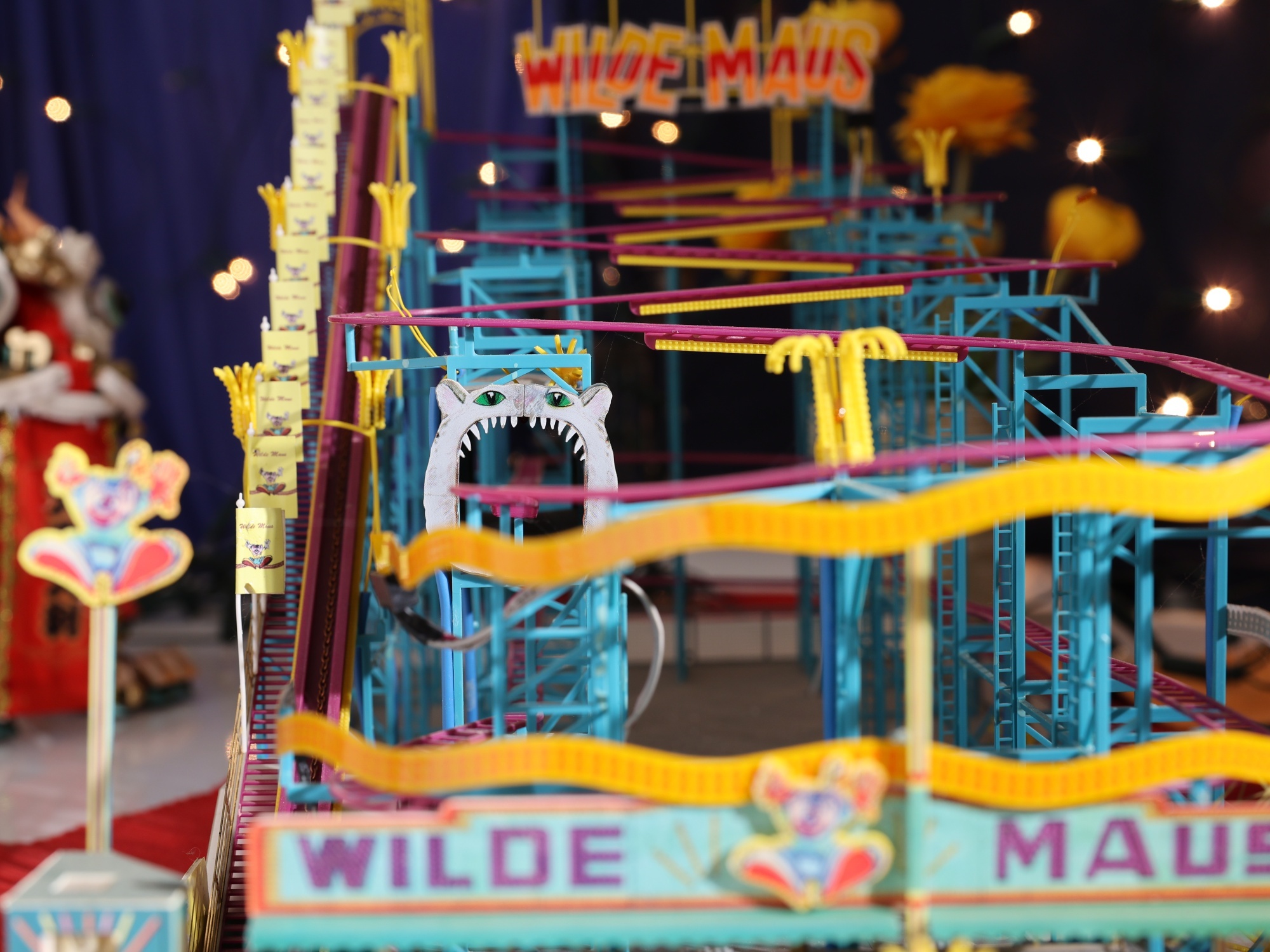} &
        \rotatebox{-90}{\hspace{3.4mm}\fnum{7.1}}
    \\
    \addlinespace[5.0pt] 
    \multirow{3}{*}[1.4mm]{\includegraphics[height=\imgcompp\textwidth, trim={350px 0px 200px 0px},clip]{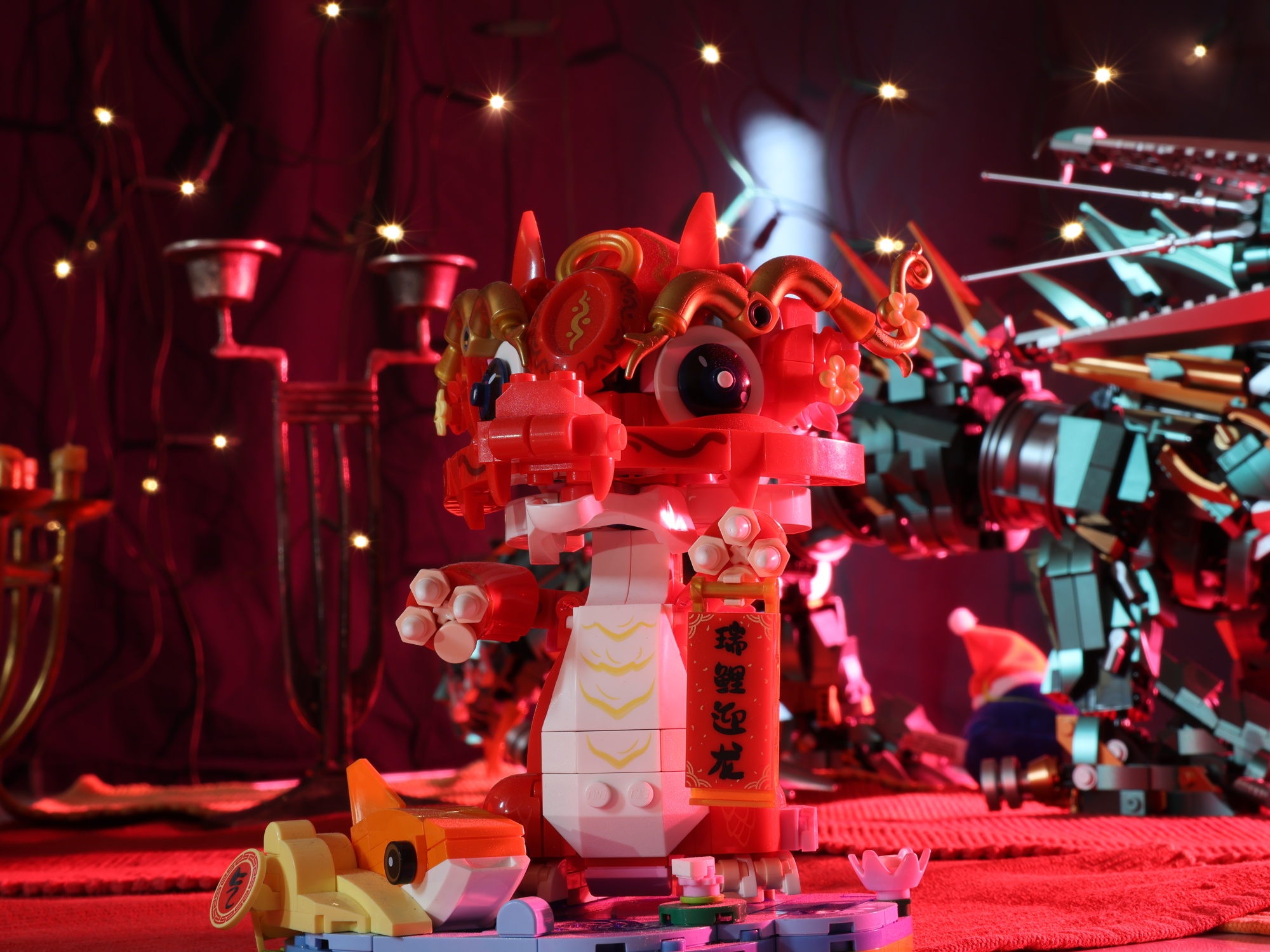}} &
        \includegraphics[width=\widthcompp\textwidth,valign=t, trim={1200px 1000px 460px 250px},clip]{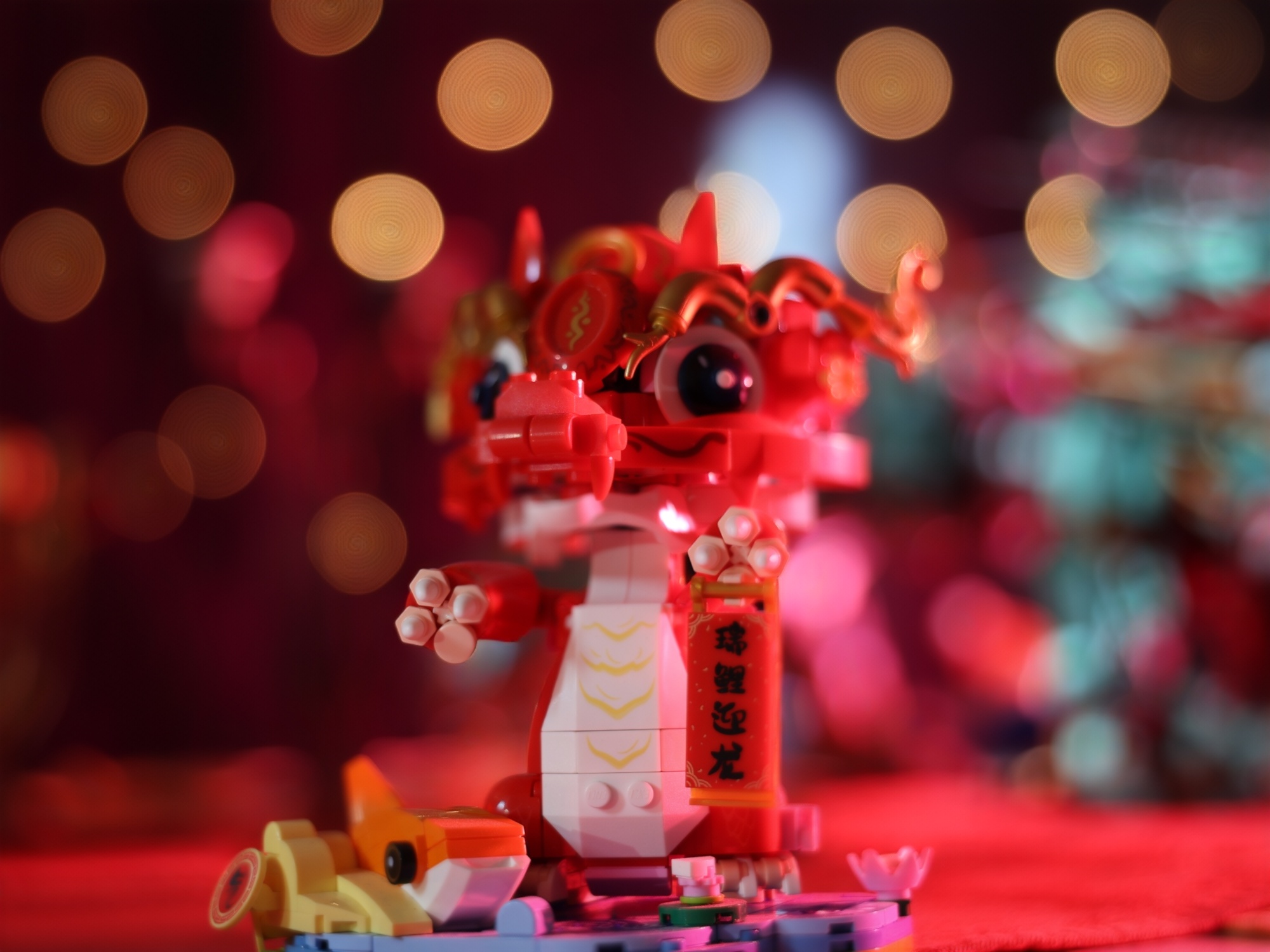} &
        \includegraphics[width=\widthcompp\textwidth,valign=t, trim={1200px 1000px 460px 250px},clip]{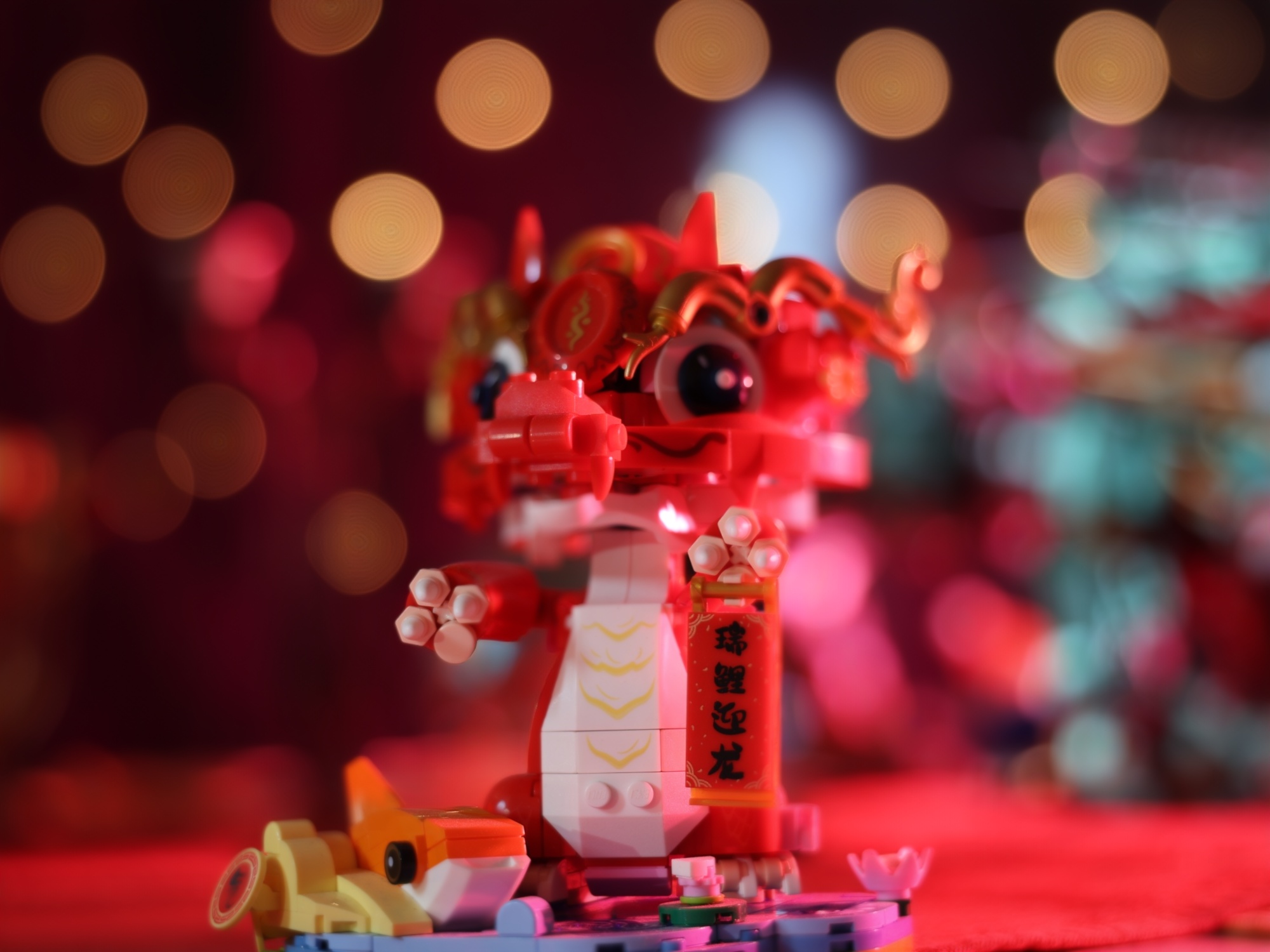} &
        \includegraphics[width=\widthcompp\textwidth,valign=t, trim={1200px 1000px 460px 250px},clip]{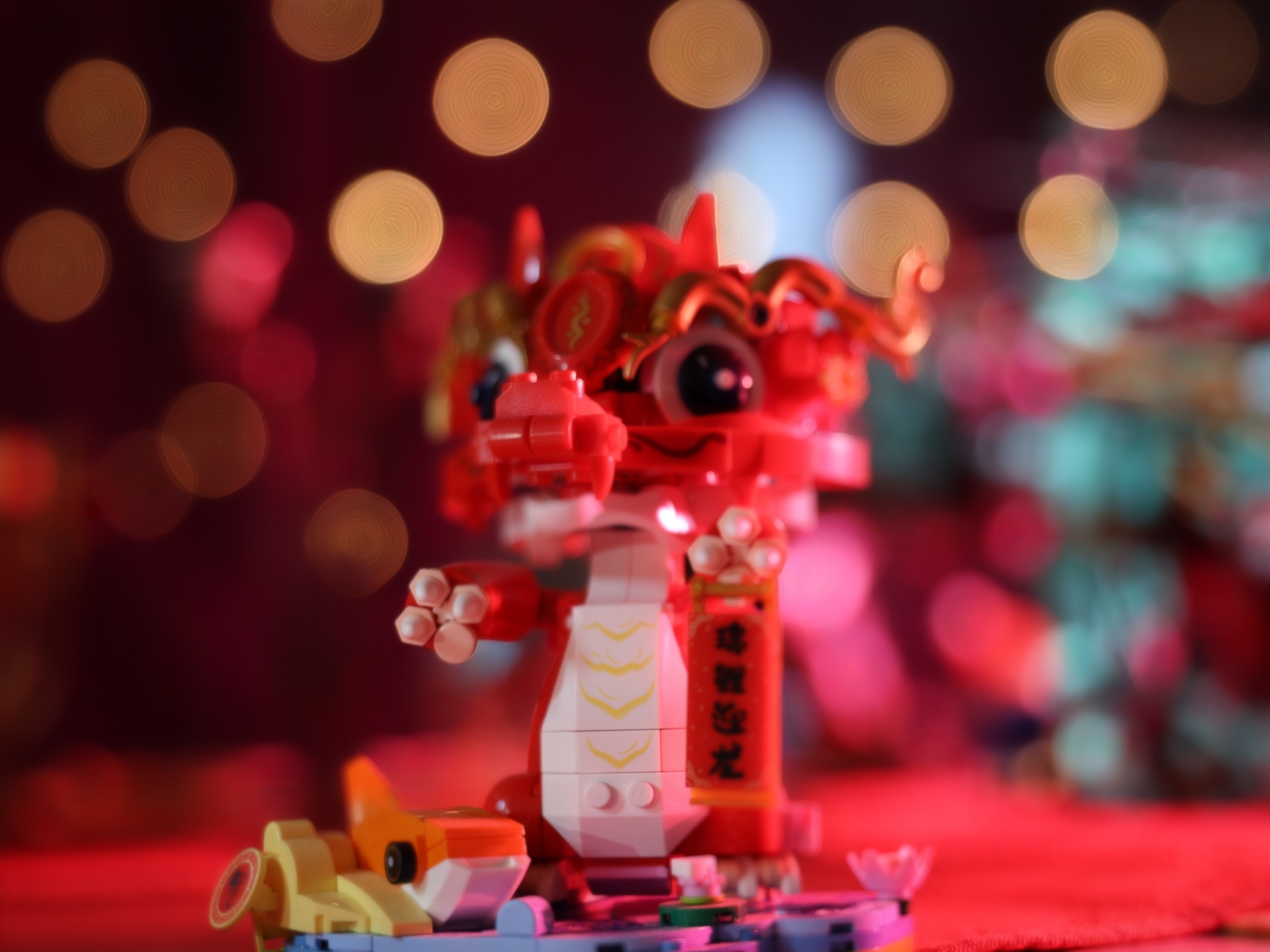} &
        \includegraphics[width=\widthcompp\textwidth,valign=t, trim={1200px 1000px 460px 250px},clip]{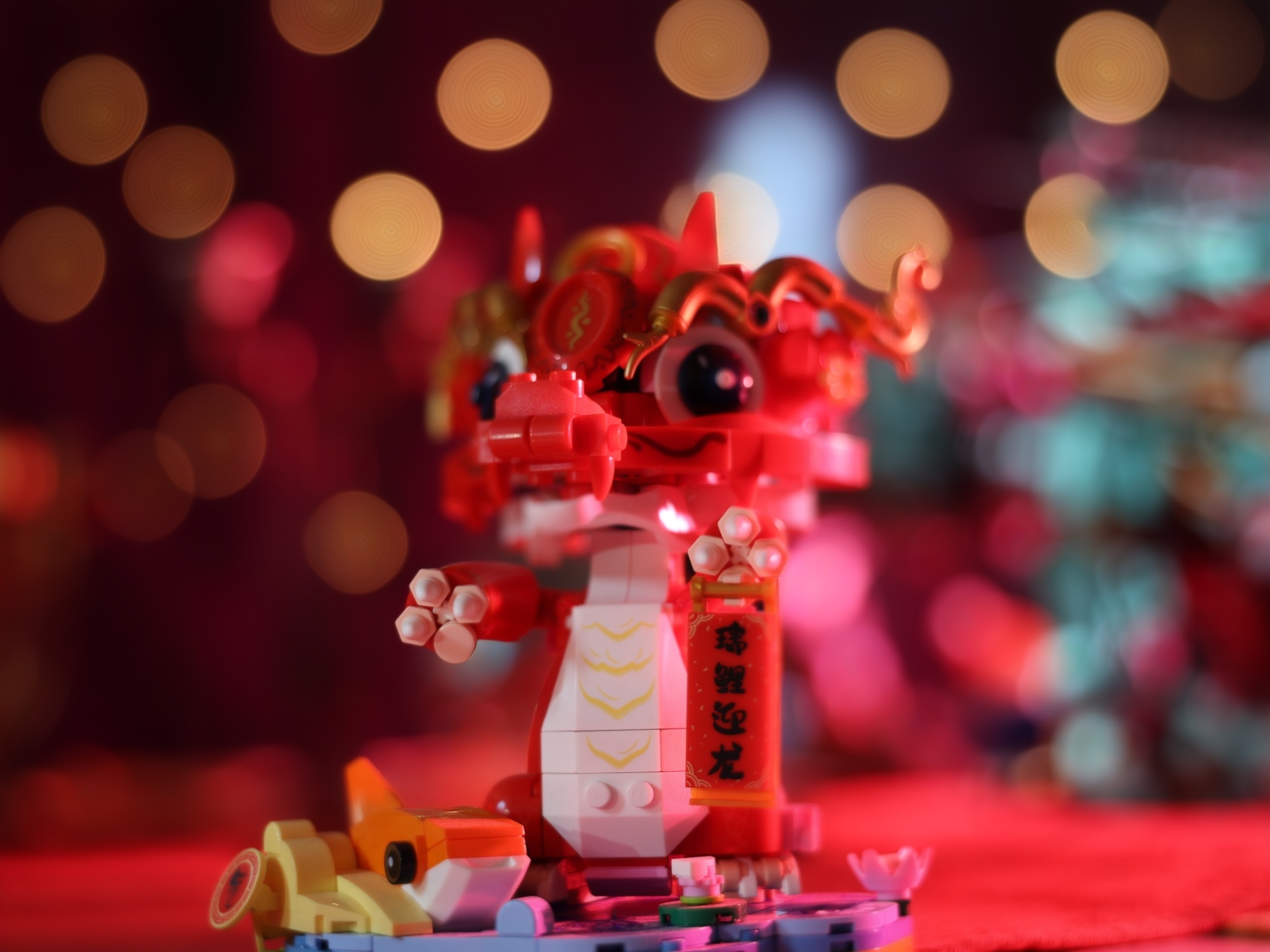} &
        \includegraphics[width=\widthcompp\textwidth,valign=t, trim={1200px 1000px 460px 250px},clip]{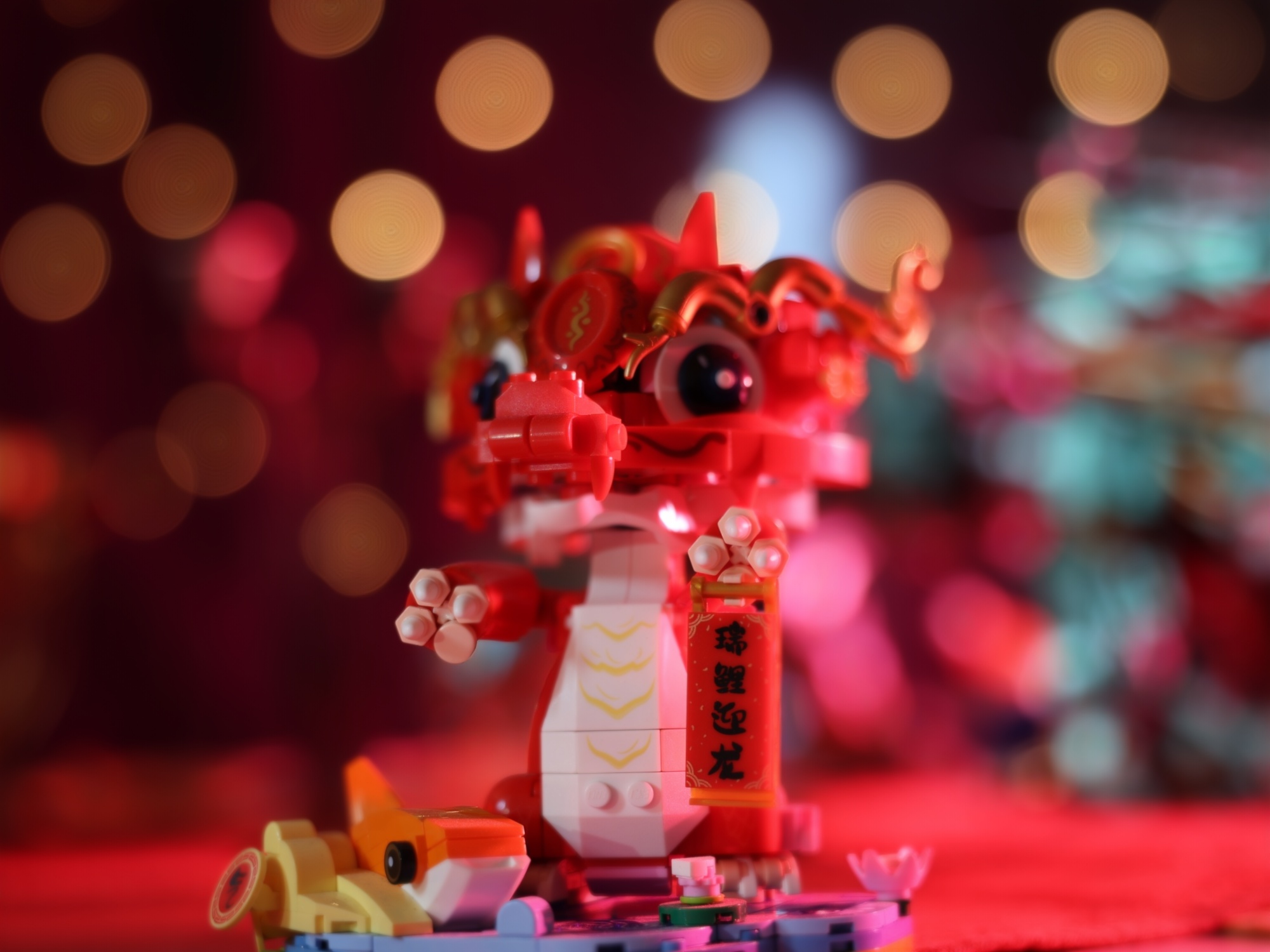} &
        \includegraphics[width=\widthcompp\textwidth,valign=t, trim={1200px 1000px 460px 250px},clip]{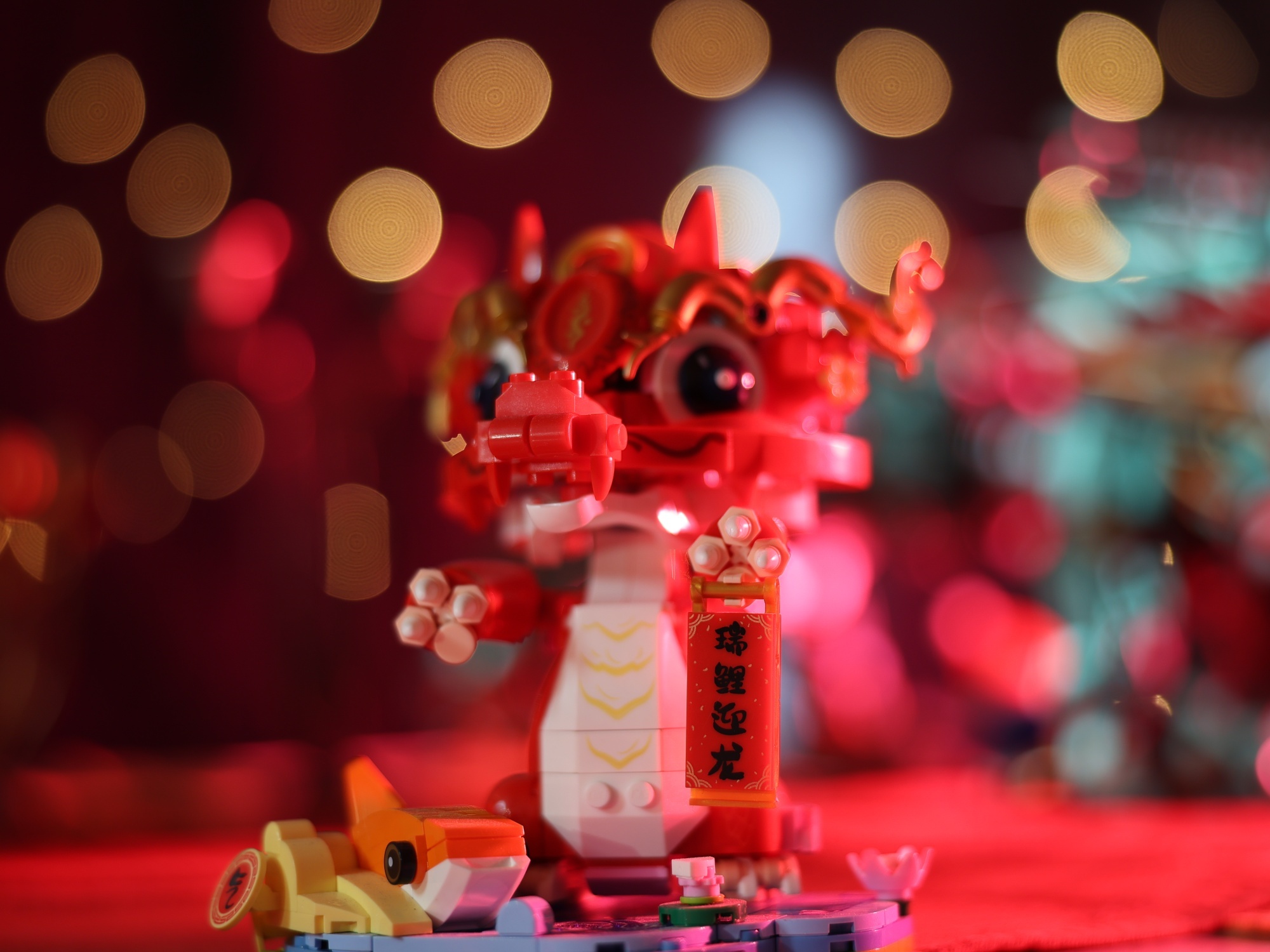} &
        \rotatebox{-90}{\hspace{3.4mm}\fnum{2.0}}
        \\
        \addlinespace[2.0pt]
        & 
        \includegraphics[width=\widthcompp\textwidth,valign=t, trim={1200px 1000px 460px 250px},clip]{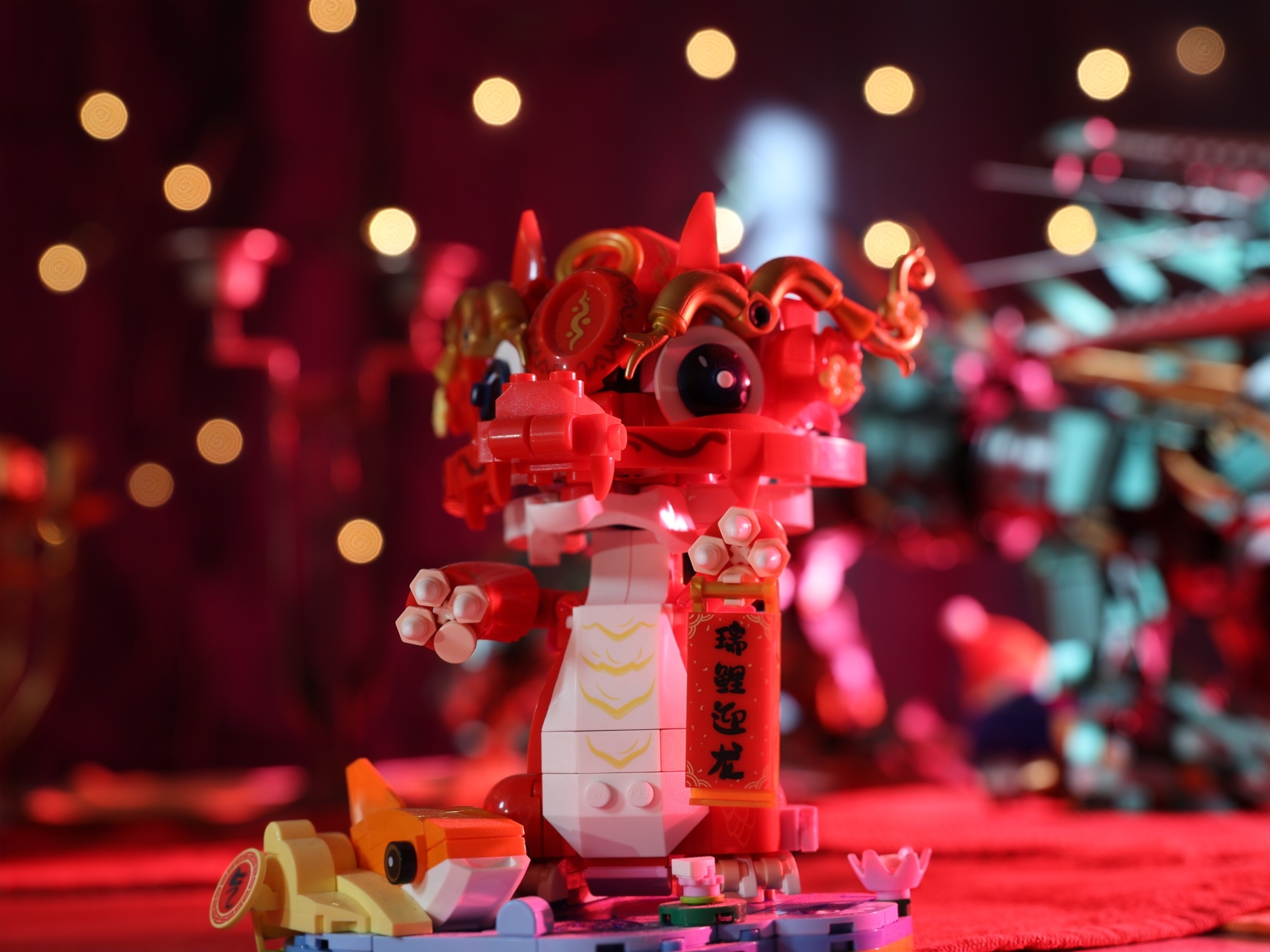} &
        \includegraphics[width=\widthcompp\textwidth,valign=t, trim={1200px 1000px 460px 250px},clip]{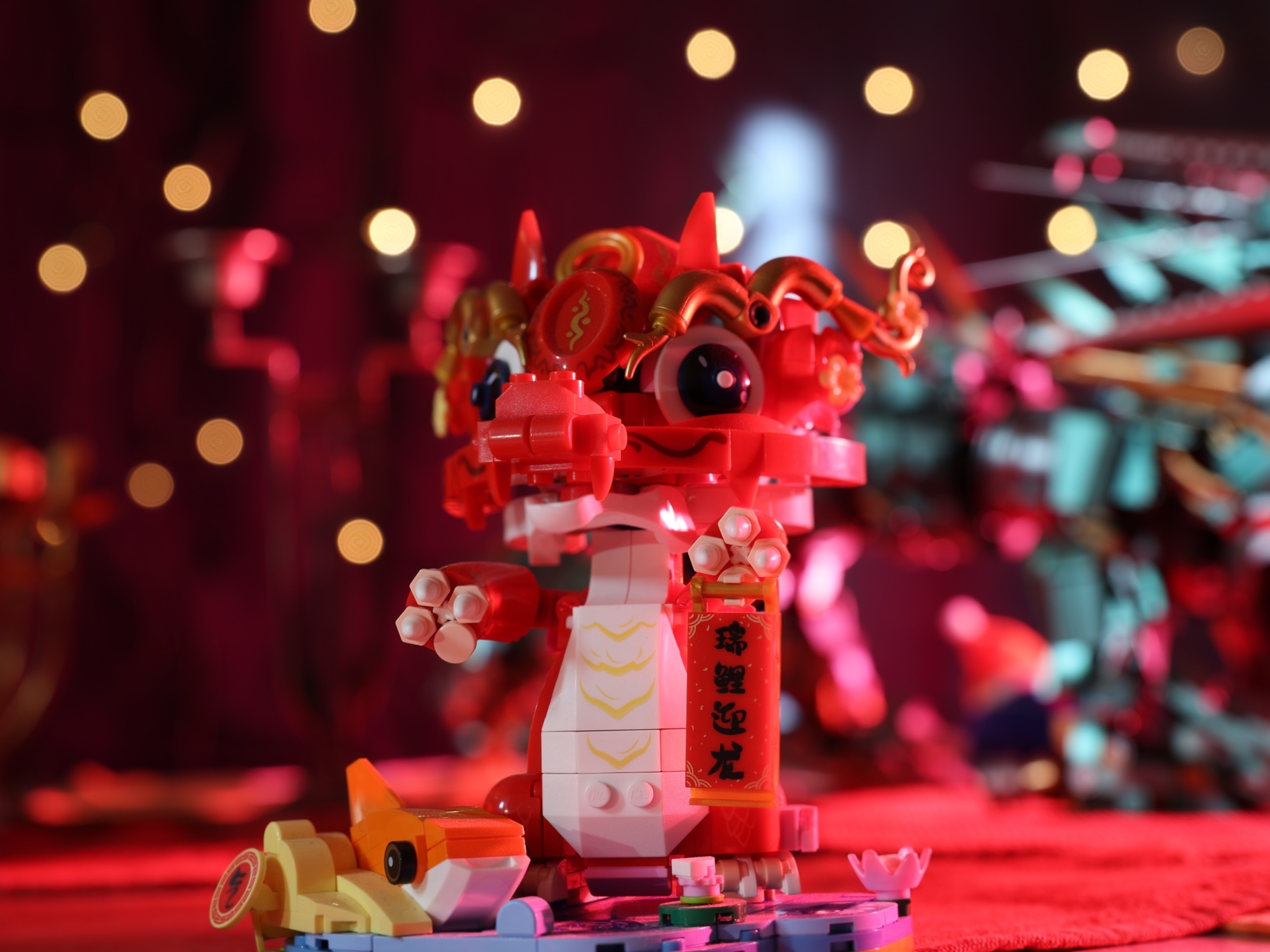} &
        \includegraphics[width=\widthcompp\textwidth,valign=t, trim={1200px 1000px 460px 250px},clip]{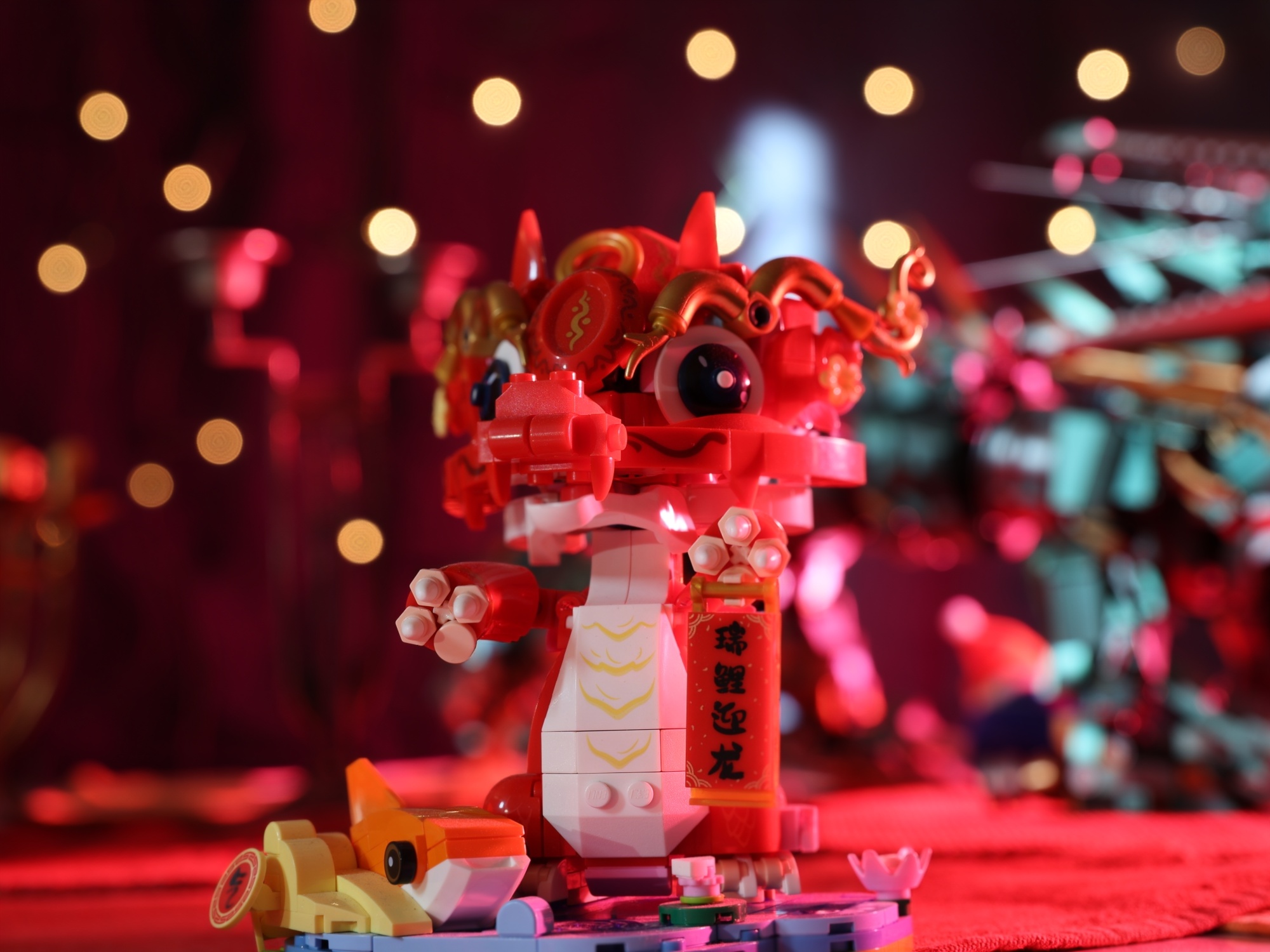} &
        \includegraphics[width=\widthcompp\textwidth,valign=t, trim={1200px 1000px 460px 250px},clip]{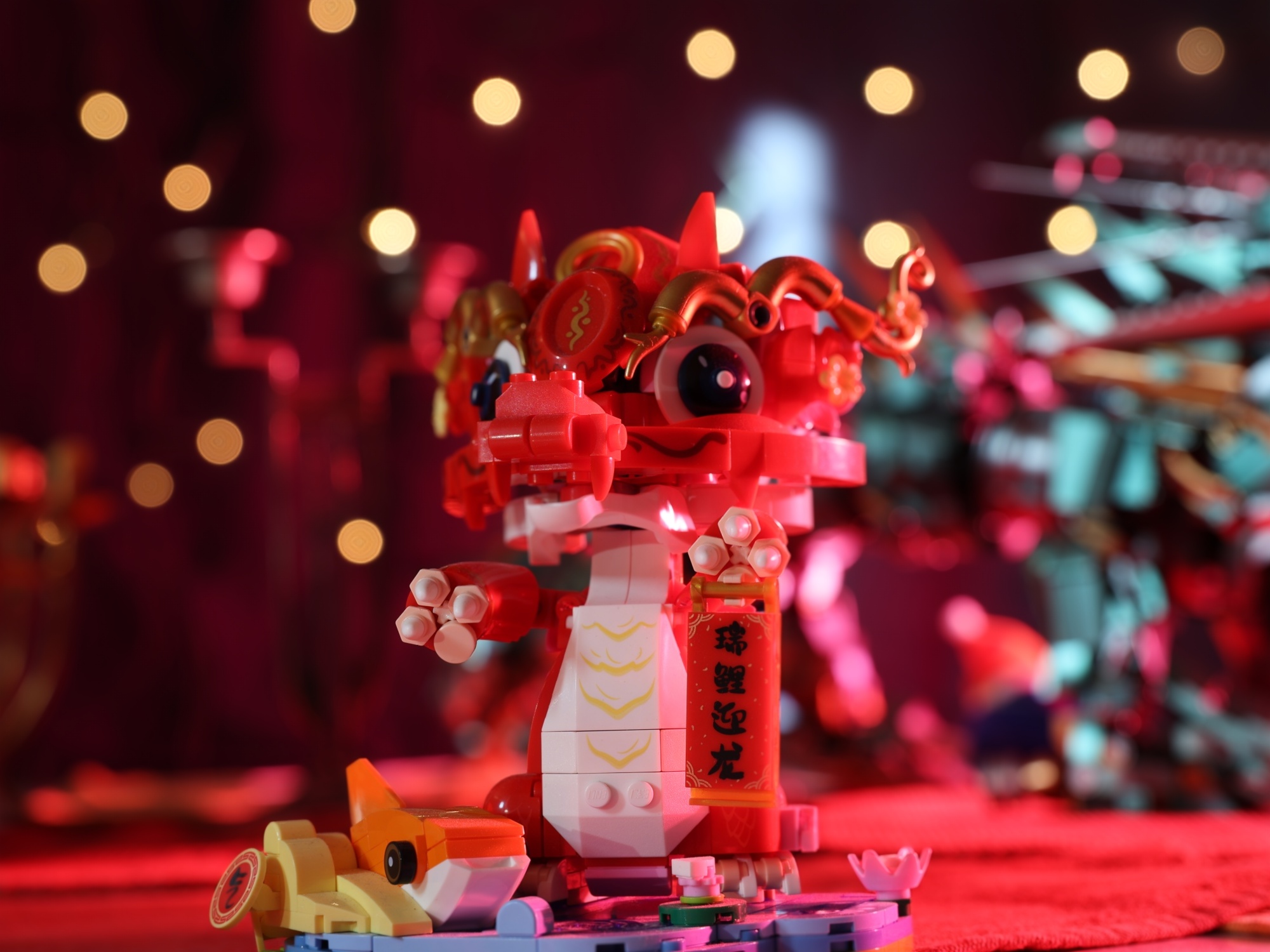} &
        \includegraphics[width=\widthcompp\textwidth,valign=t, trim={1200px 1000px 460px 250px},clip]{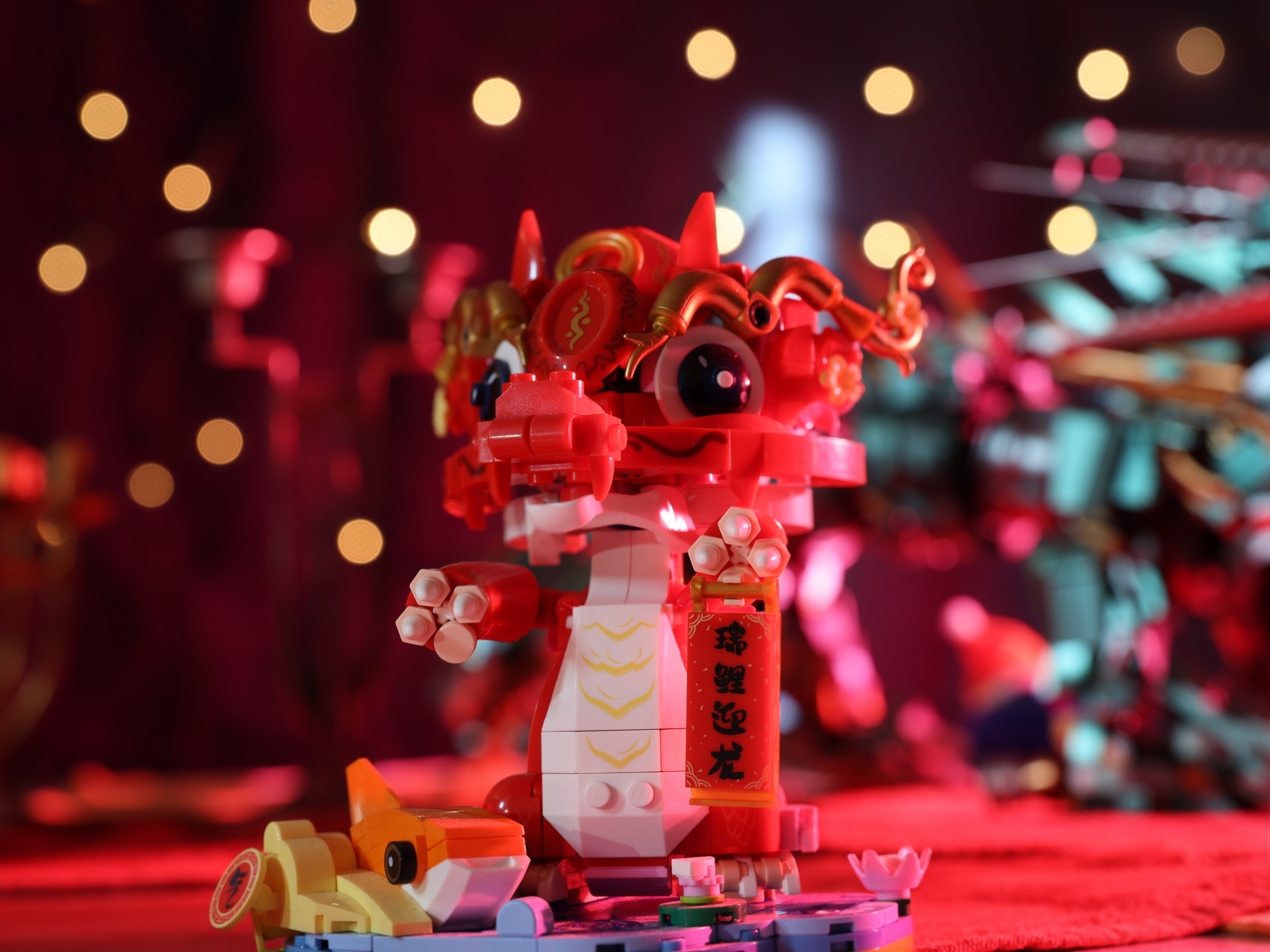} &
        \includegraphics[width=\widthcompp\textwidth,valign=t, trim={1200px 1000px 460px 250px},clip]{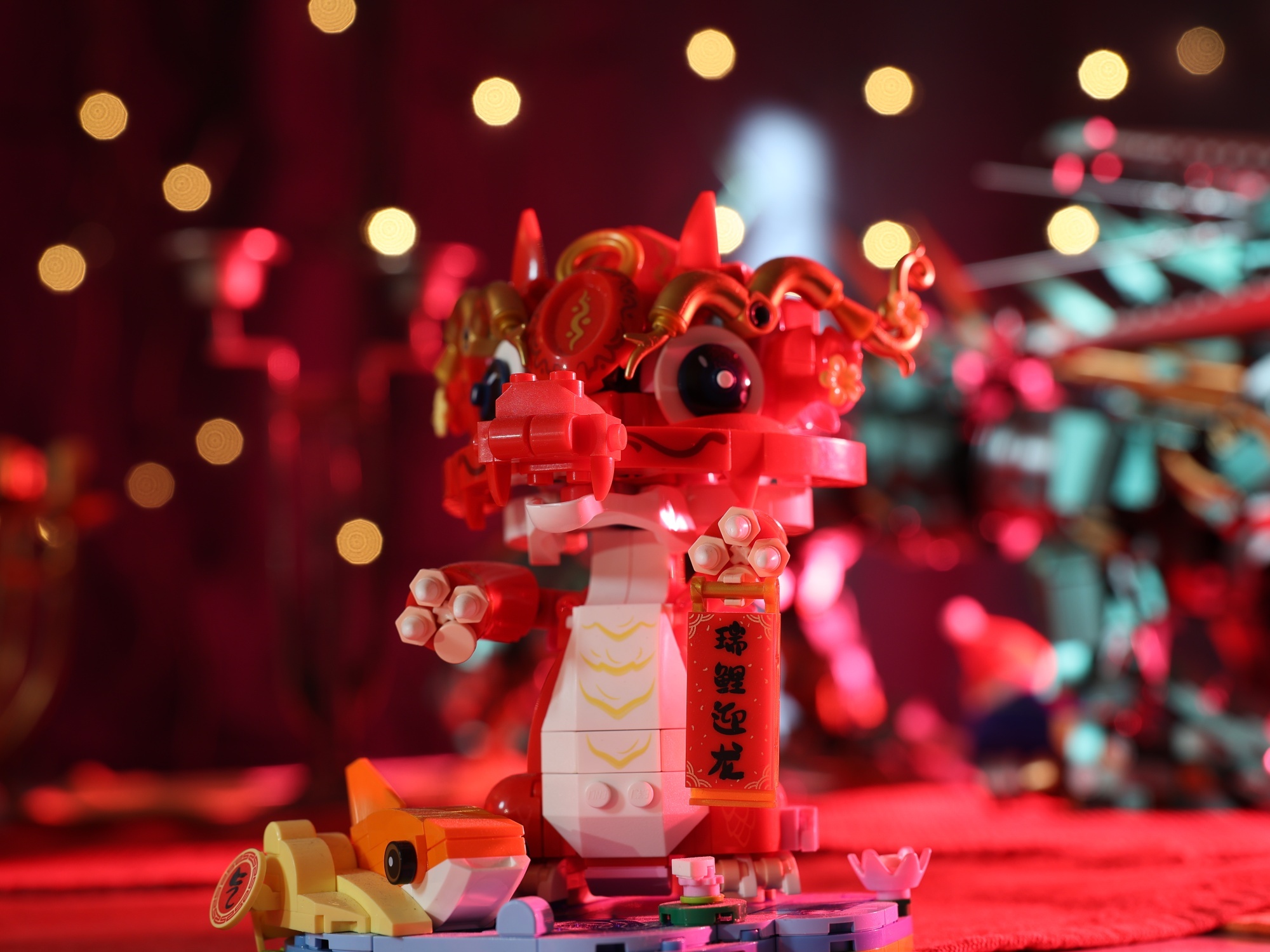} &
        \rotatebox{-90}{\hspace{3.4mm}\fnum{5.0}}
    \\
    \addlinespace[5.0pt]
    \multirow{3}{*}[1.4mm]{\includegraphics[height=\imgcomp\textwidth, trim={0px 200px 0px 000px},clip]{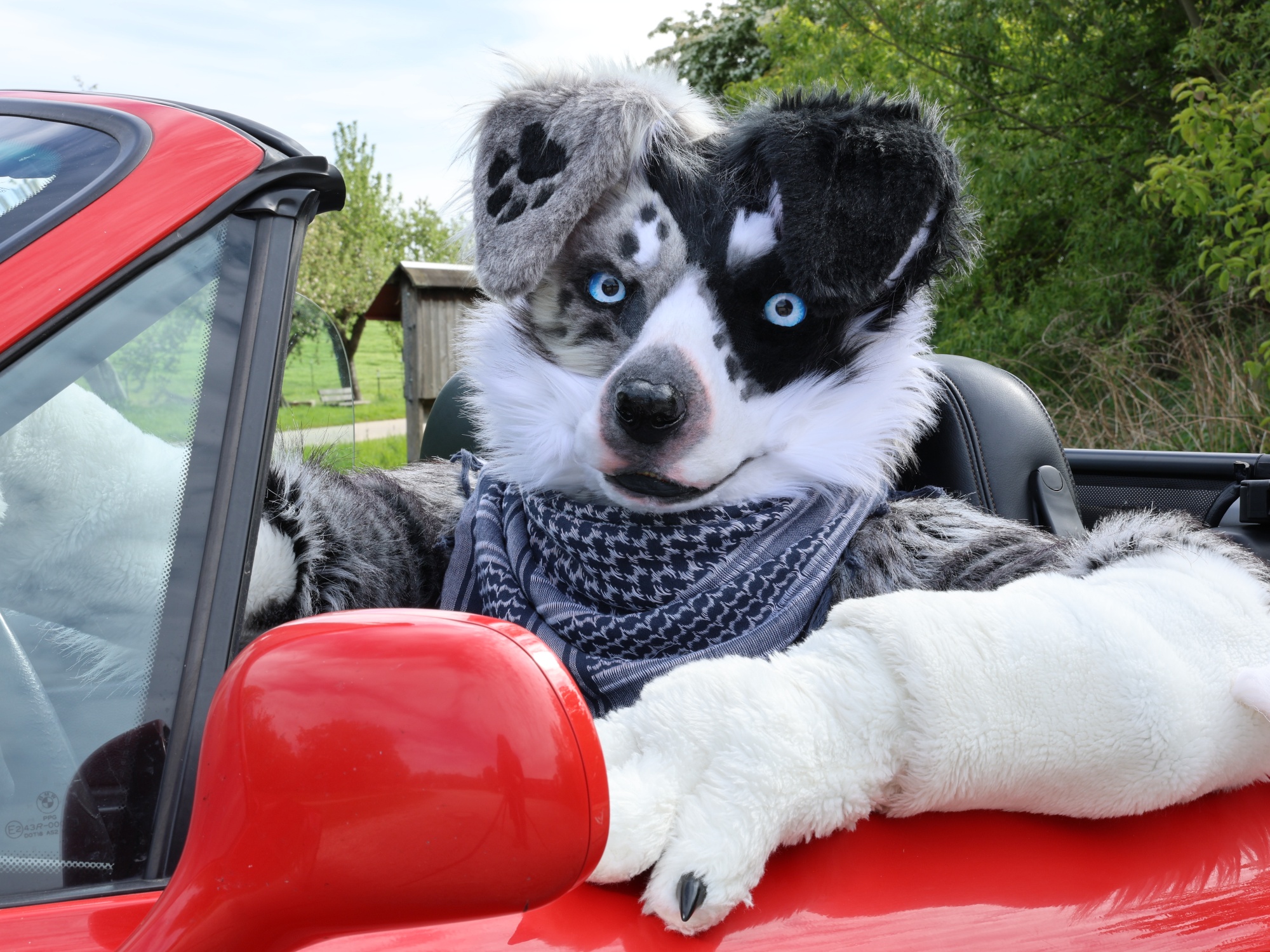}} &
        \includegraphics[width=\widthcompp\textwidth,valign=t, trim={700px 976px 960px 200px},clip]{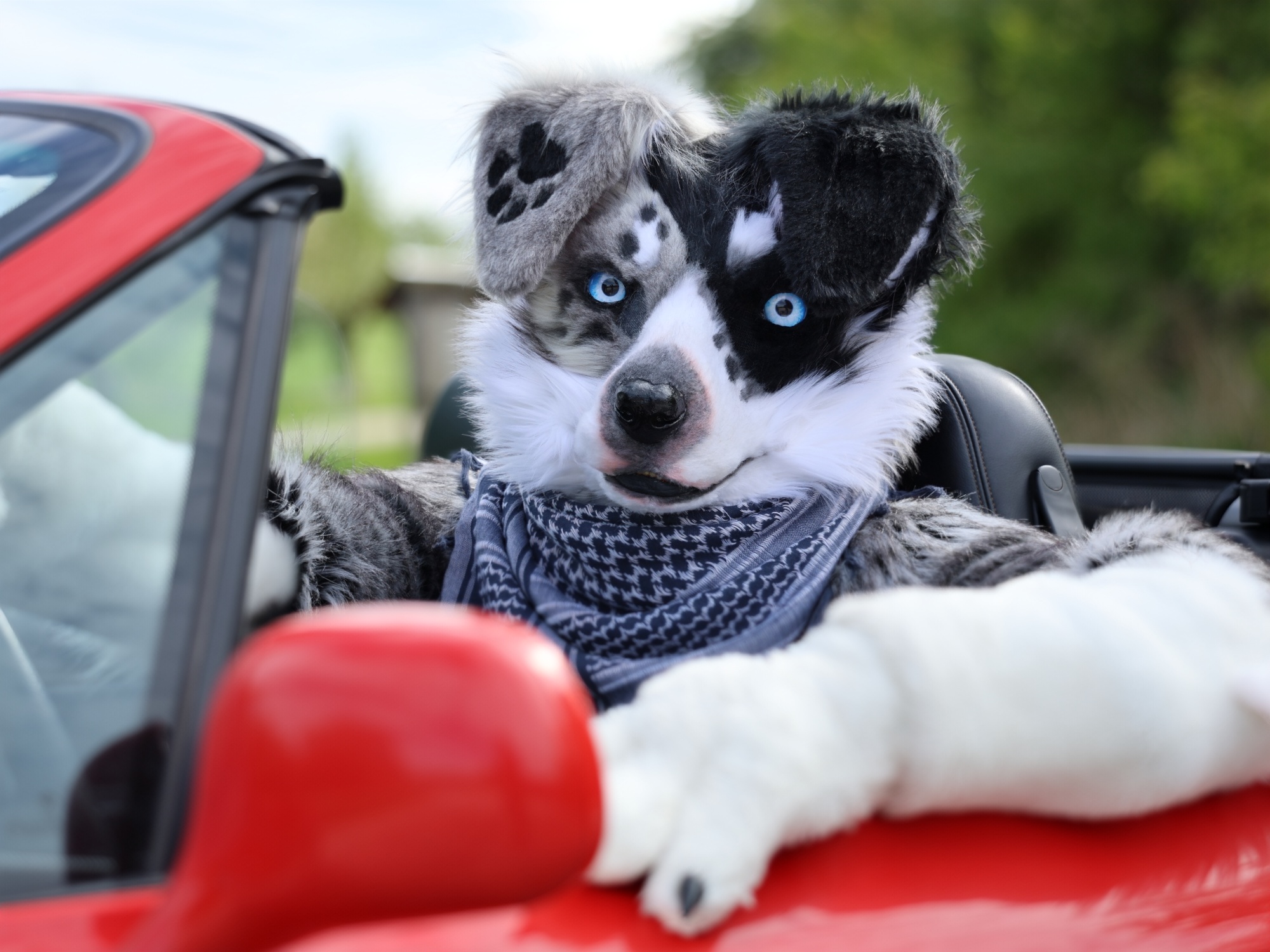} &
        \includegraphics[width=\widthcompp\textwidth,valign=t, trim={700px 976px 960px 200px},clip]{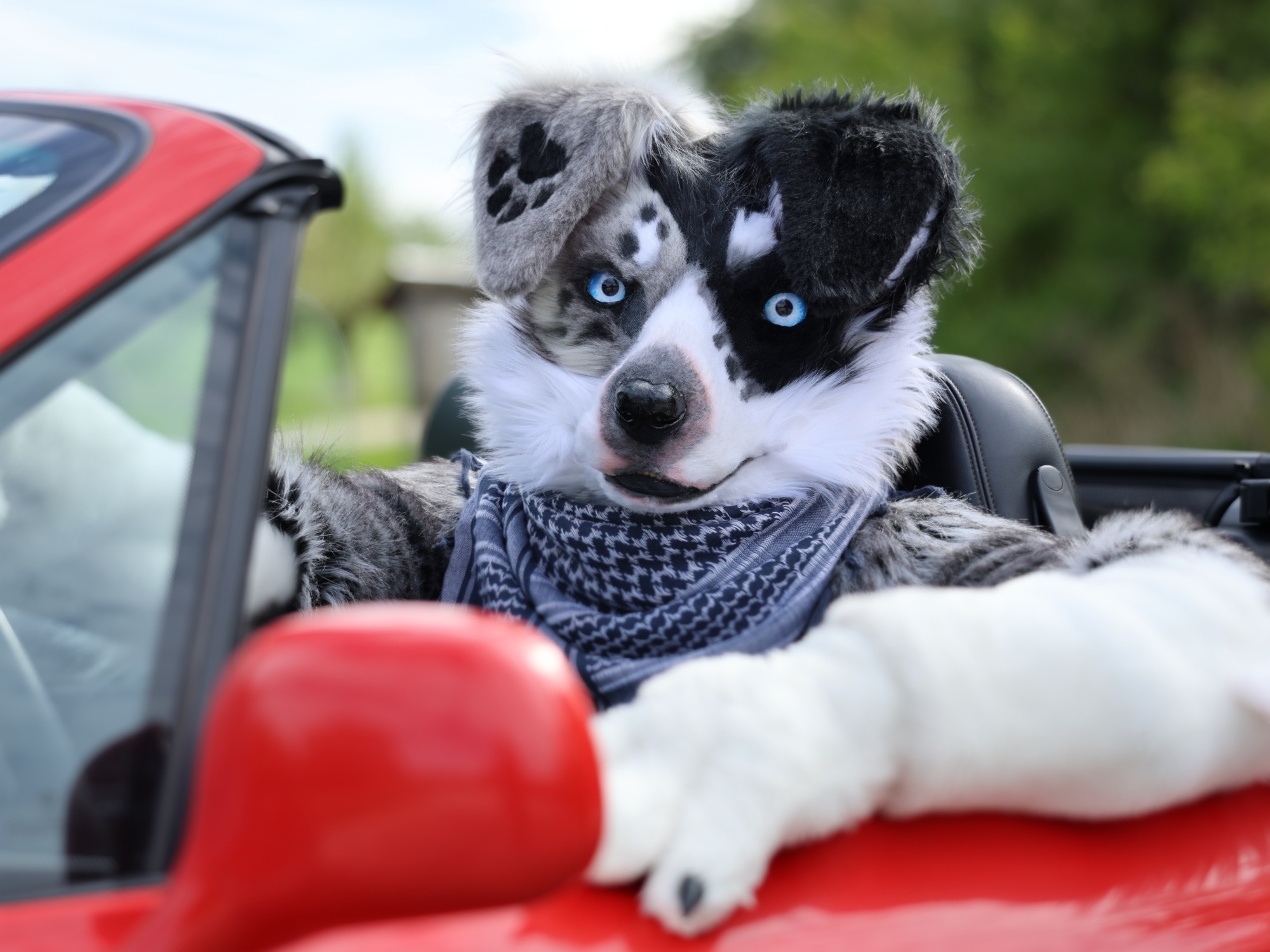} &
        \includegraphics[width=\widthcompp\textwidth,valign=t, trim={700px 976px 960px 200px},clip]{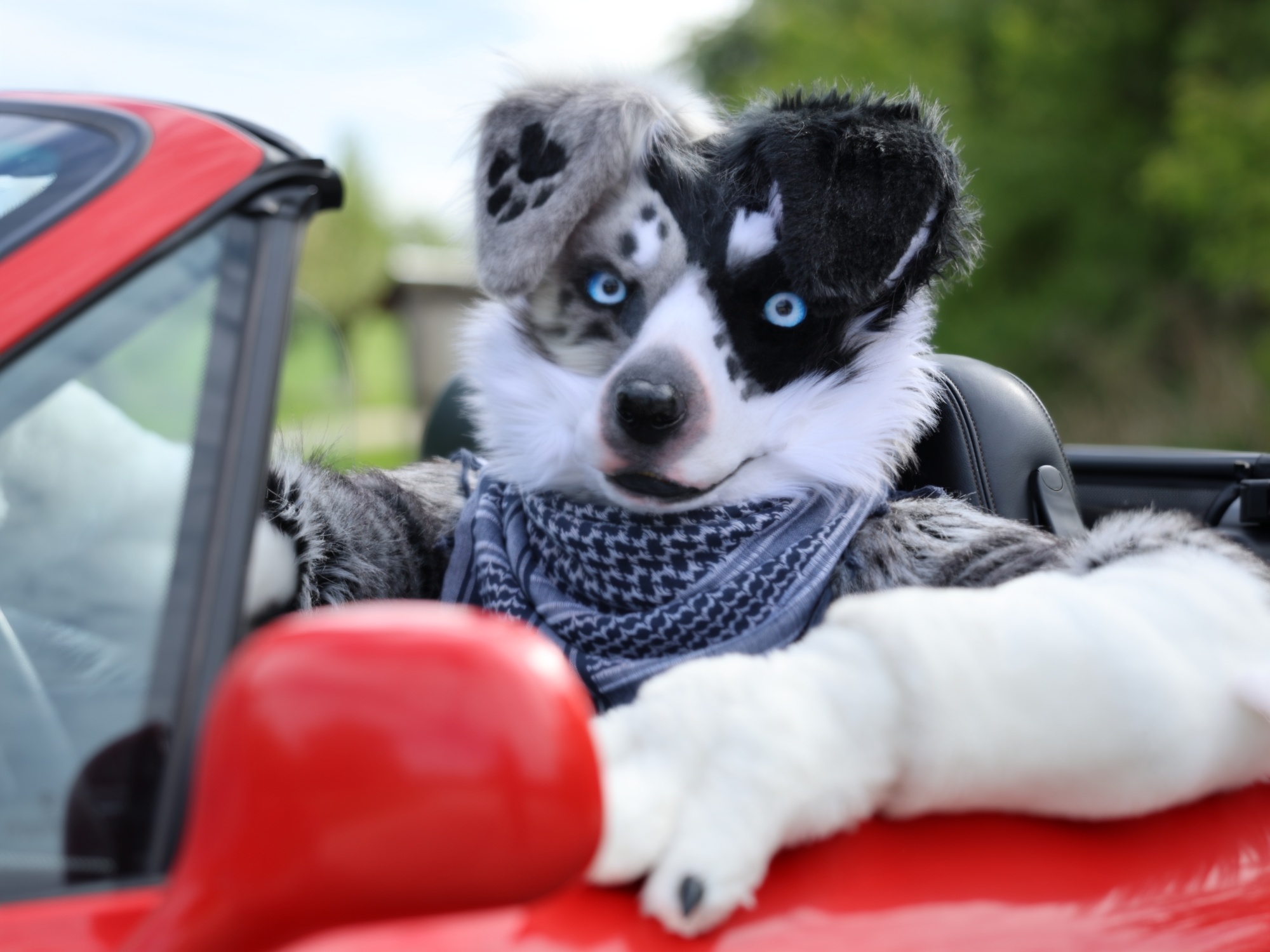} &
        \includegraphics[width=\widthcompp\textwidth,valign=t, trim={700px 976px 960px 200px},clip]{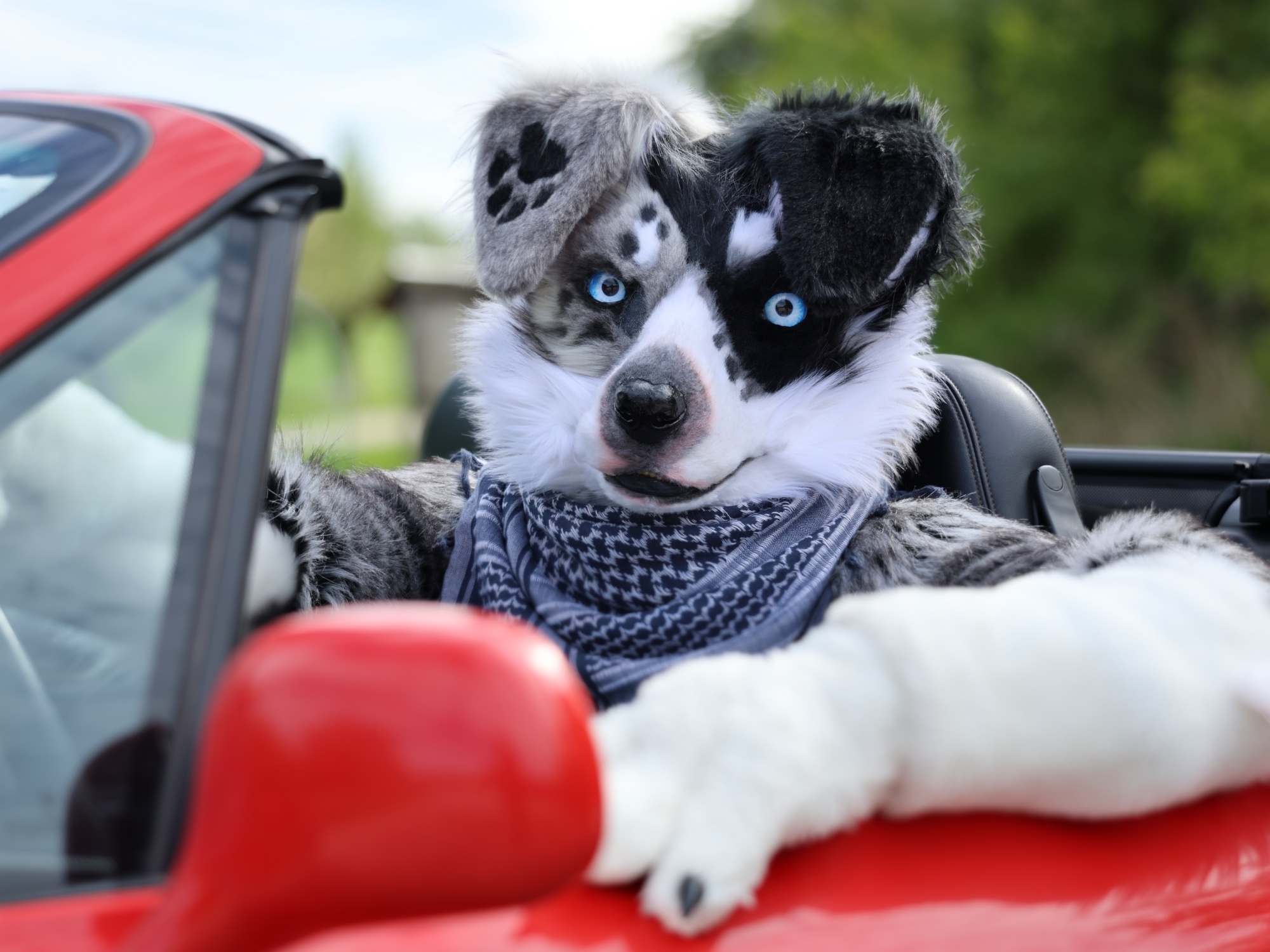} &
        \includegraphics[width=\widthcompp\textwidth,valign=t, trim={700px 976px 960px 200px},clip]{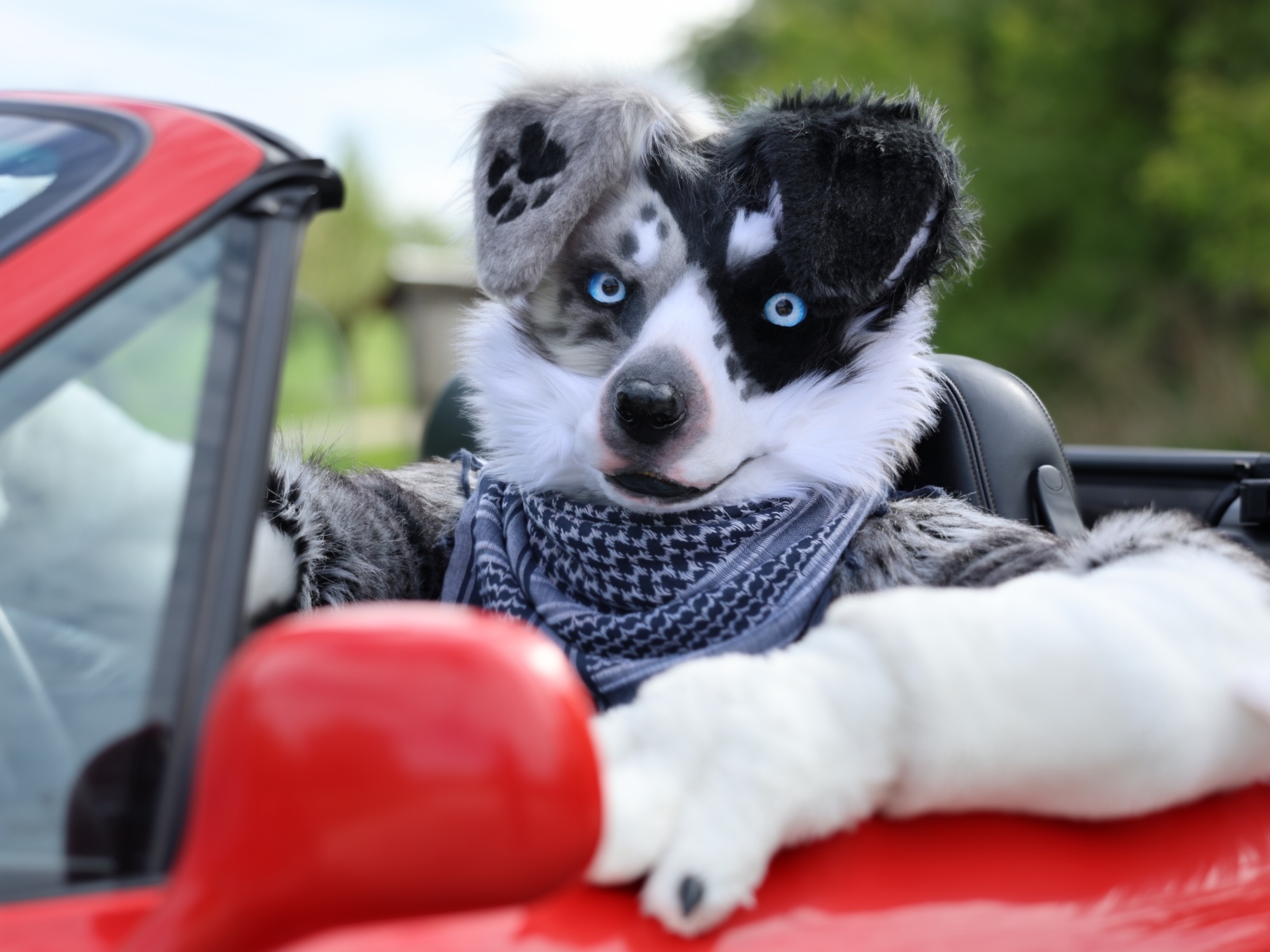} &
        \includegraphics[width=\widthcompp\textwidth,valign=t, trim={700px 976px 960px 200px},clip]{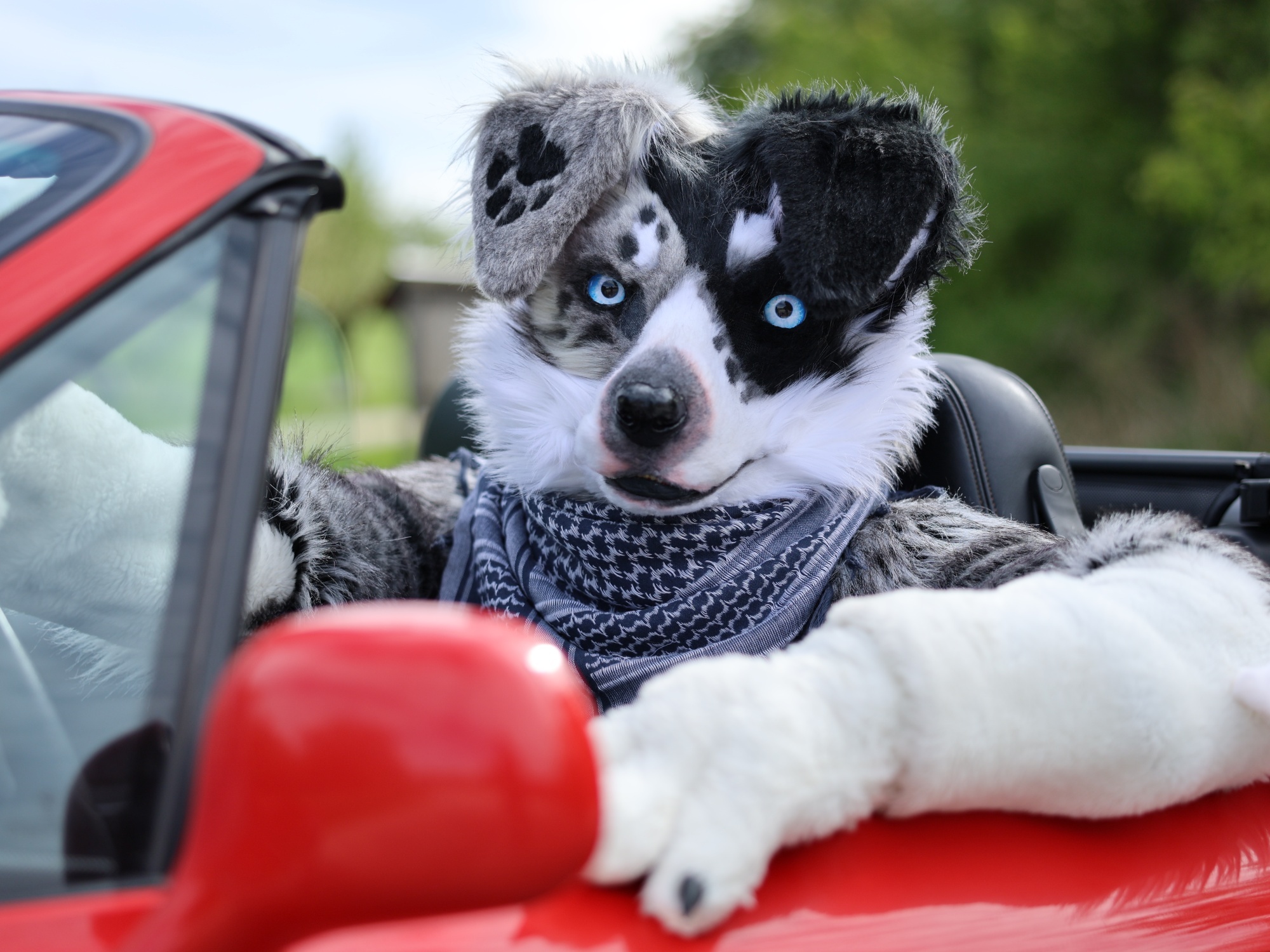} &
        \rotatebox{-90}{\hspace{5.5mm}\fnum{2.0}}
    \\
\end{tabular}
\end{center}
\vspace{-3mm}
\caption{Qualitative comparison between the top-performing methods of the challenge. Please zoom in to note details.}
\label{fig:quali-t1}
\end{figure*}

\cref{tab:quanti_t1} provides the quantitative evaluation of the submissions for both fidelity and perceptual tracks. 
Rankings for each metric are shown as superscripts. 
Many solutions improve upon the large version of the Bokehlicious baseline architecture in terms of reconstruction fidelity; however, in our user study, only the solution from YuFans was preferred over it.
Qualitative examples of the best submissions in this challenge are shown in \cref{fig:quali-t1}.

Because all top solutions are built around the provided Baseline methodology, visual differences are very minor in \cref{fig:quali-t1}, which is also reflected by closely clustered MOS scores.
Moreover, fidelity and perceptual rankings correlate only weakly: the top ranked method in the fidelity track only placed fourth in the perceptual track with a notable difference in MOS.
When inspecting \cref{fig:quali-t1}, the submissions from \textit{NJUST-KMG} and \textit{CV SVNIT} exhibit artifacts, in particular when rendering a strong and distinct Bokeh effect.
For example, the third case shows two misaligned PSFs blended on top of another, possibly caused by output fusion of separate architectural branches, while in the fourth foreground sharpness and detail are reduced.
Similarly, the geometric self-assembly approach of YuFans shows a slight loss of detail, particularly in the representation of the PSF structure.

Overall, these observations suggest that common qualitative metrics used in this iteration might not fully capture perceptual quality for Bokeh rendering: with NJUST-KMG and CV SVNIT, which perform best in PSNR, SSIM and LPIPS respectively, both suffering from image artifacts that reduce their user-study scores.

\section{Challenge Methods}
\label{sec:methods}

\subsection{Davinci}

The network structure proposed by Davinci consists of two distinct stages: the first stage employs the Bokehlicious~\cite{seizinger2025bokehlicious} architecture for controllable bokeh rendering, while the second stage is built upon the NAFNe~\cite{chen2022simple} architecture for refinement.

\noindent\textbf{Coarse Stage: }In the first stage, the network takes the input image and generates a preliminary bokeh rendering result. 
This stage utilizes the Bokehlicious model, which features an efficient CNN en-/decoder with residuals to maintain image detail~\cite{seizinger2025bokehlicious}. 
For deep feature processing, it uses a number of Residual Groups (RGs). 
The core innovation of this stage is the Aperture-Aware Attention mechanism, which mimics the physical lens aperture by adapting the width of the attention mask according to the desired f-stop. 
This allows the network to intuitively control the bokeh strength and generate complex blur kernels. 
The features extracted in this stage serve as visual prompts, providing essential contextual information for the subsequent refinement stage.

\noindent \textbf{Refinement Stage: }The second stage leverages the NAFNet~\cite{chen2022simple} architecture to further refine the coarse output. 
NAFNet represents a significant advancement in lightweight image restoration networks, distinguished by its deliberate exclusion of nonlinear activation functions within its main computational pathways. 
This design choice prevents the disruption of high-frequency information that is crucial for preserving fine image details. The core component of this stage is the NAFNet block, which comprises several key elements:

\noindent \textit{Multi-scale convolutional layers:} These layers are designed to capture features at different spatial scales, allowing the network to handle complex structural variations and refine boundaries effectively.
    
\noindent \textit{Residual connections:} These connections facilitate the seamless flow of information through the network, mitigating the vanishing gradient problem. They enable the network to focus on learning the residual refinements rather than reconstructing the entire image from scratch.
    
\noindent \textit{Channel attention mechanisms:} Integrated within the NAFNet block, these mechanisms allow the network to adaptively emphasize or suppress features across different channels based on their relevance to the refinement task.

During this refined adjustment stage, the network leverages the visual prompts from the Bokehlicious coarse stage via a Prompt Calibration Block. 
By incorporating these prompts, the NAFNet structure can better understand the semantic content and structural boundaries of the image. 
This enables it to accurately separate the background from the subject, recover finer details such as intricate foreground hair, and correct any residual artifacts introduced in the first stage. 
The refinement stage thus ensures that the final output maintains a highly photorealistic bokeh effect while preserving rigorous texture and structural integrity in the in-focus regions.

\subsection{NJUST-KMG}

\begin{figure}
    \centering
    \includegraphics[width=0.99\linewidth]{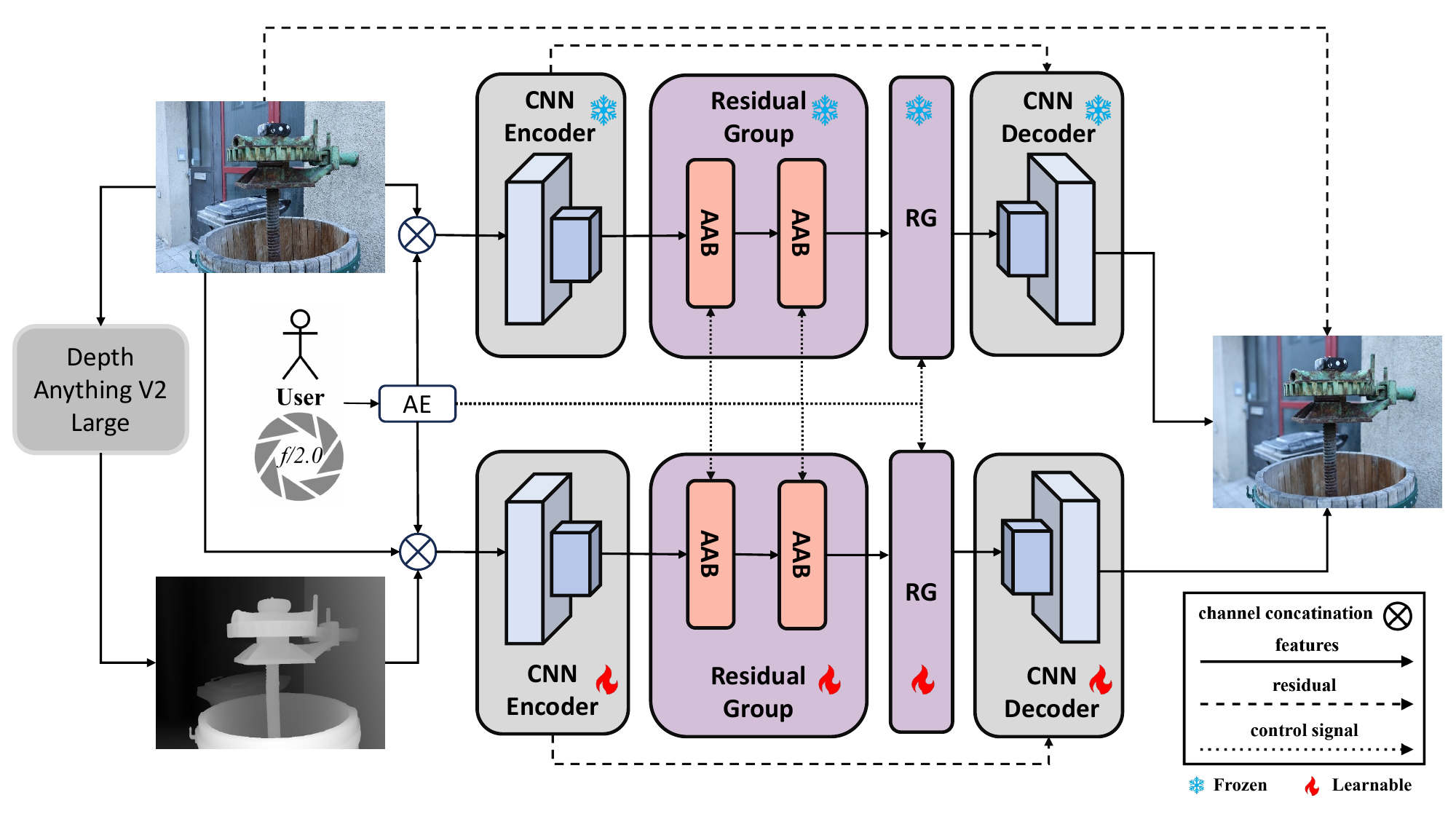}
    \vspace{-3mm}
    \caption{
        Overview of the framework proposed by NJUST-KMG. The upper part corresponds to the original Bokehlicious module \cite{seizinger2025bokehlicious}, while the lower part illustrates the enhanced module with depth map integration. During training, the upper module is frozen and only the lower module is optimized.
    }
    \label{fig:NJUST-KMG}
    \vspace{-3mm}
\end{figure}

The NJUST-KMG team proposes a dual-controllable bokeh rendering framework, as illustrated in Fig.~\ref{fig:NJUST-KMG}. 
Previous work~\cite{zhu2025bokehdiff} shows that depth maps can alleviate the burden of geometric reasoning, allowing the model to focus on defocus rendering.
Inspired by this observation, they employ Depth Anything V2 Large \cite{yang2024depth} to generate depth maps for all input images.

Building upon Bokehlicious$L$ \cite{seizinger2025bokehlicious}, this team extends the network inputs by concatenating the depth map with the RGB image along the channel dimension. 
The depth information is then propagated throughout the network and participates in all subsequent computations.

However, due to the discrepancy between predicted depth maps and true scene depth, a progressive depth-masking strategy is introduced. 
Specifically, large regions of the depth maps are randomly masked at early training stages, then the masking ratio is gradually reduced as training proceeds, mitigating the risk of over-reliance on noisy depth cues.

To further improve robustness, a model averaging strategy is implemented by ensembling (1:1) the outputs of the original Bokehlicious-L model and the proposed depth-augmented model, with additional test-time augmentation during inference.

\subsection{YuFans}


The solution by YuFans exploits the existing capability of the Bokehlicious-Large model~\cite{seizinger2025bokehlicious} through inference-time augmentation.
Specifically, its performance is enhanced through 8$\times$ geometric test-time augmentation (TTA), averaging predictions over 4 rotations $\times$ 2 flips.
A detailed description of the findings of this team has been submitted to the workshop~\cite{ntire26geometric}.

\subsection{CV SVNIT}


The \textbf{HAFT} (Hybrid Aperture-conditioned Feature Transformer) architecture proposed by CV SVNIT, builds upon the \emph{Bokehlicious} architecture, enhancing aperture-aware conditioning and introducing depth-guided refinement for controllable bokeh rendering. 
The overall pipeline consists of three main stages: aperture encoding, backbone feature extraction, and depth-guided refinement.

\noindent\textbf{Aperture Encoding: }
The camera aperture value (f-stop) $f$ determines the strength of the background blur. 
To provide this information to the network, the scalar aperture value is first transformed into a high-dimensional representation using Fourier positional encoding with eight frequency bands. 
The encoded features are then passed through a two-layer multilayer perceptron (MLP) with GELU activation to generate a 64-dimensional aperture embedding $\mathbf{e}_f$.
This embedding represents aperture-dependent information and is used to condition the feature extraction layers of the network.

\noindent\textbf{Backbone Feature Extraction}

The U-Net backbone receives a 7-channel input consisting of:
\textbf{(1)} RGB all-in-focus image (3 channels)
\textbf{(2)} 2-channel positional encoding map
\textbf{(3)} Bokeh strength map (1 channel)
\textbf{(4)} Circle-of-Confusion (CoC) map (1 channel)

The Circle-of-Confusion map provides a physics-based spatial prior for defocus blur and is computed as

\[
\mathrm{CoC}(p) =
\mathrm{clip}\left(
\frac{|D(p) - \tilde{D}|}{f},
0, 1
\right)
\]

where $D(p)$ represents the depth value at pixel $p$ and $\tilde{D}$ denotes the median scene depth.

The aperture embedding $\mathbf{e}_f$ is additionally injected into each encoder and decoder block, instead of only during the input and transformer bottleneck as in~\cite{seizinger2025bokehlicious}.

\noindent\textbf{Depth-guided Refinement:}

The lightweight refinement module aims to improve the quality of subject boundaries and reduce artifacts near depth discontinuities within the initial bokeh-rendered image $\hat{I}$. The refinement module combines the predicted image with additional guidance signals including the depth map $D$, a foreground mask $M$, a focal prior map and a soft focus weight indicating regions that should remain sharp.

These features are fused through convolutional layers and residual blocks to predict a refinement residual $\Delta$. The final output image is obtained as $I^* = \hat{I} + \Delta (1 - W_{\text{focus}}).$
The final convolution layer of the Depth-guided Refinement stage is initialized with zero weights so that the refinement stage initially behaves like an identity mapping and gradually learns to improve rendering quality during training.
\subsection{BIT\_ssvgg}

\begin{figure}
    \centering
    \includegraphics[width=0.99\linewidth]{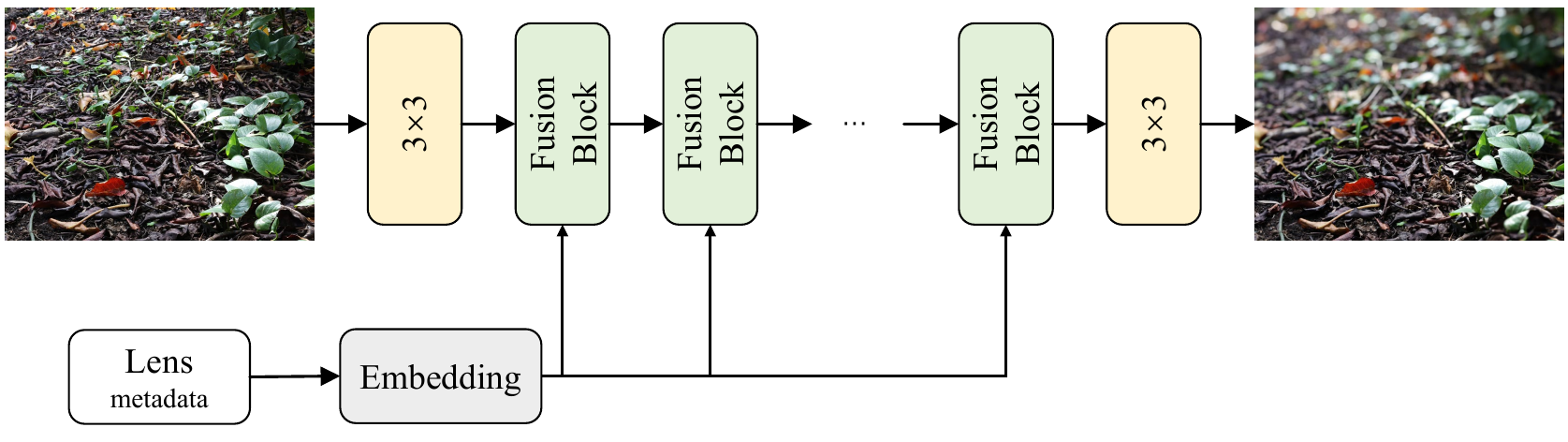}
    \vspace{-3mm}
    \caption{
        Overview of the method proposed by BIT\_ssvgg.
    }
    \label{fig:BIT_ssvgg}
    \vspace{-3mm}
\end{figure}

To address the controllable bokeh rendering task, team BIT\_ssvgg proposes a lightweight conditional image restoration network based on the NAFNet architecture and block design. 
They introduce an explicit aperture-aware conditioning mechanism that enables continuous control over the strength of defocus blur, incorporating the aperture ratio between source and target images as a physically meaningful control signal as shown in \cref{fig:BIT_ssvgg}.

Specifically, when given an input image captured with a small aperture (e.g., \fnum{22}), along with a target aperture value, the aperture ratio between the source and target settings is computed. This ratio serves as a physically interpretable representation of the desired blur strength, reflecting the relative change in depth-of-field. 
The ratio is further transformed using a logarithmic mapping and embedded into a low-dimensional feature vector via a lightweight multi-layer perceptron.

To incorporate aperture-aware control a feature-wise modulation mechanism is introduced to the NAFNet blocks, transforming them into Fusion blocks.
Specifically, the embedded aperture ratio is projected to generate channel-wise scaling and shifting parameters, which are applied to intermediate normalized features via an affine transformation. 
This design enables the network to adapt its feature processing dynamically according to the desired bokeh strengt, allowing a single model to generalize across multiple aperture settings while maintaining stable and interpretable behavior.
\subsection{Centre Borelli}

\begin{figure}
    \centering
    \includegraphics[width=0.99\linewidth]{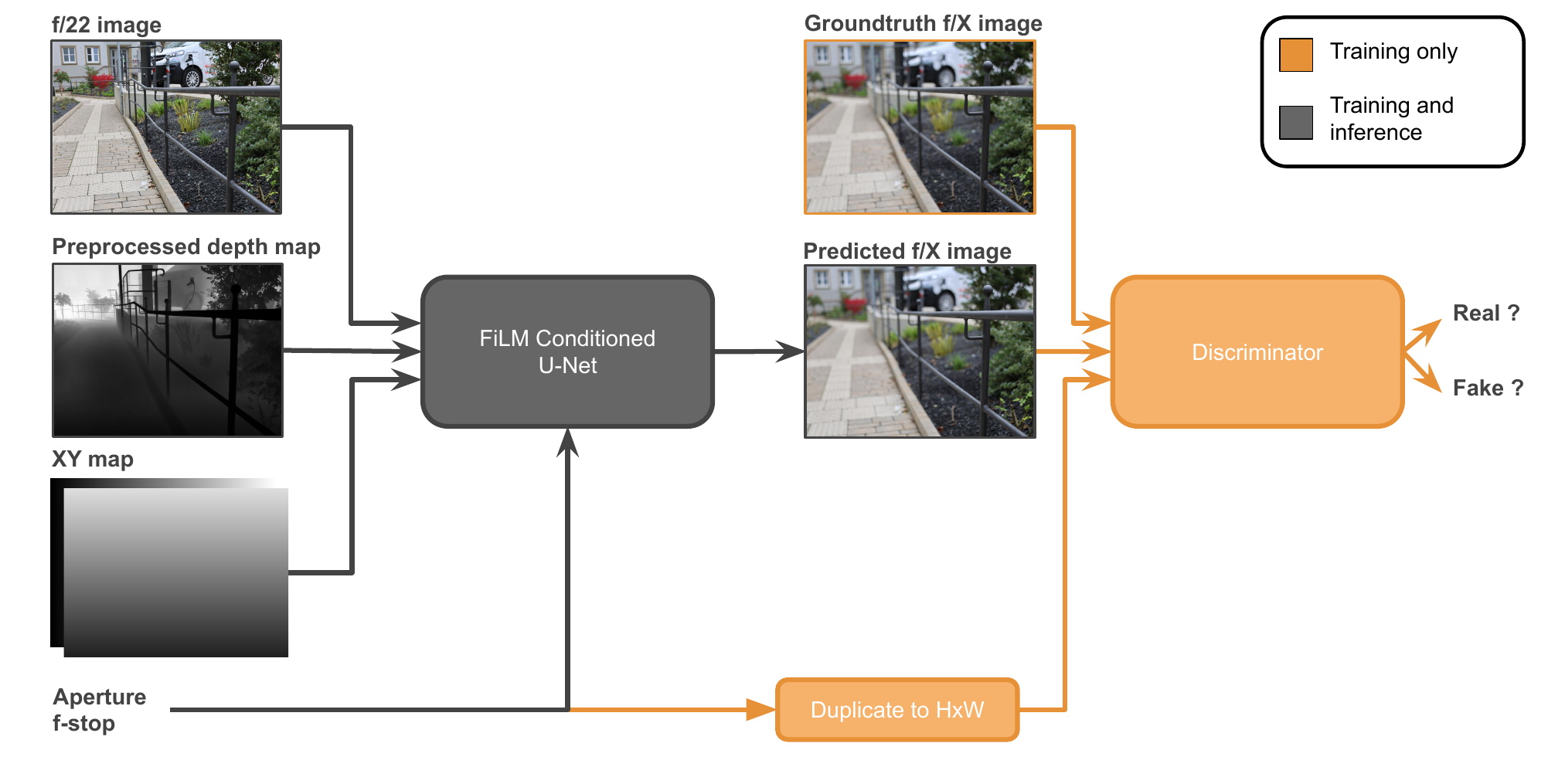}
    \vspace{-3mm}
    \caption{
        Overview of DALU-Net as proposed by Centre Borelli. 
        The generator takes an RGB image together with a monocular depth map and an  $(x,y)$ coordinate map.
        The target aperture is injected into the generator via FiLM conditioning in conditional residual blocks. 
        During adversarial fine-tuning, the discriminator is additionally conditioned on aperture by concatenating a single-channel bokeh-strength map to its input.
    }
    \label{fig:CentreBorelli}
    \vspace{-3mm}
\end{figure}

Team Centre Borelli introduces DALU-Net as visualized in \cref{fig:CentreBorelli}, a depth and location conditioned U-Net for realistic bokeh synthesis that is trained with an adversarial discriminator loss. 
The proposed method takes an all-in-focus RGB image together with two auxiliary inputs: 
(1) a monocular depth map and (2) an $(x,y)$ coordinate map, following prior work that uses spatial coordinates for bokeh rendering~\cite{peng2023selective,seizinger2023bokeh}. 
The depth map is estimated from the input \fnum{22} image using a pre-trained Marigold model~\cite{ke2024repurposing, ke2025marigold}, while the coordinate map provides explicit spatial information for each pixel. 
These inputs help the network capture both depth-dependent blur and location-dependent bokeh variations across the image plane.

The network is based on a U-Net encoder--decoder with skip connections, consisting of a stem convolution, three encoder stages, a bottleneck, and three decoder stages followed by a reconstruction head. 
Each stage uses conditional residual blocks with Group Normalization, SiLU activations, and convolution layers. 
Aperture information is injected through FiLM conditioning~\cite{perez2018film}: the input $f$-number is embedded into a conditioning vector and projected into channel-wise scale and bias parameters, which modulate intermediate features inside each residual block. 
This allows the model to adapt its behavior to different aperture settings and learn aperture-aware blur rendering. A global residual connection is added from the source RGB image to the final prediction, helping preserve scene structure and stabilize training. 
During adversarial fine-tuning, the discriminator is also conditioned on the target aperture by expanding the scalar aperture condition to a single-channel bokeh-strength map and concatinating it with the discriminator input.
\subsection{higasa}

\begin{figure}
    \centering
    \includegraphics[width=0.99\linewidth]{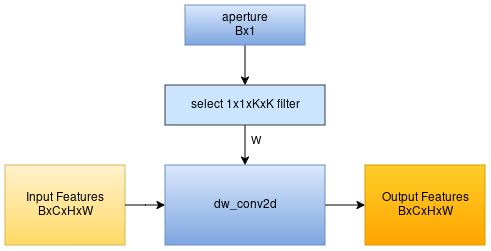}
    \vspace{-3mm}
    \caption{
        The Controllable Bokeh block proposed by higasa.
    }
    \label{fig:higasa}
    \vspace{-3mm}
\end{figure}

The overall approach proposed by higasa extends the classic NAFNet~\cite{chen2022simple} architecture towards controllable bokeh rendering.
Their main contribution is an aperture-dependent selectable-convolution kernel as shown in \cref{fig:higasa}.
Furthermore, they introduce a Correction block that performs aperture and focal length dependent sampling to mitigate misalignment caused by the bokeh effect and lens distortion using a global sampling operation.

The aperture-dependent selectable-convolution kernel block contains a single depthwise convolution kernel for each discrete aperture value. While the aperture-dependent selectable-convolution kernel block is located in the middle of the NAFNet architecture, the Correction block is placed on each decoder skip connection. The Correction block itself contains only a pair of trainable parameters per aperture and focal length value. These two parameters control linear and quadratic grid distortions, respectively.
\subsection{NTR}

\begin{figure}
    \centering
    \includegraphics[width=0.99\linewidth]{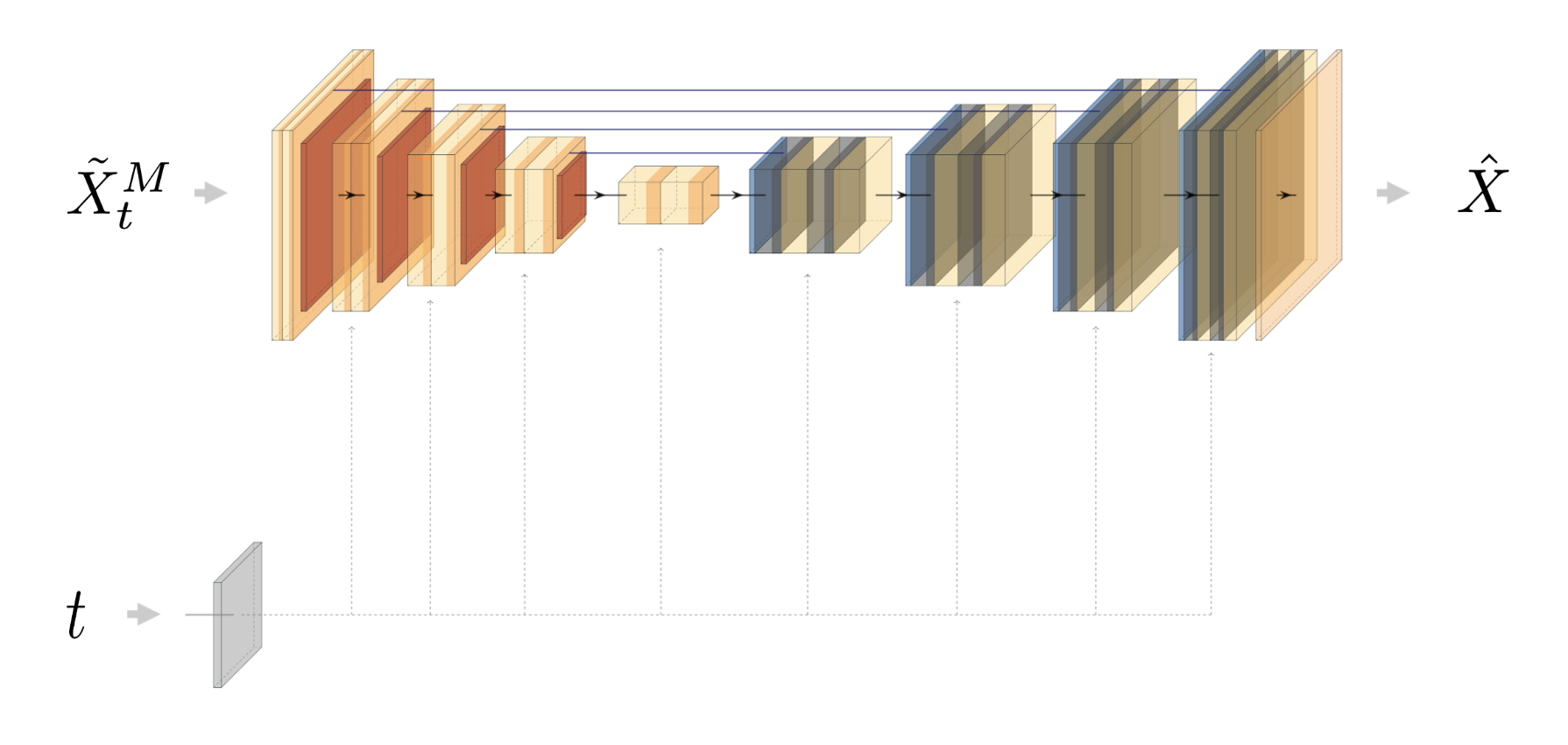}
    \vspace{-3mm}
    \caption{
        The TimeDiffiT architecture proposed by NTR. 
        During MDAE pretraining, a doubly-corrupted input and noise level $\sigma_t$ are fed to the time-conditioned U-Net, producing a restored output. 
        At fine-tuning, the $f/22$ source image replaces the corrupted input.
    }
    \label{fig:NTR}
    \vspace{-3mm}
\end{figure}

The NTR team proposes to fine-tune \textbf{TimeDiffiT} (Time-conditional Diffusion Transformer), a U-Net encoder-decoder with time conditioning, on the Bokeh Rendering task using an L1 pixel reconstruction loss.
The model is initialized from a checkpoint pretrained via Masked Diffusion Autoencoding (MDAE)~\cite{tu2025score}, a self-supervised framework that applies dual corruption: 
\textbf{(1)} random spatial masking of $16{\times}16$ blocks (masking ratio $\in[1\%,75\%]$), and
\textbf{(2)} VE-SDE noise injection~\cite{song2021scorebased}.
The MDAE loss combines an MAE reconstruction term~\cite{he2022masked} with a Corruption2Self diffusion matching term~\cite{tu2025score}.
After MDAE pretraining for 65 epochs, the full network is fine-tuned end-to-end on the Bokeh rendering task.
During inference, tiled processing at $896\times896$px with stride 384px with an additional 8-way geometric self-ensemble~\cite{timofte2016seven} is used.

\section{Conclusion}
The first Controllable Bokeh Rendering Challenge at NTIRE promoted a relatively under-studied but highly interesting task.
A large number of researchers participated in the challenge, with a select group submitting their solution for final evaluation and ranking.
Participants simultaneously target both fidelity and perceptual quality, evaluated across two distinct tracks.
Although no solution in this initial iteration of the challenge has significantly outperformed our baseline, many interesting ideas were explored, advancing the study of Bokeh rendering.
In addition, numerous participants in the challenge contributed substantive feedback and recommendations that will inform the design of subsequent editions.
 
\section*{Acknowledgments}
This work was partially supported by the Humboldt Foundation. We thank the NTIRE 2026 sponsors: OPPO, Kuaishou, and the University of Wurzburg (Computer Vision Lab).

{\small
\bibliographystyle{ieee_fullname}
\bibliography{egbib}
}

\clearpage

\appendix
\section{Teams and Affiliations}
\label{sec:team}

\subsection*{NTIRE 2026 Controllable Bokeh Rendering Challenge Team}
\noindent\textit{\textbf{Members:}}\\ 
\textit{Tim Seizinger}$^1$, Florin-Alexandru Vasluianu$^1$, Jeffrey Chen$^1$,
Zhuyun Zhou$^1$, Zongwei Wu$^1$, Radu Timofte$^1$\\
\noindent\textit{\textbf{Affiliations: }}\\
$^1$ Computer Vision Lab, IFI \& CAIDAS, University of W\"urzburg \\



\subsection*{Davinci}
\noindent\textit{\textbf{Members:}}\\ 
\textit{Dafeng Zhang$^1$
}\\
\noindent\textit{\textbf{Affiliations: }}\\
$^1$ Solo Developer \\


\subsection*{NJUST-KMG}
\noindent\textit{\textbf{Title: }}\\
 Dual-controllable Bokeh Rendering \\
\noindent\textit{\textbf{Members:}}\\ 
\textit{Yipeng Lin$^1$},
Fengqiang Wan$^1$, Kan Lv$^1$,Yang Yang$^{1}$\\
\noindent\textit{\textbf{Affiliations: }}\\
$^1$ Nanjing University of Science and Technology\\


\subsection*{CV SVNIT}
\noindent\textit{\textbf{Title: }}\\
 HAFT: Hybrid Aperture-conditioned Feature Transformer for Controllable Bokeh Rendering\\
\noindent\textit{\textbf{Members:}}\\
\textit{Divyavardhan Singh$^1$},
Hariom Thacker$^1$, Aanchal Maurya$^1$, Hammad$^1$, Kishor Upla$^1$, Kiran Raja$^1$
\\
\noindent\textit{\textbf{Affiliations: }}\\
$^1$ Sardar Vallabhbhai National Institute Of Technology, Surat \\


\subsection*{YuFans}
\noindent\textit{\textbf{Title: }}\\
Bokehlicious-Large with $8\times$ Geometric TTA\\
\noindent\textit{\textbf{Members:}}\\ 
\textit{Wei Zhou$^1$}, Hongyu Huang$^2$\\
\noindent\textit{\textbf{Affiliations: }}\\
$^1$ National University of Singapore\\
$^2$ Zhejiang University\\


\newpage
\subsection*{BIT\_ssvgg}
\noindent\textit{\textbf{Title: }}\\
 A Lightweight Conditional Network for Controllable Bokeh Rendering\\
\noindent\textit{\textbf{Members:}}\\ 
\textit{Hao Yang$^1$},
Ruikun Zhang$^{1}$, 
Liyuan Pan$^1$\\
\noindent\textit{\textbf{Affiliations: }}\\
$^1$ School of Computer Science \& Technology, Beijing Institute of Technology \\


\subsection*{Centre Borelli}
\noindent\textit{\textbf{Title: }}\\
Bokeh simulation with depth and location conditioned U-Net\\
\noindent\textit{\textbf{
Members:}}\\ 
\textit{Yujin Cho$^1$
}\\
\noindent\textit{\textbf{Affiliations: }}\\
$^1$ ENS Paris-Saclay \\



\subsection*{higasa}
\noindent\textit{\textbf{Title: }}\\
Aperture and focal length dependent sampling\\
\noindent\textit{\textbf{Members:}}\\ 
\textit{Grigory Malivenko$^1$}\\
\noindent\textit{\textbf{Affiliations: }}\\
$^1$ Solo Developer \\

\subsection*{NTR}
\noindent\textit{\textbf{Title: }}\\
 MDAE-Pretrained TimeDiffiT for Controllable Aperture Bokeh Rendering\\
\noindent\textit{\textbf{Members:}}\\
\textit{Jiachen Tu$^1$},
Yaoxin Jiang$^1$, Guoyi Xu$^1$, Jiajia Liu$^1$, Yaokun Shi$^1$\\
\noindent\textit{\textbf{Affiliations: }}\\
$^1$ University of Illinois at Urbana-Champaign \\

\end{document}